\documentclass[MAL,biber]{nowfnt} % creates the journal version, needs biber version
\usepackage[utf8]{inputenc}
\usepackage{amssymb,hyperref,amsmath,color}

\allowdisplaybreaks

\DeclareMathOperator*{\argmin}{argmin}

\DeclareSourcemap{
\maps[datatype=bibtex]{
\map[overwrite]{
\step[fieldset=isbn, null]
\step[fieldset=issn, null]
\step[fieldsource=doi, final]
\step[fieldset=url, null]
}}}

\DeclareCiteCommand{\citeyear}
{\usebibmacro{prenote}}
{\bibhyperref{\printfield{year}}\bibhyperref{\printfield{extrayear}}}
{\multicitedelim}
{\usebibmacro{postnote}}
\title{User-friendly Introduction to PAC-Bayes Bounds}

\maintitleauthorlist{
Pierre Alquier \\
ESSEC Business School \\
pierre.alquier.stat@gmail.com
}

\issuesetup
{%
 copyrightowner={P.~Alquier},
 volume        = 17,
 issue         = 2,
 pubyear       = 2024,
 copyrightyear = 2024,
 isbn          = 978-1-63828-326-3,
 eisbn         = 978-1-63828-327-0,
 doi           = 10.1561/2200000100,
 firstpage     = 174, %Explain
 lastpage      = 303
 }

\usepackage{mwe}

\author[1]{Alquier,Pierre}

\affil[1]{ESSEC Business School, Asia-Pacific Campus, Singapore; pierre.alquier.stat@gmail.com}

\articledatabox{\nowfntstandardcitation}

\begin{document}

\makeabstracttitle

\begin{abstract}
Aggregated predictors are obtained by making a set of basic predictors vote according to some weights, that is, to some probability distribution.
Randomized predictors are obtained by sampling in a set of basic predictors, according to some prescribed probability distribution.

Thus, aggregated and randomized predictors have in common that their definition rely on a probability distribution on the set of predictors. In statistical learning theory, there is a set of tools designed to understand the generalization ability of such predictors: PAC-Bayesian or PAC-Bayes bounds.

Since the original PAC-Bayes bounds~\citep{sha1997,mca1998}, these tools have been considerably improved in many directions. We will for example describe a simplified version of the localization technique~\citep{cat2003,cat2007} that was missed by the community, and later rediscovered as ``mutual information bounds''. Very recently, PAC-Bayes bounds received a considerable attention. There was workshop on PAC-Bayes at NIPS 2017, {\it (Almost) 50 Shades of Bayesian Learning: PAC-Bayesian trends and insights}, organized by B. Guedj, F. Bach and P. Germain. One of the reasons of this recent interest is the successful application of these bounds to neural networks~\citep{dzi2017}. Since then, this is a recurring topic of workshops in the major machine learning conferences.

The objective of these notes is to provide an elementary introduction to PAC-Bayes bounds.
\end{abstract}

\chapter{Introduction}
\label{section:introduction}

In a supervised learning problem, such as classification or regression, we are given a data set, and we 1) {\it fix a set of predictors} and 2) {\it find a good predictor in this set}.

For example, when doing linear regression, you 1) chose to consider only linear predictors and 2) use the least-square method to chose your linear predictor.

In this tutorial, we will rather focus on ``randomized'' or ``aggregated'' predictors. By this, we mean that we will replace 2) by 2') {\it define weights on the predictors and make them vote according to these weights} or by 2'') {\it draw a predictor according to some prescribed probability distribution}.

In this first section, we will introduce the main concepts of machine learning theory, and their mathematical notations. We will briefly introduce PAC bounds, that allow to control the generalization error of a predictor. These tools will allow to formalize properly the notion of ``randomized'' or ``aggregated'' predictors, and to introduce PAC-Bayes bounds.

\section{Machine Learning and PAC Bounds}
\label{subsec:intro:ML}

\subsection{Machine learning: notations}

In a supervised learning problem, the objective is to learn from examples to assign labels to objects. Objects can be images, videos, e-mails... The set of all possible objects will be denoted by $\mathcal{X}$. In all the examples we mentioned, it is possible to encode the objects by (large enough) vectors, and thus, we will often have $\mathcal{X} \subseteq \mathbb{R}^d$, where $\mathbb{R}$ is the set of real numbers. The set of labels will be denoted by $\mathcal{Y}$.

The most classical examples of supervised learning problems are binary classification and regression. In binary classification, $\mathcal{Y}=\{0,1\}$. Examples includes spam detection: in this case, objects in $\mathcal{X}$ are e-mails, and the label is $1$ if the e-mail is a spam, and $0$ otherwise. In regression, labels can be any real number $\mathcal{Y}=\mathbb{R}$. This is the case when we try to predict a numerical quantity such as ${\rm CO}_2$ emissions, temperature, etc.

A predictor is a function $f:\mathcal{X}\rightarrow\mathcal{Y}$: for each object $x$, it returns a label $f(x)$. We are usually interested in parametric sets of predictors. That is, we consider $\{f_\theta,\theta\in\Theta\}$ where $\Theta$ is any set, called the parameter set, and each $f_\theta$ is a predictor. For example, in linear regression, a common set of predictors is $f_\theta(x) = \left<x,\theta\right> \in \mathcal{Y} = \mathbb{R}$, with $\mathcal{X} = \Theta = \mathbb{R}^d$. In classification, we can define with the same $\mathcal{X}$ and $\Theta$,
 $$ f_\theta(x) = \left\{
\begin{array}{l} 1 \text{ if }  \left<x,\theta\right> \geq 0 ,\\
0 \text{ otherwise.}
\end{array} \right.  $$
Other examples include neural networks with a fixed architecture, $\theta$ being the weights of the network.
Predictors are sometimes refered to as classifiers in the classification setting, and as regressors in regression.

Assume now that a pair label-object, $(x,y)\in\mathcal{X}\times\mathcal{Y}$, is given. A predictor $f$ will propose a prediction $f(x)$ of the label $y$. If $f(x)=y$, the predictor $f$ predicts the label correctly, otherwise, it makes a mistake. In order to quantify how serious a mistake is, we usually measure it by a loss function. In these notes, a loss function can be any function $\ell:\mathcal{Y}^2 \rightarrow [0,+\infty) $ such that $\ell(y,y)=0$ for any $y\in\mathcal{Y}$; $\ell(f(x),y)$ will be interpreted as the cost of the prediction error.
In classification, the most natural loss function is:

\noindent
$$ \ell(y',y) = \left\{
\begin{array}{l} 1 \text{ if } y'\neq y, \\
0 \text{ if } y' = y.
\end{array} \right. $$
We will refer to it as the 0-1 loss function, and will use the following shorter notation: $\ell(y',y)=\mathbf{1}(y\neq y')$. For computational reasons, it is more convenient to use convex loss functions. For example, in binary classification: $\ell(y',y)=\max(1-yy',0)$ (the hinge loss). In regression problems, the most popular examples are $\ell(y',y)=(y'-y)^2$ the quadratic loss, or $\ell(y',y)=|y'-y|$ the absolute loss. The original PAC-Bayes bounds of \citet{mca1998} were stated in the special case of the 0-1 loss, and this is also the case of most bounds published since then. However, we will see in Section~\ref{section:tight} that their extension to any bounded loss is direct. Some PAC-Bayes bounds for regression with the quadratic loss were proven for example by \citet{cat2004}. \textbf{From now, and until the end of Section~\ref{section:oracle}, we assume that $0\leq \ell \leq C$.} This is typically the case in classification with the 0-1 loss, or in regression with quadratic loss under the additional assumption that $f_\theta(x)$ and $y$ are bounded. We will discuss how to get rid of this assumption in Section~\ref{section:not:bounded:iid}.

Assume we want to build a machine to predict the label of objects it will encounter in the future. Of course, we don't know these objects in advance, nor their labels. A way to model this uncertainty is to assume that a future pair object-label is a random variable $(X,Y)$ taking values in $\mathcal{X}\times\mathcal{Y}$. Let $P$ denote the probability distribution\footnote{Formally, we can only define a probability distribution on $\mathcal{X}\times\mathcal{Y}$ if it is equipped with a $\sigma$-algebra. Let $\mathcal{B}$ be such a $\sigma$-algebra. Essentially, the only thing that matters is that the loss function $\ell$ and the predictors $f_\theta(\cdot)$ are measurable functions, which is satisfied by all classical examples. Note that $\mathcal{B}$ will no longer appear explicitly in this tutorial.} of $(X,Y)$. The expected prediction mistake is thus $\mathbb{E}_{(X,Y)\sim P} [\ell( f(X),Y ) ]$. This is usually refered to as the (generalization) risk of $f$. As it is a very important notion in machine learning, we introduce the notation
$$ R(f) = \mathbb{E}_{(X,Y)\sim P}[\ell(f(X),Y)] .$$
As we will focus on predictors in $\{f_\theta,\theta\in\Theta\}$, we define
$$ R(\theta) := R(f_\theta) $$
for short. A good strategy would be to implement in our machine a predictor $f_\theta$ such that $R(\theta)$ is as small as possible -- ideally, we should implement $f_{\theta^*}$ where $R(\theta^*) = \inf_{\theta\in\Theta}R(\theta)$, if this infimum is reached. Unfortunately, there is a major difficulty: we don't know the distribution $P$ of $(X,Y)$ in practice. Check the examples above: we are not able to describe the probability distribution of images we will see in the future, or of e-mails we will receive.

Instead, we will train our machine based on examples. That is, we assume that we can access a sample of pairs object-label, that we will call the data, or the observations: $(X_1,Y_1),\dots,(X_n,Y_n)$. {\bf From now, and until the end of Section~\ref{section:oracle}, we assume that $(X_1,Y_1),\dots,(X_n,Y_n)$ are i.i.d. from $P$}. That is, they are ``typical examples'' of the pairs object-label the machine will have to deal with in the future. For short, we put $\ell_i(\theta):=\ell(f_{\theta}(X_i),Y_i)\geq 0$. We define the empirical risk:
$$ r(\theta) = \frac{1}{n}\sum_{i=1}^n \ell_i(\theta) . $$
Note that it satisfies
$$\mathbb{E}_{(X_1,Y_1),\dots,(X_n,Y_n)}[r(\theta)]=R(\theta).$$
The notation for the previous expectation is cumbersome. From now, we will write $\mathcal{S}=[(X_1,Y_1),\dots,(X_n,Y_n)]$ and  $\mathbb{E}_{\mathcal{S}}$ (for ``expectation with respect to the sample'') instead of $\mathbb{E}_{(X_1,Y_1),\dots,(X_n,Y_n)}$. In the same way, we will write $\mathbb{P}_\mathcal{S}$ for probabilities with respect to the sample.

\enlargethispage{\baselineskip}
Finally, an estimator is a function that takes a sample of pairs object-labels of any size and returns a guess for the parameter $\theta$ we should use for future predictions. Formally,\footnote{The proper definition also requires $\hat{\theta}$ to be a measurable function of the observations, so that probabilities of events involving $\hat{\theta}$ are well defined. This is not so important here as we will soon replace the notion of estimator with a new notion.}
 $$
 \hat{\theta}: \bigcup_{n=1}^\infty (\mathcal{X}\times\mathcal{Y})^n \rightarrow \Theta .$$
For short, we write $\hat{\theta}$ instead of $\hat{\theta}((X_1,Y_1),\dots,(X_n,Y_n))$. The most famous example is the Empirical Risk Minimizer, or ERM:
 $$
\hat{\theta}_{{\rm ERM}} = \argmin_{\theta\in\Theta} r(\theta)
$$
(when this minimizer exists and is unique).

\subsection{PAC bounds}

Of course, our objective is to minimize $R$, not $r$. So the ERM strategy is motivated by the hope that these two functions are not so different, so that the minimizer of $r$ almost minimizes $R$. Let us now discuss to what extent this is true. By doing so, we will introduce some tools that will be also useful for PAC-Bayes bounds.
\begin{proposition}
\label{prop:with:hoeffding}
For any $\theta\in\Theta$, for any $\delta\in(0,1)$,
 \begin{equation}
\label{equa:appli:hoeffding:3}
 \mathbb{P}_\mathcal{S}\left(R(\theta) > r(\theta) + C \sqrt{\frac{\log\frac{1}{\delta}}{2n}} \right) \leq \delta.
\end{equation}
\end{proposition}
The proof relies on a result that will be useful in all this tutorial.
\begin{lemma}[Hoeffding's inequality]
 \label{lemma:hoeffding}
 Let $U_1,\dots,U_n$ be independent random variables taking values in an interval $[a,b]$. Then, for any $t>0$,
 $$ \mathbb{E} \left[ {\rm e}^{t \sum_{i=1}^n [ U_i - \mathbb{E}(U_i)]} \right] \leq {\rm e}^{\frac{n t^2 (b-a)^2}{8}}. $$
\end{lemma}
Hoeffding's inequality is proven for example in Chapter 2 of \citet{bou2013}, which is a highly recommended reading, but it is so classical that you can as well find it on Wikipedia.
\begin{proof}[Proof of Proposition~\ref{prop:with:hoeffding}]
Apply Lemma~\ref{lemma:hoeffding} to $U_i=\mathbb{E}[\ell_i(\theta)]-\ell_i(\theta)$:
 \begin{equation}
 \label{equa:appli:hoeffding}
 \mathbb{E}_\mathcal{S} \left[ {\rm e}^{ t n [R(\theta)-r(\theta)] } \right] \leq {\rm e}^{\frac{n t^2 C^2}{ 8}}.
 \end{equation}
Now, for any $s>0$,
\begin{align*}
\mathbb{P}_\mathcal{S}(R(\theta)-r(\theta) > s)
& =  \mathbb{P}_\mathcal{S}\left( {\rm e}^{ nt  [R(\theta)-r(\theta)]} > {\rm e}^{ nt s} \right)
\\
& \leq \frac{\mathbb{E}_\mathcal{S}\left[ {\rm e}^{ nt [R(\theta)-r(\theta)]}  \right]}{{\rm e}^{nt s}} \text{ by Markov's inequality,}
\\
& \leq {\rm e}^{\frac{n t^2 C^2}{ 8} - nt s} \text{ by~\eqref{equa:appli:hoeffding}}.
\end{align*}
In other words,
\begin{equation*}
 \mathbb{P}_\mathcal{S}(R(\theta) > r(\theta) +  s) \leq {\rm e}^{\frac{n t^2 C^2}{ 8} - nt s}.
\end{equation*}
We can make this bound as tight as possible, by optimizing our choice for $t$. Indeed, $ n t^2 C^2 / 8 - nt s $ is minimized for $t=4s/C^2$, which gives
\begin{equation}
\label{equa:appli:hoeffding:2}
 \mathbb{P}_\mathcal{S}(R(\theta) > r(\theta) +  s) \leq {\rm e}^{\frac{-2 n s^2}{C^2}}.
\end{equation}
This means that, for a given $\theta$, the empirical risk $r(\theta)$ cannot be much smaller than the risk $R(\theta)$. The order of this ``much smaller'' can be better understood by introducing
$$ \delta = {\rm e}^{\frac{-2 n s^2}{C^2}} $$
and substituting $\delta$ to $s$ in~\eqref{equa:appli:hoeffding:2}, which gives~\eqref{equa:appli:hoeffding:3}.
\end{proof}

Proposition~\ref{prop:with:hoeffding} states that $R(\theta)$ will usually not exceed $r(\theta)$ by more than a term in $1/\sqrt{n}$. This is not enough, though, to justify the use of the ERM. Indeed,~\eqref{equa:appli:hoeffding:3} is only true for the $\theta$ that was fixed above, and we cannot apply it to $\hat{\theta}_{{\rm ERM}}$ that is a function of the data.

The usual approach to control $R(\hat{\theta}_{{\rm ERM}})$ is to use the inequality
\begin{equation}
\label{equa:appli:hoeffding:4}
R(\hat{\theta}_{{\rm ERM}}) - r(\hat{\theta}_{{\rm ERM}}) \leq \sup_{\theta\in\Theta} \left[R(\theta)-r(\theta) \right],
\end{equation}
and to prove a version of~\eqref{equa:appli:hoeffding:3} that would hold uniformly on $\Theta$. As an illustration of this method, we prove the following result.
\begin{theorem}
\label{thm:erm}
Assume that ${\rm card}(\Theta)=M<+\infty$. For any $\delta\in(0,1)$,
\begin{equation*}
 \mathbb{P}_\mathcal{S}\left( R(\hat{\theta}_{{\rm ERM}}) \leq \inf_{\theta\in\Theta} r(\theta) +  C \sqrt{\frac{\log\frac{M}{\delta}}{2n}} \right) \geq 1-\delta.
\end{equation*}
\end{theorem}
\begin{proof}
As announced before the statement of the theorem, we upper bound the supremum in~\eqref{equa:appli:hoeffding:4}:
\begin{align}
 \mathbb{P}_\mathcal{S}( \sup_{\theta\in\Theta} [R(\theta) - r(\theta)] >  s)
 & =  \mathbb{P}_\mathcal{S}\left( \bigcup_{\theta\in\Theta} \Bigl\{ [R(\theta) - r(\theta)] >  s \Bigr\} \right)
 \nonumber
 \\
 & \leq \sum_{\theta\in\Theta}  \mathbb{P}_\mathcal{S}(R(\theta) > r(\theta) +  s)
 \nonumber
 \\
 & \leq M {\rm e}^{\frac{-2 n s^2}{C^2}}
 \label{equa:appli:hoeffding:5}
\end{align}
thanks to~\eqref{equa:appli:hoeffding:2}. This time, put
$
\delta = M {\rm e}^{\frac{-2 n s^2}{C^2}}
$
and plug into~\eqref{equa:appli:hoeffding:5} to get:
\begin{equation*}
 \mathbb{P}_\mathcal{S}\left(\sup_{\theta\in\Theta} [R(\theta) - r(\theta)] > C \sqrt{\frac{\log\frac{M}{\delta}}{2n}} \right) \leq \delta.
\end{equation*}
Thus, the complementary event satisfies
\begin{equation}
\label{equa:appli:hoeffding:6}
 \mathbb{P}_\mathcal{S}\left(\sup_{\theta\in\Theta} [R(\theta) - r(\theta)] \leq C \sqrt{\frac{\log\frac{M}{\delta}}{2n}} \right) \geq 1-\delta.
\end{equation}
From~\eqref{equa:appli:hoeffding:4},
\begin{equation*}
 \mathbb{P}_\mathcal{S}\left( R(\hat{\theta}_{{\rm ERM}}) \leq r(\hat{\theta}_{{\rm ERM}}) + C \sqrt{\frac{\log\frac{M}{\delta}}{2n}} \right) \geq 1-\delta
\end{equation*}
and note that, as $\Theta$ is finite, $r(\hat{\theta}_{{\rm ERM}}) = \inf_{\theta\in\Theta} r(\theta) $.
\end{proof}

Bounds in the form of~Theorem~\ref{thm:erm} are called Probably Approximately Correct (PAC) bounds, because $r(\hat{\theta}_{{\rm ERM}})$ \textit{approximates} $ R(\hat{\theta}_{{\rm ERM}})$ within $ C \sqrt{ \log(M/\delta)/2n }$ with \textit{probability} $1-\delta$. This terminology was introduced by \citet{val1984}.

\enlargethispage{-2\baselineskip}
\begin{remark}
The proofs of Proposition~\ref{prop:with:hoeffding} and Theorem~\ref{thm:erm} used, in addition to Hoeffding's inequality, two tricks that we will reuse very often when we will prove PAC-Bayes bounds:
\begin{itemize}
\item given a random variable $U$ and $s\in\mathbb{R}$, for any $t>0$,
$$
 \mathbb{P}\left( U > s \right)
 = 
 \mathbb{P}\left( {\rm e}^{tU} > {\rm e}^{ts} \right)
 \leq \frac{\mathbb{E}\left( {\rm e}^{tU} \right) }{{\rm e}^{ts}}
$$
thanks to Markov inequality. The combo ``exponential + Markov inequality'' is known as \textbf{Chernoff's bounding technique}. It is is of course very useful together with exponential inequalities like Hoeffding's inequality.
\item given a finite number of random variables $U_1,\dots,U_M$,
\begin{align*}
\mathbb{P}\left(\sup_{1\leq i \leq M} U_i > s \right)
&
= \mathbb{P}\left(\bigcup_{1\leq i \leq M} \Bigl\{ U_i > s \Bigr\} \right)
\\
& \leq \sum_{i=1}^M \mathbb{P}\left( U_i > s \right).
\end{align*}
This argument is called the \textbf{union-bound argument}.
\end{itemize}
\end{remark}

The next step in the study of the ERM would be to go beyond finite sets $\Theta$. The union bound argument has to be modified in this case, and things become a little more complicated. We will therefore stop here the study of the ERM: it is not our objective anyway.

If the reader is interested by the study of the ERM in general: \citet{vapnik1968uniform} developed the theoretical tools for this study, see the more recent monograph by \citet{vap1998}. We refer the reader to \citet{dev1996} for a beautiful and very pedagogical introduction to machine learning theory. Chapters 11 and 12 in particular are dedicated to Vapnik and Chervonenkis theory. More recent references include~\citet{giraud2021introduction,wainwright2019high}.

\section{What are PAC-Bayes Bounds?}

We are now in better position to explain what are PAC-Bayes bounds. A simple way to phrase things: PAC-Bayes bounds are generalization of the union bound argument, that will allow to deal with any parameter set $\Theta$: finite or infinite, continuous... However, a byproduct of this technique is that we will have to change the notion of estimator.

\enlargethispage{-3\baselineskip}
\begin{definition}
\label{dfn:measure:estimator}
Let $\mathcal{P}(\Theta)$ be the set of all probability distributions on $\Theta$ equipped with a $\sigma$-algebra $\mathcal{T}$.
 A data-dependent probability measure is a function:
 $$ \hat{\rho}: \bigcup_{n=1}^\infty (\mathcal{X}\times\mathcal{Y})^n \rightarrow \mathcal{P}(\Theta) $$
 with a suitable measurability condition.\footnote{I don't want to scare the reader with measurability conditions, as I will not check them in this tutorial anyway. Here, the exact condition to ensure that what follows is well defined is that for any $A\in\mathcal{T}$, the function
 $$ ((x_1,y_1),\dots,(x_n,y_n)) \mapsto \left[\hat{\rho}((x_1,y_1),\dots,(x_n,y_n))\right](A) $$
 is measurable. That is, $\hat{\rho}$ is a regular conditional probability.} We will write $\hat{\rho}$ instead of $\hat{\rho}((X_1,Y_1),\dots,(X_n,Y_n))$ for short.
\end{definition}

In practice, when you have a data-dependent probability measure, and you want to build a predictor, you can:
\begin{itemize}
 \item draw a random parameter $\tilde{\theta} \sim \hat{\rho}$, we will call this procedure ``randomized estimator''.
 \item use it to average predictors, that is, define a new predictor:
  $$ f_{\hat{\rho}}(\cdot) = \mathbb{E}_{\theta\sim\hat{\rho}} [f_\theta(\cdot)] $$
  called the aggregated predictor with weights $\hat{\rho}$.
\end{itemize}

So, with PAC-Bayes bounds, we will extend the union bound argument\footnote{See the title of van Erven's tutorial~\citep{van2014}: ``PAC-Bayes mini-tutorial: a continuous union bound''. Note, however, that it is argued by \citet{cat2007} that PAC-Bayes bounds are actually more than that, we will come back to this in Section~\ref{section:oracle}.} to infinite, uncountable sets $\Theta$, but we will obtain bounds on various risks related to data-dependent probability measures, that is:
\begin{itemize}
 \item the risk of a randomized estimator, $R(\tilde{\theta})$,
 \item or the average risk of randomized estimators, $\mathbb{E}_{\theta\sim\hat{\rho}} [R(\theta)]$,
 \item or the risk of the aggregated estimator, $R(f_{\hat{\rho}})$.
\end{itemize}

From a technical point of view, the analysis shares many similarities with the analysis of the ERM in the previous section. A key difference is that the supremum in~\eqref{equa:appli:hoeffding:4} will be replaced by
\begin{equation*}
\mathbb{E}_{\theta\sim\hat{\rho}} [R(\theta)-r(\theta)] \leq \sup_{\rho\in\mathcal{P}(\Theta)} \mathbb{E}_{\theta\sim\rho} [R(\theta)-r(\theta)] .
\end{equation*}
While this might look unnecessarily complicated at first sight, PAC-Bayes bounds will actually turn out to be extremely convenient for many reasons that we hope will become clear along the next sections:
\begin{itemize}
 \item first, they don't require the set of predictors to be finite, nor discrete. Of course, it is also possible to prove PAC bounds for the ERM when $\Theta \subset \mathbb{R}^p $ is not finite, but this leads to technical difficulties or strong restrictions such as the compactness of $\Theta$. PAC-Bayes bounds do not lead to major difficulties with unbounded parameter spaces, as will be illustrated in Example~\ref{example:gaussian:unbounded}.
 \item randomized estimators are fairly common in machine learning. This includes Bayesian estimation and related methods such as variational inference and ensemble methods. Section~\ref{section:first:step} illustrates how PAC-Bayes bounds can be applied to such estimators. Moreover, many non-randomized estimators can be derived from randomized ones: aggregation rules, majority vote classifiers, etc. The PAC-Bayes bounds on the randomized estimator often brings strong information on the de-randomized version. This will also be discussed thoroughly and illustrated in Section~\ref{section:first:step}.
 \item Bayesian estimators incorporate prior knowledge through a prior distribution $\pi$ on $\Theta$. Even though PAC-Bayes bounds can be applied to non-Bayesian estimators, a prior $\pi$ will still appear in the bound. The effect of $\pi$ on the bound will be discussed thoroughly. In particular, PAC-Bayes bounds depend not only on the minimum of the empirical risk $r(\theta)$, but on the prior probability of the level sets of $r$: in general, this can be quantified through the so-called prior-mass condition, as described in Section~\ref{section:oracle}, even though specific examples such as Example~\ref{exm:first:finite} will already illustrate this property.
 A consequence is that flater minima lead to tigther bounds. This is one of the reasons why PAC-Bayes bounds can be tight for deep learning (Section~\ref{section:tight}).
\end{itemize}
You will of course ask the question: if $\Theta$ is infinite, what will the $\log(M)$ term be replaced with? In PAC-Bayes bounds, this term will be replaced by the Kullback-Leibler divergence between $\rho$ and a fixed $\pi$ on $\Theta$ (the prior).

\enlargethispage{-2\baselineskip}
\begin{definition}
 Given two probability measures $\mu$ and $\nu$ in $\mathcal{P}(\Theta)$, the Kullback-Leibler (or simply KL) divergence between $\mu$ and $\nu$ is
 
 \noindent
 $$
 KL(\mu\|\nu) = \int \log\left(\frac{{\rm d}\mu}{{\rm d}\nu}(\theta) \right) \mu({\rm d}\theta) \in [0,+\infty]
 $$
 if $\mu$ has a density $\frac{{\rm d}\mu}{{\rm d}\nu}$ with respect to $\nu$, and $KL(\mu\|\nu)=+\infty$ otherwise.\footnote{We recall that if there is a measurable function $g$ such that for any measurable set $A$,
$$ \mu(A) = \int_A g(\theta) \nu({\rm d}\theta), $$
then this function is essentially unique. We put $\frac{{\rm d}\mu}{{\rm d}\nu}(\theta) = g(\theta) $ and refer to this function as the density of $\mu$ with respect to $\nu$.}
\end{definition}

\begin{example}
For example, if $\Theta$ is finite,
$$ KL(\mu\|\nu)= \sum_{\theta\in\Theta} \log\left(\frac{\mu(\theta)}{\nu(\theta)}\right) \mu(\theta). $$
\end{example}

The following result is well known. You can prove it using Jensen's inequality.
\begin{proposition}
\label{prop:kullback}
 For any probability measures $\mu$ and $\nu$, $KL(\mu\|\nu)\geq 0$ with equality if and only if $\mu=\nu$.
\end{proposition}

\section{Why this Tutorial?}

Since the ``PAC analysis of a Bayesian estimator'' by \citet{sha1997} and the first PAC-Bayes bounds proven by \citet{mca1998,mca1999}, many new PAC-Bayes bounds appeared (we will see that some of them can be derived from a bound due to \cite{see2002}). These bounds were used in various contexts, to solve a wide range of problems. This led to hundreds of (beautiful!) papers. The consequence of this is that it's quite difficult to be aware of all the existing work on PAC-Bayes bounds. In particular, it seems that many powerful techniques in Catoni's book~\citep{cat2007} and earlier works (Catoni, \citeyear{cat2003}; \citeyear{cat2004}) are largely ignored by the community.

On the other hand, it's not easy to enter into the PAC-Bayes literature. Most papers already assume some basic knowledge on these bounds, and Catoni's book is quite technical to begin with. The objective of these notes is thus to provide a user-friendly introduction, accessible to PhD students, that could be used as a first approach to PAC-Bayes bounds. It also provides references for more sophisticated results.

I want to mention existing short introduction to PAC-Bayes bounds, like the ones by~\citet{mca2013,van2014} and the nice introductory slides of \citet{fleuret2011}. They are very informative, and I recommend the reader to check them. However, they are focused on empirical bounds only. There are also surveys on PAC-Bayes bounds, such as~\citet[Section 5]{cho2015} or \citet{gue2019}. These papers are very useful to navigate in the ocean of publications on PAC-Bayes bounds, and they helped me a lot when I was writing this document, but might not provide enough detail for a first reading on the topic.

Finally, in order to highlight the main ideas, I will not necessarily try to present the bounds with the tightest possible constants. In particular, many oracle bounds and localized bounds in Section~\ref{section:oracle} were introduced in Catoni (\citeyear{cat2003}; \citeyear{cat2007}) with better constants. Once again, this is an \textit{introduction} to PAC-Bayes bounds. I strongly recommend the reader to check the original publications for more accurate results.

\section{Two Types of PAC Bounds, Organization of these Notes}
\label{subsec:twotypes}

It is important to make a distinction between two types of PAC bounds.

Theorem~\ref{thm:erm} is usually refered to as an {\it empirical bound}. It means that, for any $\theta$, $R(\theta)$ is upper bounded by an empirical quantity, that is, by something that we can compute when we observe the data. This allows to study the ERM as the minimizer of this bound. It also provides a numerical certificate of the generalization error of the ERM. You will really end up with something like
$$
 \mathbb{P}_\mathcal{S}\left( R(\hat{\theta}_{{\rm ERM}}) \leq 0.12 \right) \geq 0.99.
$$

However, a numerical certificate on the generalization error does not tell you one thing. Can this $0.12$ be improved using a larger sample size? Or is it the best that can be done with our set of predictors? The right tools to answer these questions are excess risk bounds, also known as oracle PAC bounds. In these bounds, you have a control of the form

\noindent
$$
 \mathbb{P}_\mathcal{S}\left( R(\hat{\theta}_{{\rm ERM}}) \leq \inf_{\theta\in\Theta}R(\theta) + r_n(\delta) \right) \geq 1-\delta,
$$
where the remainder $r_n(\delta)$ should be as small as possible and satisfy $r_n(\delta)\rightarrow 0$ when $n\rightarrow\infty$. Of course, the upper bound on $R(\hat{\theta}_{{\rm ERM}})$ cannot be computed because $R$ is unknown in practice, so it doesn't lead to a numerical certificate on $R(\hat{\theta}_{{\rm ERM}})$. Still, these bounds are very interesting, because they tell you how close you can expect $R(\hat{\theta}_{{\rm ERM}})$ to be to the smallest possible value of $R$.

In the same way, there are empirical PAC-Bayes bounds, and oracle PAC-Bayes bounds (also known as excess-risk PAC-Bayes bounds). The very first PAC-Bayes bounds by \citet{mca1998,mca1999} were empirical bounds. The first oracle PAC-Bayes bounds came later~\citep{cat2003,cat2004,zha2006,cat2007}.

In some sense, empirical PAC-Bayes bounds are more useful in practice, and oracle PAC-Bayes bounds are theoretical objects. But this might be an oversimplification. We will see that empirical bounds are tools used to prove some oracle bounds, so they are also useful in theory. On the other hand, when we design a data-dependent probability measure, we don't know if it will lead to large or small empirical bounds. A preliminary study of its theoretical properties through an oracle bound is the best way to ensure that it is efficient, and so that it has a chance to lead to small empirical bounds.

In Section~\ref{section:first:step}, we will study an example of empirical PAC-Bayes bound, essentially taken from a preprint by \citet{cat2003}. We will prove it together, play with it and modify it in many ways. In Section~\ref{section:tight}, we cover many empirical PAC-Bayes bounds, and explain the race to tighter bounds. This led to bounds that are tight enough to provide good generalization certificates for deep learning, we will discuss this based on Dziugaite and Roy's paper~\citep{dzi2017} and a more recent work by P\'erez-Ortiz, Rivasplata,  Shawe-Taylor, and Szepesv\`ari~\citep{per2020}.

In Section~\ref{section:oracle}, we will turn to oracle PAC-Bayes bounds. We will see how to derive these bounds from empirical bounds, and apply them to some classical set of predictors. We will examine the assumptions leading to fast rates in these inequalities.
Section~\ref{section:not:bounded:iid} will be devoted to the various attempts to extend PAC-Bayes bounds beyond the setting introduced in this introduction, that is: bounded loss, and i.i.d. observations. Finally, in Section~\ref{section:related} we will discuss briefly the connection between PAC-Bayes bounds and many other approaches in machine learning and statistics, including regret bounds and  Mutual Information bounds (MI).

\chapter{First Step in the PAC-Bayes World}
\label{section:first:step}

As mentioned above, there are many PAC-Bayes bounds. This section covers as a first example a bound  proven by \citet{cat2003}. Why this choice?

Well, any choice is partly arbitrary: I did my PhD thesis~\citep{alquier2006transductive} with Olivier Catoni and thus I know his works well. But, also, the objective is not to rush immediately to {\it the best} bound. Rather, this Catoni's result will be useful to illustrate how PAC-Bayes bounds work, how to use them, and explain the different variants (bounds on randomized estimators, bounds on aggregated estimators, etc.). It appears that Catoni's technique is extremely convenient to prove almost all the various type of bounds with a unified proof. Later, in Section~\ref{section:tight}, we will see many alternative empirical PAC-Bayes bounds, this will allow you to compare them, and discuss the pros and the cons of each.

\section{A Simple PAC-Bayes Bound}

\subsection{Catoni's bound~\citep{cat2003}}

From now, and until the end of these notes, {\bf let us fix a probability measure $\pi\in\mathcal{P}(\Theta)$. The measure $\pi$ will be called the prior}, because of a connection with Bayesian statistics that will be discussed in Section~\ref{section:related}. We recall that the loss function is bounded and takes values in $[0,C]$.

\begin{theorem}
\label{thm:first:bound}
For any $\lambda>0$, for any $\delta\in(0,1)$,
\begin{multline*}
 \mathbb{P}_\mathcal{S}\Biggl(\forall\rho\in\mathcal{P}(\Theta) \text{, } \mathbb{E}_{\theta\sim\rho}[ R(\theta)] \leq \mathbb{E}_{\theta\sim\rho}[ r(\theta)] 
 \\
 +  \frac{\lambda C^2}{8n} + \frac{KL(\rho\|\pi) + \log\frac{1}{\delta}}{\lambda} \Biggr)
 \geq 1-\delta.
\end{multline*}
\end{theorem}
Let us prove Theorem~\ref{thm:first:bound}. This requires a lemma that will be extremely useful in all these notes. This lemma has been known since the 50s \citep[Exercise 8.28]{kul1959} in the case of a finite $\Theta$, but the general case is due to \citet{don1975}.
\begin{lemma}[Donsker and Varadhan's variational formula]
\label{lemma:dv}
For any measurable, bounded function $h:\Theta\rightarrow\mathbb{R}$ we have:
\begin{equation*}
\log \mathbb{E}_{\theta\sim\pi}\left[{\rm e}^{h(\theta)} \right] =\sup_{\rho\in\mathcal{P}(\Theta)}\Bigl[\mathbb{E}_{\theta\sim\rho}[h(\theta)] -KL (\rho\|\pi)\Bigr].
\end{equation*}
Moreover, the supremum with respect to $\rho$ in the right-hand side is
reached for the Gibbs measure
$\pi_{h}$ defined by its density with respect to $\pi$:
\begin{equation}
\label{equa:def:gibbs:measure}
\frac{{\rm d}\pi_{h}}{{\rm d}\pi}(\theta) =  \frac{{\rm e}^{h(\theta)}}
{ \mathbb{E}_{\vartheta\sim\pi}\left[{\rm e}^{h(\vartheta)} \right] }.
\end{equation}
\end{lemma}
\noindent {\it Proof of Lemma~\ref{lemma:dv}.}
Using the definition, for any $\rho\in\mathcal{P}(\Theta)$,
\begin{align*}
 KL(\rho\|\pi_{h})
 &
 = \mathbb{E}_{\theta\sim\rho} \left[ \log\left( \frac{{\rm d}\rho}{{\rm d}\pi_{h}}(\theta) \right) \right]
 \\
 & = \mathbb{E}_{\theta\sim\rho} \left[ \log\left(  \frac
{ \mathbb{E}_{\vartheta\sim\pi}\left[{\rm e}^{h(\vartheta)} \right] }{{\rm e}^{h(\theta)}} \frac{{\rm d}\rho}{{\rm d}\pi}(\theta)  \right) \right]
 \\
 & = -\mathbb{E}_{\theta\sim\rho}[h(\theta)] + \mathbb{E}_{\theta\sim\rho} \left[ \log\left( \frac{{\rm d}\rho}{{\rm d}\pi}(\theta) \right) \right] + \log \mathbb{E}_{\vartheta\sim\pi}\left[{\rm e}^{h(\vartheta)} \right]
 \\
 & = -\mathbb{E}_{\theta\sim\rho}[h(\theta)] + KL(\rho\|\pi) + \log \mathbb{E}_{\theta\sim\pi}\left[{\rm e}^{h(\theta)} \right].
\end{align*}
Thanks to Proposition~\ref{prop:kullback}, the left hand side is nonnegative, and equal to $0$ only for $\rho=\pi_h$.
$\square$

\noindent {\it Proof of Theorem~\ref{thm:first:bound}.}
The beginning of the proof follows closely the study of the ERM and the proof of Theorem~\ref{thm:erm}. Fix $\theta\in\Theta$ and apply Hoeffding's inequality with $U_i=\mathbb{E}[\ell_i(\theta)]-\ell_i(\theta)$: for any $t>0$,
 \begin{equation*}
 \mathbb{E}_\mathcal{S} \left[ {\rm e}^{ t n [R(\theta)-r(\theta)] } \right] \leq {\rm e}^{\frac{n t^2 C^2}{ 8}}.
 \end{equation*}
We put $t=\lambda/n$, which gives:
 \begin{equation*}
 \mathbb{E}_\mathcal{S} \left[ {\rm e}^{ \lambda [R(\theta)-r(\theta)] } \right] \leq {\rm e}^{\frac{\lambda^2 C^2}{ 8 n}}.
 \end{equation*}
This is where the proof diverges from the proof of Theorem~\ref{thm:erm}. We will now integrate this bound with respect to $\pi$:
 \begin{equation*}
 \mathbb{E}_{\theta\sim\pi} \mathbb{E}_\mathcal{S} \left[ {\rm e}^{ \lambda [R(\theta)-r(\theta)] } \right] \leq {\rm e}^{\frac{\lambda^2 C^2}{ 8 n}}.
 \end{equation*}
 Thanks to Tonelli's theorem, we can exchange the integration with respect to $\theta$ and the one with respect to the sample:
 \begin{equation}
 \mathbb{E}_\mathcal{S}  \mathbb{E}_{\theta\sim\pi} \left[ {\rm e}^{ \lambda [R(\theta)-r(\theta)] } \right] \leq {\rm e}^{\frac{\lambda^2 C^2}{ 8 n}}
  \label{equa:proof:catoni:0}
 \end{equation}
 and we apply Donsker and Varadhan's variational formula (Lemma~\ref{lemma:dv}) to get:
 \begin{equation*}
 \mathbb{E}_\mathcal{S} \left[ {\rm e}^{ \sup_{\rho\in\mathcal{P}(\Theta)} \lambda \mathbb{E}_{\theta\sim\rho}[R(\theta)-r(\theta)] -KL(\rho\|\pi) } \right] \leq {\rm e}^{\frac{\lambda^2 C^2}{ 8 n}}.
 \end{equation*} 
 Rearranging terms:
 \begin{equation}
 \mathbb{E}_\mathcal{S} \left[ {\rm e}^{ \sup_{\rho\in\mathcal{P}(\Theta)} \lambda \mathbb{E}_{\theta\sim\rho}[R(\theta)-r(\theta)] -KL(\rho\|\pi)-\frac{\lambda^2 C^2}{ 8 n} } \right] \leq 1.
 \label{equa:proof:catoni}
 \end{equation}
 The end of the proof uses Chernoff bound. Fix $s>0$,
 \begin{align*}
  \mathbb{P}_\mathcal{S} & \left[  \sup_{\rho\in\mathcal{P}(\Theta)} \lambda \mathbb{E}_{\theta\sim\rho}[R(\theta)-r(\theta)] -KL(\rho\|\pi)-\frac{\lambda^2 C^2}{ 8 n}  > s \right] 
  \\
  & \leq  \mathbb{E}_\mathcal{S} \left[ {\rm e}^{ \sup_{\rho\in\mathcal{P}(\Theta)} \lambda \mathbb{E}_{\theta\sim\rho}[R(\theta)-r(\theta)] -KL(\rho\|\pi)-\frac{\lambda^2 C^2}{ 8 n} } \right] {\rm e}^{-s}
  \\
  & \leq {\rm e}^{-s}.
 \end{align*}
 Solve ${\rm e}^{-s}=\delta$, that is, put $s=\log(1/\delta)$ to get
 \begin{equation*}
 \mathbb{P}_\mathcal{S} \left[ \sup_{\rho\in\mathcal{P}(\Theta)} \lambda \mathbb{E}_{\theta\sim\rho}[R(\theta)-r(\theta)] -KL(\rho\|\pi)-\frac{\lambda^2 C^2}{ 8 n}  > \log\frac{1}{\delta} \right] \leq \delta.
 \end{equation*} 
 Rearranging terms gives:
 \begin{multline*}
 \mathbb{P}_\mathcal{S} \Biggl[ \exists \rho\in\mathcal{P}(\Theta) \text{, } \mathbb{E}_{\theta\sim\rho}[ R(\theta)] > \mathbb{E}_{\theta\sim\rho}[ r(\theta)]
 \\
 +  \frac{\lambda C^2}{8n} + \frac{KL(\rho\|\pi) + \log\frac{1}{\delta}}{\lambda} \Biggr]
 \leq \delta.
 \end{multline*}
 Take the complement to end the proof. $\square$

\subsection{Exact minimization of the bound}

We remind that the bound in Theorem~\ref{thm:erm},
\begin{equation*}
 \mathbb{P}_\mathcal{S}\left(\forall\theta\in\Theta\text{, } R(\theta) \leq r(\theta) +  C \sqrt{\frac{\log\frac{M}{\delta}}{2n}} \right) \geq 1-\delta,
\end{equation*}
motivated the introduction of $\hat{\theta}_{{\rm ERM}}$, the minimizer of $r$.

Exactly in the same way, the bound in Theorem~\ref{thm:first:bound},
\begin{multline*}
 \mathbb{P}_\mathcal{S}\Biggl(\forall\rho\in\mathcal{P}(\Theta) \text{, } \mathbb{E}_{\theta\sim\rho}[ R(\theta)] \leq \mathbb{E}_{\theta\sim\rho}[ r(\theta)]
 \\
 +  \frac{\lambda C^2}{8n} + \frac{KL(\rho\|\pi) + \log\frac{1}{\delta}}{\lambda} \Biggr) \geq 1-\delta,
\end{multline*}
motivates the study of a data-dependent probability measure $\hat{\rho}_\lambda$ that would be defined as:
$$
\hat{\rho}_\lambda = \argmin_{\rho\in\mathcal{P}(\Theta)} \left\{ \mathbb{E}_{\theta\sim\rho}[ r(\theta)]  + \frac{KL(\rho\|\pi) }{\lambda} \right\}.
$$
But does such a minimizer exist? It turns out that the answer is yes, thanks to Donsker and Varadhan's variational formula again! Indeed, to minimize:
$$
 \mathbb{E}_{\theta\sim\rho}[ r(\theta)]  + \frac{KL(\rho\|\pi) }{\lambda} 
$$
is equivalent to maximize
$$
-\lambda \mathbb{E}_{\theta\sim\rho}[ r(\theta)] - KL(\rho\|\pi)
$$
which is exactly what the variational inequality does, with $h(\theta)=-\lambda r(\theta)$. We know that the minimum is reached for $\rho=\pi_{-\lambda r}$ as defined in~\eqref{equa:def:gibbs:measure}. Let us summarize this in following definition and corollary.
\begin{definition}
In the whole tutorial we will let $\hat{\rho}_\lambda$ denote ``the Gibbs posterior'' given by $\hat{\rho_\lambda} = \pi_{-\lambda r}$, that is:
\begin{equation}
\label{equa:def:gibbs:posterior}
\hat{\rho}_\lambda({\rm d}\theta) =  \frac{{\rm e}^{-\lambda r(\theta)} \pi({\rm d}\theta)}
{ \mathbb{E}_{\vartheta\sim\pi}\left[{\rm e}^{-\lambda r(\vartheta)} \right] } .
\end{equation}
\end{definition}
\begin{corollary}
\label{cor:first:bound}
The Gibbs posterior is the minimizer of the right-hand side of Theorem~\ref{thm:first:bound}:
$$
\hat{\rho}_\lambda = \argmin_{\rho\in\mathcal{P}(\Theta)} \left\{ \mathbb{E}_{\theta\sim\rho}[ r(\theta)]  + \frac{KL(\rho\|\pi) }{\lambda} \right\}.
$$
As a consequence, for any $\lambda>0$, for any $\delta\in(0,1)$,
\begin{multline*}
 \mathbb{P}_\mathcal{S}\left( \mathbb{E}_{\theta\sim\hat{\rho}_\lambda}[ R(\theta)] \leq \inf_{\rho\in\mathcal{P}(\Theta)}\left[  \mathbb{E}_{\theta\sim\rho}[ r(\theta)] +  \frac{\lambda C^2}{8n} + \frac{KL(\rho\|\pi) + \log\frac{1}{\delta}}{\lambda} \right]\right)
 \\
 \geq 1-\delta.
\end{multline*}
\end{corollary}

\subsection{Some examples, and non-exact minimization of the bound}
\label{subsec:non:exact}

When you see something like:
$$
\mathbb{E}_{\theta\sim\rho}[ r(\theta)] +  \frac{\lambda C^2}{8n} + \frac{KL(\rho\|\pi) + \log\frac{1}{\delta}}{\lambda},
$$
I'm not sure you immediately see what is the order of magnitude of the bound. I don't. In general, when you apply such a general bound to a set of predictors, I think it is quite important to make the bound more explicit. Let us process a few examples (I advise you to do the calculations on your own in these examples, and in other examples).

\begin{example}[Finite case]
\label{exm:first:finite}
Let us start with the special case where $\Theta$ is a finite set, that is, ${\rm card}(\Theta)=M<+\infty$. We begin with the application of Corollary~\ref{cor:first:bound}. In this case, the Gibbs posterior $\hat{\rho}_\lambda$ of~\eqref{equa:def:gibbs:posterior} is a probability on the finite set $\Theta$ given by
$$
\hat{\rho}_\lambda(\theta) = \frac{{\rm e}^{-\lambda r(\theta)}\pi(\theta) }{\sum_{\vartheta\in\Theta} {\rm e}^{-\lambda r(\vartheta)}\pi(\vartheta)}.
$$
and we have, with probability at least $1-\delta$:
\begin{equation}
\label{equa:exm:thm:first:bound:discr:1}
 \mathbb{E}_{\theta\sim\hat{\rho}_\lambda}[ R(\theta)] \leq \inf_{\rho\in\mathcal{P}(\Theta)}\left[  \mathbb{E}_{\theta\sim\rho}[ r(\theta)] +  \frac{\lambda C^2}{8n} + \frac{KL(\rho\|\pi) + \log\frac{1}{\delta}}{\lambda} \right].
\end{equation}
When we apply Donsker and Varadhan's variational formula (again!) to the right-hand side, we obtain:
\begin{equation}
\label{equa:exm:thm:first:bound:discr:1:bis}
 \mathbb{E}_{\theta\sim\hat{\rho}_\lambda}[ R(\theta)] \leq \frac{-1}{\lambda} \log \sum_{\theta\in\Theta} \pi(\theta) {\rm e}^{- \lambda r(\theta)}   +  \frac{\lambda C^2}{8n} + \frac{\log \frac{1}{\delta}}{\lambda}.
\end{equation}
By using, for any $\theta\in\Theta$, $0 \leq \exp(-\lambda r(\theta)) \leq \exp(-\lambda \inf_{\vartheta\in\Theta} r(\vartheta)) $, we show:
\begin{equation}
 \label{equa:soft:max}
 \inf_{\theta\in\Theta} r(\theta) \leq
 \frac{-1}{\lambda} \log \sum_{\theta\in\Theta} \pi(\theta){\rm e}^{- \lambda r(\theta)}  \leq \inf_{\vartheta\in\Theta} \left[ r(\theta) + \frac{\log \frac{1}{\pi(\theta)}}{\lambda}\right].
\end{equation}
So,~\eqref{equa:exm:thm:first:bound:discr:1:bis} leads to the more explicit (but less tight!) bound:
\begin{equation*}
 \mathbb{E}_{\theta\sim\hat{\rho}_\lambda}[ R(\theta)] \leq \inf_{\theta\in\Theta} \left[ r(\theta) +  \frac{\lambda C^2}{8n} + \frac{ \log\frac{1}{\pi(\theta) \delta}}{\lambda} \right] .
\end{equation*}
This gives us an intuition on the role of the prior $\pi$: the bound will be tight if there is a $\theta$ such that $r(\theta)$ and $1/\pi(\theta)$ are small simulataneously. However, $\pi$ cannot be large everywhere: its total mass is constrained to be $1$. The larger $\Theta$ is, the more we have to ``spread the mass'' of $\pi$, which will increase $1/\pi(\theta)$ in the bound.
This is clear if we use a uniform prior: $\pi(\theta)=1/M$. In this case, the choice $\lambda=\textcolor{red}{\sqrt{8n\log(M/\delta)/C^2}}$ actually minimizes the right-hand side, and leads to:
\begin{equation}
\label{bound:example:finite}
 \mathbb{P}_\mathcal{S}\left( \mathbb{E}_{\theta\sim\hat{\rho}_\lambda}[ R(\theta)]  \leq \inf_{\theta\in\Theta}  r(\theta) +  C\sqrt{\frac{\log\frac{M}{\delta}}{2n}} \right) \geq 1-\delta.
\end{equation}
That is, the Gibbs posterior $\hat{\rho}_\lambda$ satisfies the same bound as the ERM in Theorem~\ref{thm:erm}. Note that it does not mean that $\hat{\rho}_\lambda$ and $\hat{\theta}_{{\rm ERM}}$ are equivalent! The PAC-Bayes bound in~\eqref{equa:exm:thm:first:bound:discr:1:bis} can actually be tighter, as shown by a closer examination of~\eqref{equa:soft:max}: the term in the middle can be arbitrarily close to the left-hand side if the empirical risk of all parameters is close enough to $\inf_{\theta\in\Theta} r(\theta)$. We will discuss in Section~\ref{section:oracle} a so-called prior mass condition that relates more generally the mass of the level sets of $r$ under $\pi$ to the tightness of PAC-Bayes bounds.

Note that the optimization with respect to $\lambda$ is a little more problematic when $\pi$ is not uniform, because the optimal $\lambda$ would depend on the data, which is not allowed - at the moment. We will come back to the choice of $\lambda$ in the general case soon.

\enlargethispage{-\baselineskip}
Let us also consider the statement of Theorem~\ref{thm:first:bound} in this case. With probability at least $1-\delta$, we have:
\begin{equation*}
 \forall\rho\in\mathcal{P}(\Theta)\text{, } \mathbb{E}_{\theta\sim\rho}[ R(\theta)] \leq \mathbb{E}_{\theta\sim\rho}[ r(\theta)] +  \frac{\lambda C^2}{8n} + \frac{KL(\rho\|\pi) + \log\frac{1}{\delta}}{\lambda}.
\end{equation*}
As the bound holds for all $\rho\in\mathcal{P}(\Theta)$, it holds in particular for all $\rho$ in the set of Dirac masses $\{\delta_\theta,\theta\in\Theta\}$. Obviously:
$$ \mathbb{E}_{\vartheta\sim\delta_\theta}[ r(\vartheta)] = r(\theta) $$
and
$$
KL(\delta_\theta\|\pi)= \sum_{\vartheta\in\Theta} \log\left(\frac{\delta_{\theta}(\vartheta) }{\pi(\vartheta)}\right) \delta_{\theta}(\vartheta) = \log \frac{1}{\pi(\theta)}.
$$
This gives:
$$
\forall \theta\in\Theta\text{, }
 R(\theta) \leq r(\theta) +  \frac{\lambda C^2}{8n} + \frac{\log\frac{1}{\pi(\theta)} + \log\frac{1}{\delta}}{\lambda}
 $$
 and, when $\pi$ is uniform:
$$
\forall \theta\in\Theta\text{, }
 R(\theta) \leq r(\theta) +  \frac{\lambda C^2}{8n} + \frac{\log\frac{M}{\delta}}{\lambda}.
 $$
 As this bound holds for any $\theta$, it holds in particular for the ERM, which gives:
$$
 R(\hat{\theta}_{{\rm ERM}}) \leq \inf_{\theta\in\Theta} r(\theta) +  \frac{\lambda C^2}{8n} + \frac{\log\frac{M}{\delta}}{\lambda}
 $$
and, once again with the choice $\lambda=\textcolor{red}{\sqrt{8n\log(M/\delta)/C^2}}$, we recover exactly the result of Theorem~\ref{thm:erm}:
$$
 R(\hat{\theta}_{{\rm ERM}}) \leq \inf_{\theta\in\Theta} r(\theta) +  C\sqrt{\frac{\log\frac{M}{\delta}}{2n}}.
 $$
\end{example}
The previous example leads to important remarks:
\begin{itemize}
 \item PAC-Bayes bounds can be used to prove generalization bounds for Gibbs posteriors, but sometimes they can also be used to study more classical estimators, like the ERM. This actually goes far beyond the ERM. Many recent papers use PAC-Bayes bounds to study non-Bayesian robust estimators of the mean and the covariance matrix of heavy-tailed random vectors. This is discussed further in Sections~\ref{subset:ERM:final} and~\ref{subset:ERM:final:bis}.
 \item the choice of $\lambda$ has a different status when you study the Gibbs posterior $\hat{\rho}_\lambda$ and the ERM. Indeed, in the bound on the ERM, $\lambda$ is chosen to minimize the bound, but the estimation procedure is not affected by $\lambda$. The bound for the Gibbs posterior is also minimized with respect to $\lambda$, but $\hat{\rho}_\lambda$ depends on $\lambda$. So, if you make a mistake when choosing $\lambda$, this will have bad consequences not only on the bound, but also on the practical performances of the method. This means also that if the bound is not tight, it is likely that the $\lambda$ obtained by minimizing the bound will not lead to good performances in practice. (We present in Section~\ref{section:tight} bounds that do not depend on a parameter like $\lambda$).
\end{itemize}

\begin{example}[Lipschitz loss and Gaussian priors]
\label{exm:lipschitz:gauss}
 Let us switch to the continuous case, so that we can derive from PAC-Bayes bounds some results that could not be obtained only with a union bound argument. We consider the case where $\Theta=\mathbb{R}^d$, the function $\theta\mapsto\ell(f_\theta(x),y)$ is $L$-Lipschitz for any $x$ and $y$, and the prior $\pi$ is a centered Gaussian: $\pi=\mathcal{N}(0,\sigma^2 I_d)$ where $I_d$ is the $d\times d$ identity matrix.
 
 Let us, as in the previous example, study first the Gibbs posterior, by an application of Corollary~\ref{cor:first:bound}. With probability at least $1-\delta$,
 $$
 \mathbb{E}_{\theta\sim\hat{\rho}_\lambda}[ R(\theta)] \leq \inf_{\rho\in\mathcal{P}(\Theta)}\left[  \mathbb{E}_{\theta\sim\rho}[ r(\theta)] +  \frac{\lambda C^2}{8n} + \frac{KL(\rho\|\pi) + \log\frac{1}{\delta}}{\lambda} \right].
$$
Once again, the right-hand side is an infimum over all possible probability distributions $\rho$, but it is easier to restrict to Gaussian distributions here. So:

\noindent
\begin{equation}
\label{equa:exm:thm:first:bound:gauss}
 \mathbb{E}_{\theta\sim\hat{\rho}_\lambda}[ R(\theta)] \leq \inf_{
 \begin{tiny}
 \begin{array}{c}
 \rho=\mathcal{N}(m,s^2 I_d)
 \\
 m\in\mathbb{R}^d, s>0
 \end{array}
 \end{tiny}
 }\left[  \mathbb{E}_{\theta\sim\rho}[ r(\theta)] +  \frac{\lambda C^2}{8n} + \frac{KL(\rho\|\pi) + \log\frac{1}{\delta}}{\lambda} \right].
\end{equation}
Indeed, it is well known that, for $\rho=\mathcal{N}(m,s^2 I_d)$ and $\pi=\mathcal{N}(0,\sigma^2 I_d)$,
$$
KL(\rho\|\pi) = \frac{\|m\|^2}{2\sigma^2} + \frac{d}{2}\left[\frac{s^2}{\sigma^2} + \log\frac{\sigma^2}{s^2} -1 \right].
$$
Moreover, the risk $r$ inherits the Lipschitz property of the loss, that is, for any $(\theta,\vartheta)\in\Theta^2$, $r(\theta)\leq r(\vartheta) + L\|\vartheta-\theta\|$. So, for $\rho = \mathcal{N}(m,s^2 I_d)$,
\begin{align*}
 \mathbb{E}_{\theta\sim\rho}[ r(\theta)]
 & \leq r(m) + L \mathbb{E}_{\theta\sim\rho}[\|\theta-m\| ]
 \\
 & \leq r(m) + L \sqrt{\mathbb{E}_{\theta\sim\rho}[ \|\theta-m\|^2 ]} \text{ by Jensen's inequality}
 \\
 & =  r(m) + L s \sqrt{d}.
\end{align*}
Plugging this into~\eqref{equa:exm:thm:first:bound:gauss} gives:
\begin{multline*}
 \mathbb{E}_{\theta\sim\hat{\rho}_\lambda}[ R(\theta)]
 \leq \inf_{
 m\in\mathbb{R}^d, s>0
 }\Biggl[  r(m) + L s \sqrt{d} +  \frac{\lambda C^2}{8n} 
 \\
 + \frac{ \frac{\|m\|^2}{2\sigma^2} + \frac{d}{2}\left[\frac{s^2}{\sigma^2} + \log\frac{\sigma^2}{s^2} -1 \right] + \log\frac{1}{\delta}}{\lambda} \Biggr].
\end{multline*} \enlargethispage{\baselineskip}
It is possible to minimize the bound completely in $s$, but for now, we will just consider the choice $s=\sigma/\sqrt{n}$, which gives:
\begin{multline*}
 \mathbb{E}_{\theta\sim\hat{\rho}_\lambda}[ R(\theta)]
 \\
 \leq \inf_{
 m\in\mathbb{R}^d
 }\left[  r(m) + L \sigma\sqrt{\frac{d}{n}} +  \frac{\lambda C^2}{8n} + \frac{ \frac{\|m\|^2}{2\sigma^2} + \frac{d}{2}\left[\frac{1}{n}-1 + \log(n)\right] + \log\frac{1}{\delta}}{\lambda} \right]
 \\
 \leq \inf_{
 m\in\mathbb{R}^d
 }\left[  r(m) + L \sigma\sqrt{\frac{d}{n}} +  \frac{\lambda C^2}{8n} + \frac{ \frac{\|m\|^2}{2\sigma^2} + \frac{d}{2}\log(n) + \log\frac{1}{\delta}}{\lambda} \right].
\end{multline*}
It is not possible to optimize the bound with respect to $\lambda$ as the optimal value would depend on $m$... We will show how to get rid of this limitation soon. However, there is a way to simplify the bound (by making it worse!): to restrict the infimum on $m$ to $\|m\|\leq B$ for some $B>0$. Then we have:

\noindent
$$
 \mathbb{E}_{\theta\sim\hat{\rho}_\lambda}[ R(\theta)]
 \leq \inf_{
 m:\|m\|\leq B
 }  r(m) + L \sigma\sqrt{\frac{d}{n}} +  \frac{\lambda C^2}{8n} + \frac{ \frac{B^2}{2\sigma^2} + \frac{d}{2}\log(n) + \log\frac{1}{\delta}}{\lambda}.
$$
In this case, we see that the optimal $\lambda$ is
$$ \lambda = \frac{1}{C}\sqrt{8n\left(\frac{B^2}{2\sigma^2} + \frac{d}{2}\log(n) + \log\frac{1}{\delta}\right)} $$
which gives:
$$
 \mathbb{E}_{\theta\sim\hat{\rho}_\lambda}[ R(\theta)]
 \leq \inf_{m:\|m\|\leq B }  r(m) + L \sigma\sqrt{\frac{d}{n}} + C \sqrt{ \frac{\frac{B^2}{2\sigma^2} + \frac{d}{2}\log(n) + \log\frac{1}{\delta}}{2n} }.
$$
Note that our choice of $\lambda$ might look a bit weird, as it depends on the confidence level $\delta$. This can be avoided by taking:
$$ \lambda = \frac{1}{C}\sqrt{8n\left(\frac{B^2}{2\sigma^2} + \frac{d}{2}\log(n) \right)} $$
instead (check what bound you obtain by doing so!).

Finally, as in the previous example, we can also start from the statement of Theorem~\ref{thm:first:bound}: with probability at least $1-\delta$,
\begin{equation*}
 \forall\rho\in\mathcal{P}(\Theta)\text{, } \mathbb{E}_{\theta\sim\rho}[ R(\theta)] \leq \mathbb{E}_{\theta\sim\rho}[ r(\theta)] +  \frac{\lambda C^2}{8n} + \frac{KL(\rho\|\pi) + \log\frac{1}{\delta}}{\lambda},
\end{equation*}
and restrict here $\rho$ to the set of Gaussian distributions $\mathcal{N}(m,s^2 I_d)$. This leads to the definition of a new data-dependent probability measure, $\tilde{\rho}_\lambda=\mathcal{N}(\tilde{m},\tilde{s}^2 I_d)$ where
\begin{multline}
(\tilde{m},\tilde{s})
= \argmin_{m\in\mathbb{R}^d,s>0} \Biggl\{
\mathbb{E}_{\theta\sim\mathcal{N}(m,s^2 I_d)}[r(\theta)] 
\\
+  \frac{\lambda C^2}{8n} + \frac{\frac{\|m\|^2}{2\sigma^2} + \frac{d}{2}\left[\frac{s^2}{\sigma^2} + \log\frac{\sigma^2}{s^2} -1 \right] + \log\frac{1}{\delta}}{\lambda} \Biggr\}
\label{equa:argmin:notreached}
\end{multline}
when this is well defined (we can adapt the definition of  $\tilde{\rho}_\lambda$ to the situation when there is no minimum, we postpone this to the next paragraph).
While the Gibbs posterior $\hat{\rho}_\lambda$ can be quite a complicated object, one simply has to solve this minimization problem to get $\tilde{\rho}_\lambda$. The probability $\tilde{\rho}_\lambda$ is actually a special case of what is called a variational approximation of $\hat{\rho}_\lambda$. Variational approximations are very popular in statistics and machine learning, and were indeed analyzed through PAC-Bayes bounds~\citep{alq2016,alq2020,yan2020}. We will come back to this in Section~\ref{section:related}. For now, following the same computations, and using the same choice of $\lambda$ as for $\hat{\rho}_\lambda$, we obtain the same bound:
$$
 \mathbb{E}_{\theta\sim\tilde{\rho}_\lambda}[ R(\theta)]
 \leq \inf_{m:\|m\|\leq B }  r(m) + L \sigma\sqrt{\frac{d}{n}} + C \sqrt{ \frac{\frac{B^2}{2\sigma^2} + \frac{d}{2}\log(n) + \log\frac{1}{\delta}}{2n} }.
$$

Note that this approach can still be used in the case where the minimum is not reached in the definition of $(\tilde{m},\tilde{s})$ in~\eqref{equa:argmin:notreached}. For any $\epsilon>0$, we define $(\tilde{m}_\epsilon,\tilde{s}_\epsilon)$ as an $\epsilon$-minimizer, that is, it reaches the infimum in~\eqref{equa:argmin:notreached} up to $\epsilon$.
Taking $\tilde{\rho}_\lambda$ as $\tilde{\rho}_\lambda=\mathcal{N}(\tilde{m}_{\epsilon},\tilde{s}^2_{\epsilon} I_d)$ with $\epsilon = 1/n$ leads to
\begin{multline*}
 \mathbb{E}_{\theta\sim\tilde{\rho}_\lambda}[ R(\theta)]
 \\
 \leq \inf_{m:\|m\|\leq B }  r(m) + L \sigma\sqrt{\frac{d}{n}} + C \sqrt{ \frac{\frac{B^2}{2\sigma^2} + \frac{d}{2}\log(n) + \log\frac{1}{\delta}}{2n} } + \frac{1}{n} .
\end{multline*}

\end{example}

\enlargethispage{-2\baselineskip}
\begin{example}[Model aggregation, model selection]
\label{exm:model:selection}
In the case where we have many sets of predictors, say $\Theta_1,\dots,\Theta_M$, equipped with priors $\pi_1,\dots,\pi_M$ respectively, it is possible to define a prior on $\Theta= \bigcup_{j=1}^M \Theta_j$. For the sake of simplicity, assume that the $\Theta_j$'s are disjoint, and let $p=(p(1),\dots,p(M))$ be a probability distribution on $\{1,\dots,M\}$. We simply put:
$$ \pi = \sum_{j=1}^{M}p(j) \pi_j . $$
The minimization of the bound in Theorem~\ref{thm:first:bound} leads to the Gibbs posterior $\hat{\rho}_\lambda$ that will put mass on all the $\Theta_j$ in general, so this is a model aggregation procedure in the spirit of
\citet{mca1999}. On the other hand, we can also restrict the minimization in the PAC-Bayes bound to distributions that would charge only one of the models, that is, to
$\rho \in \mathcal{P}(\Theta_1)\cup \dots \cup \mathcal{P}(\Theta_M)$. Theorem~\ref{thm:first:bound} becomes:

\noindent
\begin{multline*}
\mathbb{P}_{\mathcal{S}}\Biggl( \forall j\in\{1,\dots,M\},\forall \rho  \in \mathcal{P}(\Theta_j)\text{, }
\mathbb{E}_{\theta\sim\rho}[ R(\theta)]
\\
\leq \mathbb{E}_{\theta\sim\rho}[ r(\theta)] +  \frac{\lambda C^2}{8n} + \frac{KL(\rho\|\pi) + \log\frac{1}{\delta}}{\lambda} \Biggr) \geq 1-\delta,
\end{multline*}
that is
\begin{multline*}
\mathbb{P}_{\mathcal{S}}\Biggl( \forall j\in\{1,\dots,M\},\forall \rho  \in \mathcal{P}(\Theta_j)\text{, }
\mathbb{E}_{\theta\sim\rho}[ R(\theta)]
\\
\leq \mathbb{E}_{\theta\sim\rho}[ r(\theta)] +  \frac{\lambda C^2}{8n} + \frac{KL(\rho\|\pi_j) + \log\frac{1}{p(j)} + \log\frac{1}{\delta}}{\lambda} \Biggr) \geq 1-\delta.
\end{multline*}
Thus, we can propose the following procedure:
\begin{itemize}
 \item first, we build the Gibbs posterior for each model $j$,
$$
\hat{\rho}_\lambda^{(j)}({\rm d}\theta)
=  \frac{{\rm e}^{-\lambda r(\theta)}\pi_j( {\rm d}\theta) }{ \int_{\Theta_j} {\rm e}^{-\lambda r(\vartheta)}\pi_j({\rm d}\vartheta)},
$$
\item then, model selection:
$$ \hat{j} = \argmin_{1\leq j \leq M}  \left\{ \mathbb{E}_{\theta\sim\hat{\rho}_\lambda^{(j)}}[ r(\theta)] + \frac{KL(\hat{\rho}_\lambda^{(j)}\|\pi_j) + \log\frac{1}{p(j)} }{\lambda}  \right\}. $$
\end{itemize}
The obtained $\hat{j}$ satisfies:
\begin{multline*}
\mathbb{P}_{\mathcal{S}}\Biggl(
\mathbb{E}_{\theta\sim\hat{\rho}_{\lambda}^{(\hat{j})}}[ R(\theta)]
\leq \min_{1\leq j\leq M } \inf_{\rho\in\mathcal{P}(\Theta_j)}\Biggl\{ \mathbb{E}_{\theta\sim\rho}[ r(\theta)]
\\
+ \frac{KL(\rho\|\pi_j) + \log\frac{1}{p(j)} + \log\frac{1}{\delta}}{\lambda} \Biggr\} \Biggr) \geq 1-\delta.
\end{multline*}
\end{example}

\subsection{The choice of $\lambda$}
\label{subsec:choice:lambda}

As discussed earlier, it is in general not possible to optimize the right-hand side of the PAC-Bayes equality with respect to $\lambda$. For example, in~\eqref{equa:exm:thm:first:bound:discr:1}, the optimal value of $\lambda$ could depend on $\rho$, which is not allowed by Theorem~\ref{thm:first:bound}. In the previous examples, we have seen that in some situations, if one is lucky enough, the optimal $\lambda$ actually does not depend on $\rho$, but we still need a procedure for the general case.

A natural idea is to propose a finite grid $\Lambda\subset (0,+\infty)$ and to minimize over this grid, which can be justified by a union bound argument.
\begin{theorem}
\label{thm:first:bound:lambda:grid}
Let $\Lambda\subset (0,+\infty)$ be a finite set. For any $\delta\in(0,1)$,
\begin{multline*}
 \mathbb{P}_\mathcal{S}\Biggl(\forall\rho\in\mathcal{P}(\Theta),\forall \lambda\in\Lambda \text{, }
 \mathbb{E}_{\theta\sim\rho}[ R(\theta)]
 \\
 \leq \mathbb{E}_{\theta\sim\rho}[ r(\theta)] +  \frac{\lambda C^2}{8n} + \frac{KL(\rho\|\pi) + \log\frac{{\rm card}(\Lambda)}{\delta}}{\lambda}\Biggr) \geq 1-\delta.
\end{multline*}
\end{theorem}

\noindent {\it Proof.} Fix $\lambda\in\Lambda$, and then follow the proof of Theorem~\ref{thm:first:bound} until~\eqref{equa:proof:catoni}:
$$
 \mathbb{E}_\mathcal{S} \left[ {\rm e}^{ \sup_{\rho\in\mathcal{P}(\Theta)} \lambda \mathbb{E}_{\theta\sim\rho}[R(\theta)-r(\theta)] -KL(\rho\|\pi)-\frac{\lambda^2 C^2}{ 8 n} } \right] \leq 1.
$$
Sum over $\lambda\in\Lambda$ to get:
$$
 \sum_{\lambda\in\Lambda} \mathbb{E}_\mathcal{S} \left[ {\rm e}^{ \sup_{\rho\in\mathcal{P}(\Theta)} \lambda \mathbb{E}_{\theta\sim\rho}[R(\theta)-r(\theta)] -KL(\rho\|\pi)-\frac{\lambda^2 C^2}{ 8 n} } \right] \leq {\rm card}(\Lambda)
$$
and so
$$
 \mathbb{E}_\mathcal{S} \left[ {\rm e}^{ \sup_{\rho\in\mathcal{P}(\Theta),\lambda\in\Lambda} \lambda \mathbb{E}_{\theta\sim\rho}[R(\theta)-r(\theta)] -KL(\rho\|\pi)-\frac{\lambda^2 C^2}{ 8 n} } \right] \leq {\rm card}(\Lambda).
$$
The end of the proof is as for Theorem~\ref{thm:first:bound}, we start with Chernoff bound. Fix $s>0$,
 \begin{align*}
  \mathbb{P}_\mathcal{S} & \left[  \sup_{\rho\in\mathcal{P}(\Theta),\lambda\in\Lambda} \lambda \mathbb{E}_{\theta\sim\rho}[R(\theta)-r(\theta)] -KL(\rho\|\pi)-\frac{\lambda^2 C^2}{ 8 n}  > s \right] 
  \\
  & \leq  \mathbb{E}_\mathcal{S} \left[ {\rm e}^{ \sup_{\rho\in\mathcal{P}(\Theta),\lambda\in\Lambda} \lambda \mathbb{E}_{\theta\sim\rho}[R(\theta)-r(\theta)] -KL(\rho\|\pi)-\frac{\lambda^2 C^2}{ 8 n} } \right] {\rm e}^{-s}
  \\
  & \leq {\rm card}(\Lambda) {\rm e}^{-s}.
 \end{align*}
 Solve ${\rm card}(\Lambda){\rm e}^{-s}=\delta$, that is, put $s=\log({\rm card}(\Lambda)/\delta)$ to get
 \small
 \begin{equation*}
 \mathbb{P}_\mathcal{S} \left[ \sup_{\rho\in\mathcal{P}(\Theta)} \lambda \mathbb{E}_{\theta\sim\rho}[R(\theta)-r(\theta)] -KL(\rho\|\pi)-\frac{\lambda^2 C^2}{ 8 n}  > \log\frac{{\rm card}(\Lambda)}{\delta} \right] \leq \delta.
 \end{equation*} \normalsize
 Rearranging terms gives:
 \begin{multline*}
 \mathbb{P}_\mathcal{S} \Biggl[ \exists \rho\in\mathcal{P}(\Theta),\exists \lambda\in\Lambda \text{, } \mathbb{E}_{\theta\sim\rho}[ R(\theta)]
> \mathbb{E}_{\theta\sim\rho}[ r(\theta)]
\\
+  \frac{\lambda C^2}{8n} + \frac{KL(\rho\|\pi) + \log\frac{{\rm card}(\Lambda)}{\delta}}{\lambda} \Biggr] \leq \delta.
 \end{multline*}
 Take the complement to get the statement of the theorem. $\square$
 
This leads to the following procedure. First, we remind that, for a fixed $\lambda$, the minimizer of the bound is $\hat{\rho}_\lambda = \pi_{-\lambda r}$. Then, we put:
\begin{align}
\nonumber
\hat{\rho} & = \hat{\rho}_{\hat{\lambda}} \text{ where }
\\
\hat{\lambda} & =\argmin_{\lambda\in\Lambda} \left\{
\mathbb{E}_{\theta\sim \pi_{-\lambda r}}[ r(\theta)] +  \frac{\lambda C^2}{8n} + \frac{KL(\pi_{-\lambda r}\|\pi) + \log\frac{{\rm card}(\Lambda)}{\delta}}{\lambda}
\right\}.
\label{equa:first:bound:lambda:grid}
\end{align}
We have immediately the following result.
\begin{corollary}
\label{cor:first:bound:lambda:grid}
Define $\hat{\rho}$ as in~\eqref{equa:first:bound:lambda:grid}, for any $\delta\in(0,1)$ we have
\begin{multline*}
 \mathbb{P}_\mathcal{S}\Biggr( \mathbb{E}_{\theta\sim\hat{\rho}}[ R(\theta)]
 \leq \inf_{
\begin{tiny}
 \begin{array}{c}
  \rho\in\mathcal{P}(\Theta)
  \\
  \lambda\in\Lambda
 \end{array}
\end{tiny}
 }\Biggl[  \mathbb{E}_{\theta\sim\rho}[ r(\theta)] +  \frac{\lambda C^2}{8n}
 \\
 + \frac{KL(\rho\|\pi) + \log\frac{{\rm card}(\Lambda)}{\delta}}{\lambda} \Biggr]\Biggr) \geq 1-\delta.
\end{multline*}
\end{corollary}

We could for example propose an arithmetic grid $ \Lambda = \{1,2,\dots,n \}$. The bound in Corollary~\ref{cor:first:bound:lambda:grid} becomes:
$$
\mathbb{E}_{\theta\sim\hat{\rho}}[ R(\theta)]
 \\
 \leq \inf_{
\begin{tiny}
 \begin{array}{c}
  \rho\in\mathcal{P}(\Theta)
  \\
  \lambda=1,\dots,n
 \end{array}
\end{tiny}
 }\left[  \mathbb{E}_{\theta\sim\rho}[ r(\theta)] +  \frac{\lambda C^2}{8n} + \frac{KL(\rho\|\pi) + \log\frac{n}{\delta}}{\lambda} \right]
$$
It is also possible to transform the optimization on a discrete grid by an optimization on a continuous grid. Indeed, for any $\lambda\in[1,n]$, we simply apply the bound to the integer part of $\lambda$, $\lfloor \lambda \rfloor$, and remark that we can upper bound $\lfloor \lambda \rfloor \leq \lambda$ and $1/\lfloor \lambda \rfloor \leq 1/(\lambda-1)$. So the bound becomes:
$$
\mathbb{E}_{\theta\sim\hat{\rho}}[ R(\theta)]
 \\
 \leq \inf_{
\begin{tiny}
 \begin{array}{c}
  \rho\in\mathcal{P}(\Theta)
  \\
  \lambda\in[1,n]
 \end{array}
\end{tiny}
 }\left[  \mathbb{E}_{\theta\sim\rho}[ r(\theta)] +  \frac{\lambda C^2}{8n} + \frac{KL(\rho\|\pi) + \log\frac{n}{\delta}}{\lambda-1} \right].
$$

The arithmetic grid is not be the best choice, though: the $\log(n)$ term can be improved. In order to optimize hyperparameters in PAC-Bayes bounds, \citet{langford2002not} used a geometric grid $\Lambda=\{{\rm e}^k,k\in\mathbb{N} \}\cap[1,n]$, the same choice was used later by \citet{cat2003} and \citet[Theorem 1.2.8 page 13]{cat2007}. Using such a grid in Corollary~\ref{cor:first:bound:lambda:grid} we get
$$
\mathbb{E}_{\theta\sim\hat{\rho}}[ R(\theta)]
 \\
 \leq \inf_{
\begin{tiny}
 \begin{array}{c}
  \rho\in\mathcal{P}(\Theta)
  \\
  \lambda\in[1,n]
 \end{array}
\end{tiny}
 }\left[  \mathbb{E}_{\theta\sim\rho}[ r(\theta)] +  \frac{\lambda C^2}{8n} + \frac{KL(\rho\|\pi) + \log\frac{\log n}{\delta}}{\lambda/{\rm e}} \right].
$$

We conclude this discussion on the choice of $\lambda$ by mentioning that there are other PAC-Bayes bounds, as the one by \citet{mca1999}, where there is no parameter $\lambda$ to optimize. We will study these bounds in Section~\ref{section:tight}.

\let\mysectionmark\sectionmark
\renewcommand\sectionmark[1]{}
\section[PAC-Bayes Bound on Aggregation of Predictors and \\ Weighted Majority Vote]{PAC-Bayes Bound on Aggregation of Predictors and Weighted Majority Vote}
\let\sectionmark\mysectionmark
\sectionmark{PAC-Bayes Bound on Aggregation of Predictors and Weighted Majority Vote}

In the introduction, right after Definition~\ref{dfn:measure:estimator}, I promised that PAC-Bayes bound would allow to control
\begin{itemize}
 \item the risk of randomized predictors,
 \item the expected risk of randomized predictors,
 \item the risk of averaged predictors.
\end{itemize}
But so far, we only focused on the expected risk of randomized predictors (the second bullet point). In this subsection, we provide some bounds on averaged predictors, and in the next one, we will focus on the risk of randomized predictors.

We start by a very simple remark. When the loss function
$u\mapsto \ell(u,y)$ is convex for any $y$, then the risk $R(\theta)=R(f_\theta)$ is a convex function of $f_\theta$. Thus, Jensen's inequality ensures:
$$
\mathbb{E}_{\theta\sim\rho}[R(f_{\theta})]
\geq R( \mathbb{E}_{\theta\sim\rho}[f_{\theta}]).
$$
Plugging this into Corollary~\ref{cor:first:bound} gives immediately the following result.

\begin{corollary}
 \label{cor:first:bound:convex}
 Assume that $\forall y\in\mathcal{Y}$, $u\mapsto \ell(u,y)$ is convex. Define
 $$ \hat{f}_{\hat{\rho}_\lambda}(\cdot) = \mathbb{E}_{\theta\sim\hat{\rho}_\lambda}[f_\theta(\cdot)]. $$
 For any $\lambda>0$, for any $\delta\in(0,1)$,
 \small
\begin{equation*}
 \mathbb{P}_\mathcal{S}\left( R( \hat{f}_{\hat{\rho}_\lambda}) \leq \inf_{\rho\in\mathcal{P}(\Theta)} \left[ \mathbb{E}_{\theta\sim\rho}[ r(\theta)] +  \frac{\lambda C^2}{8n} + \frac{KL(\rho\|\pi) + \log\frac{1}{\delta}}{\lambda} \right] \right) \geq 1-\delta.
\end{equation*}\normalsize
\end{corollary}
That is, in the case of a convex loss function, like the quadratic loss or the hinge loss, PAC-Bayes bounds also provide bounds on the risk of aggregated predictors.

It is also possible to study $R ( \mathbb{E}_{\theta\sim\rho}[f_{\theta}]) -\mathbb{E}_{\theta\sim\rho}[R(f_{\theta})]$ under other assumptions. For example, we can use the Lipschitz property as in Example~\ref{exm:lipschitz:gauss}.
In the case of the quadratic loss, we have $R( \mathbb{E}_{\theta\sim\rho}[f_{\theta}])-\mathbb{E}_{\theta\sim\rho}[R(f_{\theta})]
 = \mathbb{E}_{X} [ {\rm Var}_{\theta\sim\rho} ( f_\theta(X)) ]$, this fact was used to provide tight bounds on $R( \mathbb{E}_{\theta\sim\rho}[f_{\theta}])$ \citep[page 22]{audPHD}. A similar idea was used later beyond regression to control $R( \mathbb{E}_{\theta\sim\rho}[f_{\theta}])-\mathbb{E}_{\theta\sim\rho}[R(f_{\theta})]$ by a variance term \citep{germain2015risk,mas2019}.

However, note that in the case of classification, $f_\theta(x)\in\{0,1\}$ for any $x$ and $\theta$, but $ \hat{f}_{\hat{\rho}_\lambda}(x)$ can be any value in $[0,1]$. We can define a new classifier by $\hat{f}_{\hat{\rho}_\lambda}^{{\rm maj}}(x) = 1$ if $ \hat{f}_{\hat{\rho}_\lambda}(x)\geq 1/2$ and $\hat{f}_{\hat{\rho}_\lambda}^{{\rm maj}}(x) = 0$ otherwise. This is known as the weighted majority vote classifier. Adaptations of PAC-Bayes bounds to control the risk of weighted majority vote classifiers and variants were proven, and this is still an important research direction~\citep{lan2002,lacasse2006pac,laviolette2011pac,lorenzen2019pac,masegosa2020second,wu2021cheby,wu2022split}.

\section{PAC-Bayes Bound on a Single Draw from the Posterior}

\begin{theorem}
\label{thm:pac:bayes:bound:random}
 For any $\lambda>0$, for any $\delta\in(0,1)$, for any data-dependent probability measure $\tilde{\rho}$,
\begin{equation*}
 \mathbb{P}_\mathcal{S} \mathbb{P}_{\tilde{\theta}\sim\tilde{\rho} } \left( R( \tilde{\theta} ) \leq  r(\tilde{\theta}) +  \frac{\lambda C^2}{8n} + \frac{\log \frac{{\rm d}\rho}{{\rm d}\pi}(\tilde{\theta}) + \log\frac{1}{\delta}}{\lambda} \right) \geq 1-\delta.
\end{equation*} 
\end{theorem}
This bound simply says that if you draw $\tilde{\theta}$ from, for example, the Gibbs posterior $\hat{\rho}_\lambda$ (defined in~\eqref{equa:def:gibbs:posterior}), you have the bound on $R( \tilde{\theta} )$ that holds with large probability simultaneously on the drawing of the sample and of $\tilde{\theta}$.

\noindent {\it Proof.}
Once again, we follow the proof of Theorem~\ref{thm:first:bound}, until~\eqref{equa:proof:catoni:0}:
 \begin{equation*}
 \mathbb{E}_\mathcal{S}  \mathbb{E}_{\theta\sim\pi} \left[ {\rm e}^{ \lambda [R(\theta)-r(\theta)] } \right] \leq {\rm e}^{\frac{\lambda^2 C^2}{ 8 n}}.
 \end{equation*}
 Now, for any nonnegative function $h$,
\begin{align*}
\mathbb{E}_{\theta\sim\pi}[h(\theta)]
& =
\int h(\theta) \pi({\rm d}\theta)
\\
& \geq  \int_{  \left\{ \frac{{\rm d}\tilde{\rho}}{{\rm d}\pi}(\theta) >0 \right\} } h(\theta) \pi({\rm d}\theta)
\\
& = \int_{  \left\{ \frac{{\rm d}\tilde{\rho}}{{\rm d}\pi}(\theta) >0 \right\} } h(\theta) \frac{{\rm d}\pi}{{\rm d}\tilde{\rho}}(\theta)  \tilde{\rho}({\rm d}\theta)
\\
& = \mathbb{E}_{\theta\sim\tilde{\rho}}\left[ h(\theta) {\rm e}^{-\log\frac{{\rm d}\tilde{\rho}}{{\rm d}\pi}(\theta) }  \right]
\end{align*}
and in particular:
 \begin{equation*}
 \mathbb{E}_\mathcal{S}  \mathbb{E}_{\theta\sim\tilde{\rho}} \left[ {\rm e}^{ \lambda [R(\theta)-r(\theta)] -\log\frac{{\rm d}\tilde{\rho}}{{\rm d}\pi}(\theta) } \right] \leq {\rm e}^{\frac{\lambda^2 C^2}{ 8 n}}.
 \end{equation*}
We could go through the proof until the end, but you can now guess that it's essentially Chernoff bound + rearrangement of the terms. $\square$

\citet{clerico2022pacdet} proposed a very clever use of such a PAC-Bayes bound that does not actually require to sample from $\tilde{\rho}$. They study a sequence $(\theta_t)_{t=0,\dots,T}$ obtained by gradient descent, with a random initialization $\theta_0$. The prior $\pi$ is taken as the distribution of $\theta_0$, and $\tilde{\rho}$ is defined implicitly as the distribution of $\theta_T$. They apply a PAC-Bayes bound similar to Theorem~\ref{thm:pac:bayes:bound:random} and provide an explicit upper bound on $\log \frac{{\rm d}\rho}{{\rm d}\pi}$. This analysis gives a bound on $R(\theta_T)$, that is, a bound on the risk of a deterministic predictor (up to the initialization). This is related to information bounds on similar algorithms that will be discussed in Subsection~\ref{subsubsec:MI}.

\section{Bound in Expectation}

We end this section by one more variant of the initial PAC-Bayes bound in Theorem~\ref{thm:first:bound}: a bound in expectation with respect to the sample.

\begin{theorem}
\label{thm:first:bound:exp}
For any $\lambda>0$, for any data-dependent probability measure $\tilde{\rho}$,
\begin{equation*}
 \mathbb{E}_\mathcal{S} \mathbb{E}_{\theta\sim\tilde{\rho}}[ R(\theta)] \leq  \mathbb{E}_\mathcal{S} \mathbb{E}_{\theta\sim\tilde{\rho}}\left[ r(\theta) +  \frac{\lambda C^2}{8n} + \frac{KL(\tilde{\rho}\|\pi) }{\lambda} \right].
\end{equation*}
In particular, for $\tilde{\rho} = \hat{\rho}_\lambda$ the Gibbs posterior,
\begin{equation*}
 \mathbb{E}_\mathcal{S} \mathbb{E}_{\theta\sim\hat{\rho}_\lambda}[ R(\theta)] \leq  \mathbb{E}_\mathcal{S} \left[ \inf_{\rho\in\mathcal{P}(\theta)} \mathbb{E}_{\theta\sim\rho}[ r(\theta)] +  \frac{\lambda C^2}{8n} + \frac{KL(\rho\|\pi) }{\lambda} \right].
\end{equation*}
\end{theorem}
These bounds in expectation are very convenient tools from a pedagogical point of view. Indeed, in Section~\ref{section:oracle}, we will study oracle PAC-Bayes inequalities. While it is possible to derive oracle PAC-Bayes bounds both in expectation and with large probability, the one in expectation are much simpler to derive, and much shorter. Thus, Section~\ref{section:oracle} will mostly cover PAC-Bayes oracle bounds in expectation, and we refer the reader to \citet{cat2003,cat2007} for the corresponding bounds in probability.

Note that as the bound does not hold with large probability, as the previous bounds, it is no longer a PAC bound in the proper sense: {\it Probably Approximately Correct}. A few years ago, I was attending a talk by Tsybakov where he presented some results from his paper with Dalalyan~\citep{dal2008} that can also be interpreted as a ``PAC-Bayes bound in expectation'', and he suggested the more appropriate EAC-Bayes acronym: {\it Expectedly Approximately Correct} (their paper is briefly discussed in Section~\ref{subsection:aggregation} below). I don't think this term was often reused since then. I also found recently the acronym MAC-Bayes~\citep{grunwald2021pac}: {\it Mean Approximately Correct}. In order to avoid any confusion I will stick to ``PAC-Bayes bound in expectation'', but I like EAC and MAC! Early examples of PAC-Bayes bounds in expectation were proven by \citet{alquier2006transductive,cat2007,juditsky2008learning,dal2008}.

\noindent {\it Proof.}
Once again, the beginning of the proof is the same as for Theorem~\ref{thm:first:bound}, until~\eqref{equa:proof:catoni}:
\begin{equation*}
 \mathbb{E}_\mathcal{S} \left[ {\rm e}^{ \sup_{\rho\in\mathcal{P}(\Theta)} \lambda \mathbb{E}_{\theta\sim\rho}[R(\theta)-r(\theta)] -KL(\rho\|\pi)-\frac{\lambda^2 C^2}{ 8 n} } \right] \leq 1.
\end{equation*}
This time, use Jensen's inequality to send the expectation with respect to the sample inside the exponential function:
\begin{equation*}
 {\rm e}^{  \mathbb{E}_\mathcal{S} \left[ \sup_{\rho\in\mathcal{P}(\Theta)} \lambda \mathbb{E}_{\theta\sim\rho}[R(\theta)-r(\theta)] -KL(\rho\|\pi)-\frac{\lambda^2 C^2}{ 8 n} \right] } \leq 1,
\end{equation*}
that is,
\begin{equation*}
 \mathbb{E}_\mathcal{S} \left[ \sup_{\rho\in\mathcal{P}(\Theta)} \lambda \mathbb{E}_{\theta\sim\rho}[R(\theta)-r(\theta)] -KL(\rho\|\pi)-\frac{\lambda^2 C^2}{ 8 n} \right]\leq 0.
\end{equation*}
In particular,
\begin{equation*}
 \mathbb{E}_\mathcal{S} \left[ \lambda \mathbb{E}_{\theta\sim\tilde{\rho}}[R(\theta)-r(\theta)] -KL(\tilde{\rho}\|\pi)-\frac{\lambda^2 C^2}{ 8 n} \right]\leq 0.
\end{equation*}
Rearrange terms. $\square$

\section{Applications of Empirical PAC-Bayes Bounds}

The original PAC-Bayes bounds were stated for classification \citep{mca1998} and it became soon clear that many results could be extended to any bounded loss, thus covering for example bounded regression (we discuss in Section~\ref{section:not:bounded:iid} how to get rid of the boundedness assumption). Thus, some papers are written in no specific setting, with a generic loss, that can cover classification, regression, or density estimation. For example, the result of \citet{cat2007} about classification are extended to unbounded losses in my PhD thesis~\citep[Chapter 1]{alquier2006transductive}.

\begin{sloppypar}
However, some empirical PAC-Bayes bounds were also developped or applied to specific models, sometimes taking advantage of some specificities of the model. We mention for example:\end{sloppypar}
\begin{itemize}
 \item support vector classifiers~\citep{lan2002,cat2007},
 \item random forests~\citep{lorenzen2019pac},
 \item ranking/scoring~\citep{ral2010},
 \item density estimation~\citep{hig2010,sel2012c},
 \item multiple testing~\citep{blanchard2007occam} is tackled with related techniques,
 \item deep learning: even though deep networks are trained for classification or regression, the application of PAC-Bayes bounds to deep learning is not straightforward. We discuss this in Section~\ref{section:tight} based on the work by \citet{dzi2017} and more recent references,
 \item generative adversarial networks~\citep{mbacke2023pac},
 \item unsupervised learning, including clustering~\citep{sel2012c,app2021},~representation learning~\citep{noz2019,noz2020}, variational autoencoders \citep{cherief2022vae}.
\end{itemize}
This list is of course non-exhaustive. Many more applications are presented in Section~\ref{section:oracle} (more precisely, in Section~\ref{subsec:appli:oracle}).

\chapter{Tight and Non-vacuous PAC-Bayes Bounds}
\label{section:tight}

\section{Why is there a Race to Tighter PAC-Bayes Bound?}

Let us start with a numerical application of the PAC-Bayes bounds we met in Section~\ref{section:first:step}.

First, assume we are in the classification setting with the 0-1 loss, so that $C=1$. We are given a small set of classifiers, say $M=100$, such that on the test set with size $n=1000$, the best of these classifiers has an empirical risk $r_n=0.26$. Let us use the bound in~\eqref{bound:example:finite}, that is reminded here:
\begin{equation*}
 \mathbb{P}_\mathcal{S}\left( \mathbb{E}_{\theta\sim\hat{\rho}_\lambda}[ R(\theta)]  \leq \inf_{\theta\in\Theta}  r(\theta) +  C\sqrt{\frac{\log\frac{M}{\delta}}{2n}} \right) \geq 1-\delta.
\end{equation*}
With $\delta=0.05$ this bound is:
\begin{equation*}
 \mathbb{P}_\mathcal{S}\Biggl( \mathbb{E}_{\theta\sim\hat{\rho}_\lambda}[ R(\theta)]  \leq 0.26 + \underbrace{1.\sqrt{\frac{\log \frac{100}{0.05} }{2\times 1000}}}_{\leq 0.062} \Biggr) \geq 0.95.
\end{equation*}
So the classification risk using the Gibbs posterior is smaller than $0.322$ with probability at least $95\%$.

Let us now switch to a more problematic example. We consider a very simple binary neural network, given by the following formula, for $x\in\mathbb{R}^d$, and where $\varphi$ is a nonlinear activation function (e.g $\varphi(x) = \max(x,0)$):
$$
f_{w}(x) = \mathbf{1}\left[\sum_{i=1}^M w^{(2)}_i \varphi\left(\sum_{j=1}^d w_{j,i}^{(1)} x_j \right) \geq 0\right]
$$
and the weights $w_{j,i}^{(1)}$ and $w^{(2)}_i$ are all in $\{-1,+1\}$ for $1\leq j\leq d$ and $1\leq i \leq M$. Define $\theta=(w_{1,1}^{(1)},w_{1,2}^{(1)},\dots,w_{d,M}^{(1)},w^{(2)}_1,\dots,w^{(2)}_M)$. Note that the set of all possible such networks has cardinality $2^{M(d+1)}$. Consider inputs that are $100\times 100$ greyscale images, that is, $x\in[0,1]^d$ with $d=10,000$, and a sample size $n=10,000$. With neural networks, it is often the case that a perfect classification of the training sample is possible, that is, there is a $\theta$ such that $r(\theta)=0$.

Even for a moderate number of units such as $M=100$, this leads to the PAC-Bayes bound (with $\delta=0.05$):
\begin{equation*}
 \mathbb{P}_\mathcal{S}\Biggl( \mathbb{E}_{\theta\sim\hat{\rho}_\lambda}[ R(\theta)]  \leq \underbrace{1.\sqrt{\frac{\log \frac{2^{1,000,100}}{0.05} }{2\times 10,000}}}_{\simeq 13.58} \Biggr) \geq 0.95.
\end{equation*}
So the classification risk using the Gibbs posterior is smaller than $13.58$ with probability at least $95\%$. Which is not informative at all, because we already know that the classification risk is smaller than $1$. Such a bound is usually refered to as a {\it vacuous bound}, because it does not bring any information at all. You can try to improve the bound by increasing the dataset. But you can check that even $n=1,000,000$ still leads to a vacuous bound with this network.

Various opinions on these vacuous bounds are possible:
\begin{itemize}
 \item ``theory is useless. I don't know why I would care about generalization guarantees, neural networks work in practice.'' This opinion is lazy: it's just a good excuse not to have to think about generalization guarantees. I will assume that since you are reading this tutorial, this is not your opinion.
 \item ``vacuous bounds are certainly better than no bounds at all!'' This opinion is a little cynical, it can be rephrased as ``better have a theory that doesn't work than no theory at all: at least we can claim we have a theory, and some people might even believe us''. But the theory just says nothing.
 \item ``let's get back to work, and improve the bounds''. Since the publication of the first PAC-Bayes bounds already mentioned~\citep{sha1997,mca1998,mca1999}, many variants were proven. One can try to test which one is the best in a given setting, try to improve the priors, try to refine the bound in many ways... In 2017, \citet{dzi2017} obtained non-vacuous (even though not really tight yet) PAC-Bayes bounds for practical neural networks (since then, tighter bounds were obtained by these authors and by others). This is a remarkable achievement, and this also made PAC-Bayes theory immediately more popular than it was ever before.
\end{itemize}

Let's begin this section with a review of some popular PAC-Bayes bounds: Section~\ref{subsec:few:pac}. Some of them are of historical interest, while others will illustrate various possible improvements. We will prove many of them thanks to a general bound due to~\citet{ger2009}. We will then focus more specifically on deep learning: we explain which bound, and which improvements led to tight generalization bounds for deep learning (Section \ref{subsec:neural:nets}). In particular, we will focus on a very important approach to improve the bounds: data-dependent priors.

\section{A Few PAC-Bayes Bounds}
\label{subsec:few:pac}

Note that the original works on PAC-Bayes focused only on classification with the 0-1 loss~\citep{mca1998}. However, they can be directly extended to any loss taking values in $[0,1]$~\citep[Lemma 3]{mau2004}. So, we will state all the following bounds for any $[0,1]$-valued loss $\ell$. Thus, $R$ and $r$ also take value in $[0,1]$ and $C=1$ in this section. Many of these bounds can be derived from Theorem~\ref{thm:bound:germain} due to~\citet{ger2009}. So we will postpone the proofs to after the statement of Theorem~\ref{thm:bound:germain}.

\subsection{The first PAC-Bayes bounds}

The very first paper on PAC-Bayes bounds by \citet{mca1998} was stated for a finite or denumerable set $\Theta$. Let us start with the first bound for a general $\Theta$, by \citet{mca1999}.
\begin{theorem}[Theorem 1, \citet{mca1999}]
\label{thm:mca:bound}
 For any $\delta>0$,
  \begin{multline*}
 \mathbb{P}_\mathcal{S} \Biggl[ \forall \rho\in\mathcal{P}(\Theta) \text{, } \mathbb{E}_{\theta\sim\rho}[ R(\theta)]
 \leq \mathbb{E}_{\theta\sim\rho}[ r(\theta)]
 \\
 + \sqrt{\frac{KL(\rho\|\pi) + \log\frac{1}{\delta} + \frac{5}{2}\log(n) +8 }{2n-1}} \Biggr] \geq 1-\delta.
 \end{multline*}
 \end{theorem}
 Compared to Theorem~\ref{thm:first:bound}, note that there is no parameter $\lambda$ here to optimize. On the other hand, one can no longer use Lemma~\ref{lemma:dv} to minimize the right-hand side. A way to solve this problem is to make the parameter $\lambda$ appear artificially using the inequality $\sqrt{ab}\leq a\lambda/2 + b/(2\lambda)$ for any $\lambda>0$:
   \begin{multline}
 \mathbb{P}_\mathcal{S} \Biggl[ \forall \rho\in\mathcal{P}(\Theta) \text{, } \mathbb{E}_{\theta\sim\rho}[ R(\theta)]
 \\
 \leq \mathbb{E}_{\theta\sim\rho}[ r(\theta)] + \inf_{\lambda>0} \left\{ \frac{\lambda}{4n-2} + \frac{KL(\rho\|\pi) + \log\frac{2}{\delta} + \frac{1}{2}\log(n)  }{2\lambda} \right\} \Biggr] \\
 \geq 1- \delta.
 \label{bound:mau:2}
 \end{multline}
 On the other hand, the price to pay for an optimization with respect to $\lambda$ in Theorem~\ref{thm:first:bound:lambda:grid} was a $\log(n)$ term, that is already in McAllester's bound, with an arithmetic grid, and a $\log\log(n)$ term when using a geometric grid. Of course, the geometric grid leads to an additional constraint that $\lambda\in[1,n]$, but one can easily check that this is always the case when~\eqref{bound:mau:2} is nonvacuous anyway. So, asymptotically in $n$, Theorem~\ref{thm:first:bound:lambda:grid} with a geometric grid will lead to better results than Theorem~\ref{thm:mca:bound}. On the other hand, the constants in Theorem~\ref{thm:mca:bound} are smaller, so the bound can be better for small sample sizes (a point that should not be neglected for tight certificates in practice!).
 
 It is possible to minimize the right-hand side in~\eqref{bound:mau:2} with respect to $\rho$, and this will lead to a Gibbs posterior: $\hat{\rho}=\pi_{-2\lambda r}$. It is also possible to minimize it with respect to $\lambda$, but the minimization in $\lambda$ when $\rho$ itself depends on $\lambda$ is a bit more tricky (for example, there might be local minima). We will discuss below a more recent bound due to~\citet{thi2017}, that can be minimized efficiently in both $\lambda$ and $\rho$.
 
\subsection{A key for tigher bounds: $kl$-PAC-Bayes bounds}
 
Let us now propose a completely different bound. This bound is very central in the PAC-Bayesian theory: we will see that many other bounds can be derived from this one. A first version was proven by \citet{lan2001,see2002} and is often refered to as Seeger's bound. The bound was slightly improved by \citet{mau2004}, so we will here provide Maurer's version of Seeger's bound.

Let $\mathcal{B}e(p)$ denote the probability distribution of a Bernoulli random variable $V$ with parameter $p$, that is, $\mathbb{P}(V=1)=p=1-\mathbb{P}(V=0)$. Then we have:
$$
KL(\mathcal{B}e(p)\|\mathcal{B}e(q)) = p\log\frac{p}{q} + (1-p)\log\frac{1-p}{1-q} =: kl(p\|q)
$$
which is interpreted as a statistical distance between $p$ and $q$.
\begin{theorem}[Theorem 5, \citet{mau2004}]
\label{thm:see:bound}
 For any $\delta>0$,
  \begin{multline*}
 \mathbb{P}_\mathcal{S} \Biggl[ \forall \rho\in\mathcal{P}(\Theta) \text{, } kl\Bigl( \mathbb{E}_{\theta\sim\rho}[ r(\theta)] \left\|  \mathbb{E}_{\theta\sim\rho}[ R(\theta)]\right.\Bigr)
 \\
 \leq \frac{KL(\rho\|\pi) + \log \frac{2\sqrt{n}}{\delta} }{n} \Biggr] \geq 1-\delta.
 \end{multline*}
 \end{theorem}
Under this form, the bound is not very explicit, so we will derive more explicit versions. Following \citet{see2002}, we define:
$$ kl^{-1} \left( p \left\| b \right.\right) = \sup\{q \in[0,1]: kl(p\|q) \leq b \} . $$
Then the bound becomes:
  \begin{multline*}
 \mathbb{P}_\mathcal{S} \Biggl[ \forall \rho\in\mathcal{P}(\Theta) \text{, } \mathbb{E}_{\theta\sim\rho}[ R(\theta)]
 \\
 \leq
 kl^{-1}\left( \mathbb{E}_{\theta\sim\rho}[ r(\theta)] \left\| \frac{KL(\rho\|\pi) + \log \frac{2\sqrt{n}}{\delta} }{n} \right.\right) \Biggr] \geq 1-\delta.
 \end{multline*}
We can deduce more explicit bounds from Seeger's bound simply by providing explicit bounds on the function $kl^{-1}$.

As a first application of this method, Pinsker's inequality~\citep[Theorem 4.19]{bou2013} implies $kl(p\|q)\geq 2 (p-q)^2$, and thus $kl^{-1}(p\|b) \leq  p + \sqrt{b/2}$. Thus, Theorem~\ref{thm:see:bound} implies, for any $\delta>0$,
  \begin{multline*}
 \mathbb{P}_\mathcal{S} \Biggl[ \forall \rho\in\mathcal{P}(\Theta) \text{, } \mathbb{E}_{\theta\sim\rho}[ R(\theta)]
 \leq \mathbb{E}_{\theta\sim\rho}[ r(\theta)]
 \\
 + \sqrt{\frac{KL(\rho\|\pi) + \log\frac{2\sqrt{n}}{\delta} }{2n}} \Biggr] \geq 1-\delta
 \end{multline*}
which is essentially McAllester's bound with improved constants. But we can do much better.
 
\subsection{Explicit tight bounds}
 
Based on an idea due to~\citet{mca2003},~\citet{tol2013} derived upper bounded $kl^{-1}$ using relaxed Pinsker's inequality to prove the following result.
\begin{theorem}[(3), \citet{tol2013}]
\label{thm:tol:bound}
 For any $\delta>0$,
  \begin{multline*}
 \mathbb{P}_\mathcal{S} \Biggl[ \forall \rho\in\mathcal{P}(\Theta) \text{, } \mathbb{E}_{\theta\sim\rho}[ R(\theta)] 
 \leq
 \mathbb{E}_{\theta\sim\rho}[ r(\theta)]
 \\
 + \sqrt{ 2 \mathbb{E}_{\theta\sim\rho}[ r(\theta)]  \frac{KL(\rho\|\pi) + \log \frac{2\sqrt{n}}{\delta} }{ \textcolor{red}{n} } }
 \\
 +2 \frac{KL(\rho\|\pi) + \log \frac{2\sqrt{n}}{\delta} }{\textcolor{red}{n}} \Biggr] \geq 1-\delta.
 \end{multline*}
 \end{theorem}
Note something amazing: while the dependence of this bound in $n$ is in general in $1/\sqrt{n}$, as all the PAC-Bayes bounds seen so far, it drops to $1/n$ if $ \mathbb{E}_{\theta\sim\rho}[ r(\theta)]=0$. This was actually not a surprise, because a similar phenomenon is known for the ERM~\citep{vap1998}.

More generally, we will see in Section~\ref{section:oracle} a general assumption that characterizes the best possible learning rate in classification problems. And as a special case, the noiseless case indeed leads to rates in $1/n$. Let us summarize the important take home message from Theorem~\ref{thm:tol:bound}:
\begin{itemize}
 \item the empirical PAC-Bayes bounds seen so far were in $1/\sqrt{n}$,
 \item in the noiseless case $  \mathbb{E}_{\theta\sim\rho}[ r(\theta)]=0$, it is possible to have a bound in $1/n$, on the condition that one uses the right PAC-Bayes inequality, for example Theorem~\ref{thm:tol:bound} or Theorem~\ref{thm:see:bound}.
\end{itemize}
This is very important for the application of these bounds to neural networks, as deep networks usually allow to classify the training data perfectly.

Another example of a theorem derived from Seeger's bound appears to be very convenient for a joint minimization in $\lambda$ and $\rho$, and is also extremely tight (see Subsection~\ref{subsec:neural:nets} below).
\begin{theorem}[Theorem 3, \citet{thi2017}]
\label{thm:thieman:bound}
 For any $\delta>0$,
 
 \vspace{-0.8\baselineskip}
 \small
  \begin{equation*}
 \mathbb{P}_\mathcal{S} \Biggl[ \forall \rho\in\mathcal{P}(\Theta) \text{, }  \mathbb{E}_{\theta\sim\rho}[ R(\theta)] \leq \textcolor{red}{\inf_{\lambda\in(0,2)}} \left\{ \frac{\mathbb{E}_{\theta\sim\rho}[ r(\theta)] }{1-\frac{\lambda}{2}} +  \frac{KL(\rho\|\pi) + \log \frac{2\sqrt{n}}{\delta} }{n\lambda\left(1-\frac{\lambda}{2}\right)} \right\} \Biggr] \geq 1-\delta.
 \end{equation*}\normalsize
 \end{theorem}
 Here again, we observe the $1/n$ regime when $\mathbb{E}_{\theta\sim\rho}[ r(\theta)]=0$ (for example for $\lambda=1$).

Finally, note PAC-Bayes bounds with both $1/n$ and $1/\sqrt{n}$ can be derived without using the $kl$-PAC-Bayes bound. We provide as an example this result by \citet{cat2007}.
\begin{theorem}[Theorem 1.2.6 page 11, \citet{cat2007}]
\label{thm:catoni:2}
 Define, for $a>0$, the function of $p\in(0,1)$
 $$ \Phi_a(p) = \frac{-\log\left\{1- p\left[1-{\rm e}^{-a}\right] \right\}}{a} .$$
 Then, for any $\lambda>0$, for any $\epsilon>0$,
 \small
\begin{multline}
\label{thm:bound:catoni}
 \mathbb{P}_\mathcal{S} \left\{
 \forall\rho\in\mathcal{P}(\Theta)\text{, }
 \mathbb{E}_{\theta\sim\rho}[ R(\theta)]
 \leq
 \Phi_{\frac{\lambda}{n}}^{-1}\left[
  \mathbb{E}_{\theta\sim\rho}[ r(\theta)]
  + \frac{KL(\rho\|\pi) + \log\frac{1}{\delta} }{\lambda}
  \right]
 \right\}
 \\
 \geq 1-\delta.
\end{multline}\normalsize
\end{theorem}
 Theorem~\ref{thm:catoni:2} can be proven using a bound of \citet{ger2009}, that is also used to prove the $kl$-PAC-Bayes bound. Theorem~\ref{thm:catoni:2} can be made a little more explicit.
We have
$$ \Phi_{a}^{-1}(q) = \frac{1-{\rm e}^{-a q}}{1-{\rm e}^{-a}} \leq \frac{aq}{1-{\rm e}^{-a}}  ,$$
and thus:
\begin{multline}
\label{thm:bound:catoni:2}
 \mathbb{P}_\mathcal{S} \Biggl\{
 \forall\rho\in\mathcal{P}(\Theta)\text{, }
 \mathbb{E}_{\theta\sim\rho}[ R(\theta)]
 \leq
\frac{\lambda}{n\left[1-{\rm e}^{\frac{-\lambda}{n}} \right]}\Biggl[
  \mathbb{E}_{\theta\sim\rho}[ r(\theta)]
  \\
  + \frac{KL(\rho\|\pi) + \log\frac{1}{\delta} }{\lambda}
 \Biggr]
 \Biggr\}\geq 1-\delta.
 \end{multline}

Before we prove all these bounds, let us apply them to the same examples than in Section~\ref{section:first:step} to highlight the improvements. I will actually use Theorem~\ref{thm:thieman:bound}, the application of the other bounds can be done in a similar way.
\begin{example}[Finite case]
 We start with a continuation of Example~\ref{exm:first:finite}: ${\rm card}(\Theta)=M<+\infty$, and $\pi$ is a uniform distribution.
 For any $\delta>0$,
$$
 \mathbb{P}_\mathcal{S} \left\{ \forall \lambda\in(0,2)\text{, }  \mathbb{E}_{\theta\sim\hat{\rho}_{\lambda}}[ R(\theta)] \leq \inf_{\theta\in\Theta} \frac{ r(\theta) }{1-\frac{\lambda}{2}} +  \frac{\log \frac{2 M\sqrt{n}}{\delta} }{n\lambda\left(1-\frac{\lambda}{2}\right)}
 \right\} \geq 1-\delta.
$$
If $r(\hat{\theta}_{{\rm ERM}})=0$, the inequality becomes:
$$
 \mathbb{E}_{\theta\sim\hat{\rho}_{\lambda}}[ R(\theta)] \leq  \frac{\log \frac{2 M\sqrt{n}}{\delta} }{n\lambda\left(1-\frac{\lambda}{2}\right)},
$$
which is exactly minimized for $\lambda=1$. Thus, in this case, we put $\hat{\lambda}=1$ and we obtain:
$$
 \mathbb{E}_{\theta\sim\hat{\rho}_{\hat{\lambda}}}[ R(\theta)] \leq  \frac{2 \log \frac{2 M\sqrt{n}}{\delta} }{n}.
$$
In the case where $r(\hat{\theta}_{{\rm ERM}})>0$, we could minimize the right-hand side exactly in $\lambda$. However, even an approximate minimization is enough to obtain a good bound in this toy example. For example, with
$$\hat{\lambda} = \sqrt{ \frac{\log \frac{2 M\sqrt{n}}{\delta}}{ n }}, $$
we obtain, for $n$ large enough so that $\hat{\lambda}<2$,
\begin{align*}
 \mathbb{E}_{\theta\sim\hat{\rho}_{\hat{\lambda}}}[ R(\theta)] & \leq
\frac{\inf_{\theta\in \Theta}  r(\theta) + \sqrt{\frac{ \log \frac{2 M\sqrt{n}}{\delta} }{ n} } }{ 1 - \frac{1}{2}\sqrt{ \frac{\log \frac{2 M\sqrt{n}}{\delta}}{ n }} }
\\
& = \left[\inf_{\theta\in \Theta}  r(\theta) + \sqrt{\frac{ \log \frac{2 M\sqrt{n}}{\delta} }{ n} } \right](1+o(1)).
\end{align*}
Thus, Theorem~\ref{thm:thieman:bound} allows an empirical definition of $\hat{\lambda}$ that leads essentially to the same bound than in Example~\ref{exm:first:finite}, up to the $\log n$ term, in general. Moreover, in the case $r(\hat{\theta}_{{\rm ERM}})=0$, the rate is much better.
\end{example}

\begin{example}[Gaussian priors]
\label{example:gaussian:unbounded}
We come back to the setting of Example~\ref{exm:lipschitz:gauss}: $\Theta=\mathbb{R}^d$, the function $\theta\mapsto\ell(f_\theta(x),y)$ is $L$-Lipschitz and $\pi$ is a Gaussian $\pi=\mathcal{N}(0,\sigma^2 I_d)$, where $I_d$ is the $d\times d$ identity matrix.
Plugging the formulas derived in Example~\ref{exm:lipschitz:gauss} into Theorem~\ref{thm:thieman:bound}, we obtain:
\begin{multline*}
 \mathbb{P}_\mathcal{S} \Biggl[ \forall \lambda\in(0,2) \text{, }  \mathbb{E}_{\theta\sim\hat{\rho}_{\lambda}}[ R(\theta)] \leq \inf_{m\in\mathbb{R}^d, s>0} \frac{\mathbb{E}_{\theta\sim\mathcal{N}(m,s^2 I_d)}[ r(\theta)] }{1-\frac{\lambda}{2}}
 \\
 +   \frac{ \frac{\|m\|^2}{2\sigma^2} + \frac{d}{2}\left[\frac{s^2}{\sigma^2} + \log\frac{\sigma^2}{s^2} -1 \right] + \log \frac{2\sqrt{n}}{\delta} }{n\lambda\left(1-\frac{\lambda}{2}\right)} \Biggr] \geq 1-\delta.
 \end{multline*}
 Note that, when compared to Example~\ref{exm:lipschitz:gauss}, a major advantage here is that the bound allows to minimize the right-hand side in $m$, $\sigma$ and $\lambda$ without having to impose an upper bound on $\|m\|$.
\end{example}

\subsection{Some proofs}
\enlargethispage{\baselineskip}
We start this section with a nice generalization of Theorem~\ref{thm:see:bound} by \citet{ger2009}. The first reason why we state this result is that it can be used to prove all the bounds mentioned since the beginning of this section, which will be done in this subsection.
The other reason is that it can be used to derive bounds for unbounded losses, and for non-i.i.d. observations. This point will be discussed in Section~\ref{section:not:bounded:iid}.
\begin{theorem}[Theorem 2.1, \citet{ger2009}]
\label{thm:bound:germain}
Let $\mathcal{D}:[0,1]^2\rightarrow \mathbb{R} $ be any convex function. For any $\delta>0$,
  \begin{multline*}
 \mathbb{P}_\mathcal{S} \Biggl[ \forall \rho\in\mathcal{P}(\Theta) \text{, } \mathcal{D} \left( \mathbb{E}_{\theta\sim\rho}[ r(\theta)] ,  \mathbb{E}_{\theta\sim\rho}[ R(\theta)]\right)
 \\
 \leq \frac{KL(\rho\|\pi) + \log \frac{ \mathbb{E}_{\theta\sim\pi}
 \mathbb{E}_{\mathcal{S}} {\rm e}^{n\mathcal{D}(r(\theta),R(\theta)) }
 }{\delta} }{n} \Biggr] \geq 1-\delta.
 \end{multline*}
\end{theorem}
\begin{proof}[Proof of Theorem~\ref{thm:bound:germain}]
For short, put
\begin{align*}
\mathrm{M}(\pi) &
= \mathbb{E}_{\mathcal{S}} \mathbb{E}_{\theta\sim\pi}{\rm e}^{n\mathcal{D}(r(\theta),R(\theta)) }
\\
& = \mathbb{E}_{\theta\sim\pi} \mathbb{E}_{\mathcal{S}} {\rm e}^{n\mathcal{D}(r(\theta),R(\theta)) },
\end{align*}
the second inequality being due to Tonelli's theorem, and observe that
$$
 \mathbb{E}_{\mathcal{S}}  {\rm e}^{ \sup_{\rho\in\mathcal{P}(\Theta)} \left[ n \mathbb{E}_{\theta\sim\rho}[\mathcal{D}(r(\theta),R(\theta))] - KL(\rho\|\pi) \right] } = \mathrm{M}(\pi)
$$
(using Donsker and Varadhan's inequality).
Multiply both sides by $\delta/\mathrm{M}(\pi)$ to get:
$$
 \mathbb{E}_{\mathcal{S}}  {\rm e}^{ \sup_{\rho\in\mathcal{P}(\Theta)} \left[ n \mathbb{E}_{\theta\sim\rho}[\mathcal{D}(r(\theta),R(\theta))] - KL(\rho\|\pi) \right]-\log\frac{\mathrm{M}(\pi)}{\delta} }
= \delta .
$$
Thus, with probability at least $1-\delta$,
$$
\sup_{\rho\in\mathcal{P}(\Theta)} \Bigl[ n \mathbb{E}_{\theta\sim\rho}[\mathcal{D}(r(\theta),R(\theta))]  - KL(\rho\|\pi) \Bigr] -\log\frac{\mathrm{M}(\pi)}{\delta} \leq 0
$$
and rearranging terms, for all $\rho\in\mathcal{P}(\Theta)$,
$$
\mathbb{E}_{\theta\sim\rho}[\mathcal{D}(r(\theta),R(\theta))]
\leq
\frac{KL(\rho\|\pi) + \log\frac{\mathrm{M}(\pi)}{\delta}
}{n}.
$$
Finally, thanks to the convexity of $\mathcal{D}$, this implies
$$
\mathcal{D}(\mathbb{E}_{\theta\sim\rho}[r(\theta)],\mathbb{E}_{\theta\sim\rho}[R(\theta)])
\leq
\frac{KL(\rho\|\pi) + \log\frac{\mathrm{M}(\pi)}{\delta}
}{n}.
$$
\end{proof}
We will now prove Theorem~\ref{thm:see:bound} from Theorem~\ref{thm:bound:germain}. This requires the following lemma.

\enlargethispage{-3\baselineskip}
\begin{lemma}[Theorem 1 of \cite{mau2004}]
\label{lemma:maurer:old}
For any $\theta\in\Theta$,
$$
 \mathbb{E}_{\mathcal{S}} {\rm e}^{n.kl(r(\theta)\|R(\theta)) } \leq 2\sqrt{n}.
 $$
\end{lemma}
We are now in position to prove Theorem~\ref{thm:see:bound}.
\begin{proof}[Proof of Theorem~\ref{thm:see:bound}]
 We apply Theorem~\ref{thm:bound:germain} to $\mathcal{D}(x,y) = kl(x\|y)$. We obtain, with probability at least $1-\delta$, for any $\rho\in\mathcal{P}(\Theta)$,
\begin{align*}
 kl\left( \mathbb{E}_{\theta\sim\rho}[ r(\theta)] \|  \mathbb{E}_{\theta\sim\rho}[ R(\theta)]\right)
 &
 \leq \frac{KL(\rho\|\pi) + \log \frac{ \mathbb{E}_{\theta\sim\pi}
 \mathbb{E}_{\mathcal{S}} {\rm e}^{n.kl(r(\theta)\| R(\theta)) }
 }{\delta} }{n}
 \\
 & \leq \frac{KL(\rho\|\pi) + \log \frac{ 2\sqrt{n}
 }{\delta} }{n},
 \end{align*}
 we used Lemma~\ref{lemma:maurer:old} in the second inequality.
\end{proof}
\begin{proof}[Proof of Theorem~\ref{thm:tol:bound}]
We recall the relaxed version of Pinkser's inequality~\citep[Lemma 8.4]{bou2013}: for any $(p,q)\in(0,1)^2$,
 $$
 kl(p\|q) \leq \frac{(p-q)^2}{2q}.
 $$
 This leads to
 $$  (p-q)^2 \leq 2q kl(p\|q) $$
 and thus
 \begin{equation}
 \label{piiinsker:1}
 q \leq p + \sqrt{ 2q kl(p\|q)}
 \end{equation}
 which implies, using $q\leq 1$,
  \begin{equation}
 \label{piiinsker:2}
 q \leq p + \sqrt{ 2 kl(p\|q)}.
 \end{equation}
 We plug~\eqref{piiinsker:2} into~\eqref{piiinsker:1} to get
 $$ q \leq p + \sqrt{ 2\left(p+\sqrt{ 2 kl(p\|q)}\right) kl(p\|q)}
  \leq \sqrt{ 2p kl(p\|q)} + 2 kl(p\|q) $$
and
$$ kl^{-1}(p\|b) \leq p + \sqrt{2p b} + 2b. $$
Plug this upper bound into Theorem~\ref{thm:see:bound} to get the stated inequality.
\end{proof}
\begin{proof}[Proof of Theorem~\ref{thm:thieman:bound}]
We follow the previous proof until~\eqref{piiinsker:1}. Then, remark that for any $a,b>0$,
$$ \sqrt{ab} =\inf_{\lambda>0}  \frac{\lambda a}{2} + \frac{b}{2\lambda}, $$
and apply this inequality to $a=q$ and $b = 2 kl(p\|q) $. We obtain, for any $\lambda>0$,
$$  q \leq p + \frac{q \lambda}{2} + \frac{ kl(p\|q) }{\lambda} $$
and thus
$$ q\left( 1 - \frac{\lambda}{2} \right) \leq p + \frac{ kl(p\|q) }{\lambda} .$$
If $\lambda<2$, we can divide both sides by $1-\lambda/2>0$, which gives:
$$ q \leq \frac{p}{1-\frac{\lambda}{2}} + \frac{ kl(p\|q) }{\lambda\left(1-\frac{\lambda}{2}\right)} .$$
Thus,
$$ kl^{-1}(p\|b) \leq \inf_{0\leq \lambda \leq 2} \left[ \frac{p}{1-\frac{\lambda}{2}} + \frac{ b }{\lambda\left(1-\frac{\lambda}{2}\right)} \right] .$$
Plug this upper bound into Theorem~\ref{thm:see:bound}.
\end{proof}
\begin{proof}[Sketch of the proof of Theorem~\ref{thm:bound:catoni}]
 This result can be obtained by an application of Theorem~\ref{thm:bound:germain} with  $\mathcal{D}(p,q)=-\log[1-p(1-{\rm e}^{-a})]-aq$. The control of the exponential moment
 $$ \mathrm{M}(\pi)
= \mathbb{E}_{\mathcal{S}} \mathbb{E}_{\theta\sim\pi}{\rm e}^{n\mathcal{D}(r(\theta),R(\theta)) } $$
is provided in Corollary 2.2 of~\citet{ger2009} in the case of the binary loss, and can be extended to any
$[0,1]$-valued loss thanks to Lemma 3 of \citet{mau2004}.
\end{proof}

These derivations lead to a natural question: is there a function $\mathcal{D}$ that will lead to a strict improvement of Theorem~\ref{thm:see:bound}? The question is investigated by \citet{foong2021tight}. Overall, it seems that no function $\mathcal{D}$ will lead to a bound that will be smaller in expectation than the one in Theorem~\ref{thm:see:bound}, up to the $\log(2\sqrt{n})$ term.

\subsection{PAC-Bayes-Bernstein}

The first PAC-Bayes bound introduced in Section~\ref{section:first:step}, Theorem~\ref{thm:first:bound}, uses Hoeffding's inequality to control the expectation of the exponential of the generalization error. For this reason, it is sometimes refered to as PAC-Bayes-Hoeffding bound. Most bounds seen in Section~\ref{section:tight} so far were a consequence of another control of this exponential moment: Lemma~\ref{lemma:maurer:old}.

We will conclude this necessarily non-exhaustive survey on empirical PAC-Bayes bounds by covering another family of bounds, known as PAC-Bayes-Bernstein, because they rely on Bernstein's inequality to control the exponential moment. There are actually many versions of Bernstein's inequality. For a proof of the one we state here, see \citet[Theorem 5.2.1]{cat2004}, or~\citet{mcdiarmid1998concentration}.
\begin{lemma}[Bernstein's inequality]
\label{lemma:bernstein}
Let $g$ denote the Bernstein function defined by $g(0) = 1$ and, for $x\neq 0$,
$$ g(x) = \frac{{\rm e}^x - 1 - x}{x^2}. $$
Let $U_1,\dots,U_n$ be i.i.d. random variables such that $\mathbb{E}(U_i)$ is well defined and $U_i- \mathbb{E}(U_i) \leq C$ almost surely for some $C\in\mathbb{R}$. Then
$$
\mathbb{E}\left( {\rm e}^{t \sum_{i=1}^n [U_i - \mathbb{E}(U_i) ] } \right)
\leq {\rm e}^{ g\left(C t\right) n t^2 {\rm Var}(U_i) }.
$$
\end{lemma}
Before discussing PAC-Bayes bounds based on Lemma~\ref{lemma:bernstein}, let us compare it to our application of Hoeffding's inequality in Section~\ref{section:introduction}.
As $\ell_i(\theta)\in[0,1]$ here, an application of Lemma~\ref{lemma:bernstein} to $U_i =\mathbb{E}[\ell_i(\theta)] - \ell_i(\theta) $ gives:
$$
\mathbb{E}\left( {\rm e}^{tn [R(\theta)-r(\theta)] } \right)
\leq {\rm e}^{ g\left(t\right) n t^2 {\rm Var}(\ell_i(\theta)) }.
$$
The term ${\rm Var}(\ell_i(\theta))$ is of particular interest, so we introduce a shorter notation.
\begin{definition}
 We put ${\rm V}(\theta) = {\rm Var}(\ell_i(\theta))$.
\end{definition}
In order to ease the comparison with the previous section, we also put $ \lambda = nt$.
Chernoff's bounding technique leads to
$$
 \mathbb{P}_\mathcal{S}\left(R(\theta) > r(\theta) + s \right) \leq {\rm e}^{ g\left(\frac{\lambda}{n}\right) \frac{\lambda^2}{n} {\rm V}(\theta) - \lambda n s}.
$$
We define $\delta$ as the right-hand side and consider the complementary event:
$$
 \mathbb{P}_\mathcal{S}\left(R(\theta)  \leq r(\theta) + \frac{g\left(\frac{\lambda}{n}\right) \lambda {\rm V}(\theta) }{n}  + \frac{\log \frac{1}{\delta} }{\lambda} \right) \geq 1-\delta.
$$
The exact optimization of the right-hand side in $\lambda$ is not straightforward because of the function $g$, but an approximate minimization is enough to highlight the benefits of Bernstein's inequality. In the regime $ \lambda<n$, $g(\lambda/n) \leq g(1) <1$ and thus the bound becomes
$$
 \mathbb{P}_\mathcal{S}\left(R(\theta)  \leq r(\theta) + \frac{ \lambda {\rm V}(\theta) }{n}  + \frac{\log \frac{1}{\delta} }{\lambda} \right) \geq 1-\delta.
$$
The choice $\lambda = \sqrt{ n / {\rm V}(\theta)} $ would lead to
$$
 \mathbb{P}_\mathcal{S}\left(R(\theta)  \leq r(\theta) + \sqrt{\frac{{\rm V}(\theta)}{n}}\left( 1+\log\frac{1}{\delta} \right)  \right) \geq 1-\delta.
$$
Note however that when ${\rm V}(\theta)<1/n$, $\sqrt{ n / {\rm V}(\theta)} > n$. In this case, we propose simply to take $\lambda=n$ to get:
$$
 \mathbb{P}_\mathcal{S}\left(R(\theta)  \leq r(\theta) + \frac{ 1+\log\frac{1}{\delta}}{n}  \right) \geq 1-\delta.
$$
Combining both inequalities, we obtain
\begin{equation}
\label{equa:appli:bernstein} 
 \mathbb{P}_\mathcal{S}\left(R(\theta)  \leq r(\theta) +   \min\left( \sqrt{\frac{{\rm V}(\theta)}{n}},\frac{1}{n}\right) \left( 1+\log\frac{1}{\delta} \right) \right) \geq 1-\delta.
\end{equation}
It is important to compare~\eqref{equa:appli:bernstein} to bounds obtained by Hoeffding's inequality, such as~\eqref{equa:appli:hoeffding:3}. Of course, as the loss takes values in $[0,1]$ here, we could upper bound ${\rm V}(\theta)$ by $1/4$, and in this case we obtain something comparable to~\eqref{equa:appli:hoeffding:3}. However, when ${\rm V}(\theta)$ is small,~\eqref{equa:appli:bernstein} is much better.

Using the proof technique of Section~\ref{section:first:step}, we obtain the following PAC-Bayes bound: for any $\lambda>0$, for any $\delta\in(0,1)$,
\begin{multline*}
 \mathbb{P}_\mathcal{S}\Biggl(\forall\rho\in\mathcal{P}(\Theta) \text{, } \mathbb{E}_{\theta\sim\rho}[ R(\theta)]
 \\
 \leq \mathbb{E}_{\theta\sim\rho}[ r(\theta)] +  \frac{\lambda g\left(\frac{\lambda}{n}\right) \mathbb{E}_{\theta\sim\rho}[ {\rm V}(\theta)] }{n} + \frac{KL(\rho\|\pi) + \log\frac{1}{\delta}}{\lambda} \Biggr)
 \geq 1-\delta.
\end{multline*}
Unfortunately, this bound is useless in practice, as the right-hand side depends on the unknown distribution of the observations via the variance term ${\rm V}$.

It turns out that it is possible to upper bound the term $ \mathbb{E}_{\theta\sim\rho}[ {\rm V}(\theta)]$ by an empirical variance term to obtain a so-called empirical PAC-Bayes-Bernstein bound~\citep{pmlr-v26-seldin12a,tol2013}.
For example, \citet[Theorem 3]{tol2013} provided such an upper bound, $\mathcal{V}(\rho,\pi,\delta)$ below, to prove the following result. We will omit the proof here.
\begin{theorem}[Theorem 4, \citet{tol2013}]
\label{thm:empirical:bernstein}
Fix any $c_1,c_2>0$.
Put, for any $\theta\in\Theta$,
$$ \hat{V}(\theta) = \frac{1}{n-1}\sum_{i=1}^n ( \ell_i(\theta) - r(\theta) )^2.  $$
Put
\begin{multline*}
\mathcal{V}(\rho,\pi,\delta)
=  \mathbb{E}_{\theta\sim\rho}[ \hat{V}(\theta)]
+ 2c_2 \frac{KL(\rho\|\pi) + \log \frac{2\nu_2}{\delta}}{n-1}
\\
+ (1+c_2) \sqrt{ \frac{  \mathbb{E}_{\theta\sim\rho}[ \hat{V}(\theta)] \left(KL(\rho\|\pi) + \log \frac{2\nu_2}{\delta}\right)  }{2(n-1) } }
\end{multline*}
where
$$ \nu_2 = \left\lceil \frac{1}{\log c_2} \log\left( \frac{1}{2} \sqrt{\frac{n-1}{\log \frac{2}{\delta}} +1}+\frac{1}{2} \right) \right\rceil .$$
Then, for any $\delta\in(0,1)$,
\begin{multline*}
 \mathbb{P}_\mathcal{S}\Biggl(\forall\rho\in\mathcal{P}(\Theta) \text{, } \mathbb{E}_{\theta\sim\rho}[ R(\theta)] \leq  \mathbb{E}_{\theta\sim\rho}[ r(\theta)] +
  \mathcal{B}(\rho,\pi,\delta)
 \Biggr)
 \geq 1-\delta,
\end{multline*}
where
$$
  \mathcal{B}(\rho,\pi,\delta)
  = 
 (1+c_1) \sqrt{ \frac{  ({\rm e}-2) \mathcal{V}(\rho,\pi,\delta) \left( KL(\rho\|\pi) + \log \frac{2\nu_1}{\delta}  \right) }{  n } } 
$$
for all $\rho$ such that
$$
\sqrt{\frac{ KL(\rho\|\pi) + \log \frac{2\nu_1}{\delta} }{ ({\rm e}-2)  \mathbb{E}_{\theta\sim\rho}[ \hat{V}(\theta)]}} \leq \sqrt{n}
$$
and
$$
  \mathcal{B}(\rho,\pi,\delta)
  = 2 \frac{KL(\rho\|\pi) + \log \frac{2\nu_1}{\delta}}{n}
$$
for all other $\rho$, and
$$ \nu_1 = \left\lceil \frac{1}{\log c_1} \log\left( \sqrt{\frac{({\rm e}-2)n}{4\log \frac{2}{\delta}}} \right) \right\rceil .$$
\end{theorem}
Let us compare this bound to Theorems~\ref{thm:thieman:bound} and \ref{thm:tol:bound}. If $\theta$ is such that $r(\theta)=0$, then we also have direclty $\hat{{\rm V}}(\theta)=0$. Thus, when there is a $\rho$ such that $ \rho$-almost surely, $r(\theta)=0$, then Theorem~\ref{thm:empirical:bernstein} leads to a bound in $1/n$ rather than in $1/\sqrt{n}$. Moreover, as $\ell_i(\theta)\in[0,1]$ here, $\ell_i(\theta)^2 \leq \ell_i(\theta)$ and thus, we can expect that $\hat{{\rm V}}(\theta) \leq r(\theta) $. Thus, even when there is no perfect predictor, Theorem~\ref{thm:empirical:bernstein} can lead to a tighter bound than Theorem~\ref{thm:tol:bound}. Empirical PAC-Bayes-Bernstein bounds became an important research direction, with improvements of the original bound, and many applications~\citep{tol2013,mhammedi2019pac,wu2021cheby,wu2022split,jang2023tighter}. Finally, some oracle PAC-Bayes bounds discussed in Section~\ref{section:oracle} also rely on Bernstein's inequality.

More bounds are known, but it's not possible to mention all, so I apologize if I didn't cite a bound you like, or your bound. Many refinments of the above bounds are discussed in a recent preprint by~\citet{rod2023}. Some other variants will be discussed later, especially in Section~\ref{section:related}: bounds for unbounded losses $\ell$, bounds for non i.i.d. data, and also some bounds where the KL divergence $KL(\rho\|\pi)$ is replaced by another divergence.

\section{Tight Generalization Error Bounds for Deep Learning}
\label{subsec:neural:nets}

\subsection{A milestone: non vacuous generalization error bounds for deep networks by \citet{dzi2017}}

PAC-Bayes bounds were applied to (shallow) neural networks as early as in 2002 by \citet{langford2002not}.
We also applied them with O. Wintenberger to prove that shallow networks can consistently predict time series~\citep{alq2012}. McAllester proposed an application of PAC-Bayes bounds to dropout, a tool used for training neural networks, in his tutorial~\citep{mca2013}. But none of these techniques seemed to lead to tight empirical bounds for deep networks... until 2017, when \citet{dzi2017} obtained the first non-vacuous generalization error bounds for deep networks on the MNIST dataset based on Theorem~\ref{thm:see:bound} (Seeger's bound). Since then, there was a regain of interest in PAC-Bayes bounds to obtain the tightest possible certificates.

At first sight,~\citet{dzi2017} applied Seeger's bound to a deep neural network, but many important ideas and refinements were used to lead to a non vacuous bound (some being original, and according to the authors, some based on ideas from~\cite{langford2002not}). Let us describe here briefly these ingredients, the reader should of course read the paper for more details and insightful explanations:
\begin{itemize}
 \item the posterior is constrained to be Gaussian (similar to the above ``non-exact minimization of the bound'' in Section~\ref{subsec:non:exact}): $\rho_{w,s^2}$ $= \mathcal{N}(w,s^2 I_d)$. Thus, the PAC-Bayes bound only has to be minimized with respect to the parameter $(w,s^2)$, which allows to use an optimization algorithm to minimize the bound (the authors mention that fitting Gaussian distributions to neural networks was already proposed by \citet{hinton1993keeping} based on the MDL principle, which will be discussed in Section~\ref{section:related}).
 \item the choice of an adequate upper bound on $kl^{-1}$ in Seeger's bound in order to make the bound easier to minimize.
 \item Seeger's bound holds for the $0-1$ loss, but Dziugaite and Roy upper bounded the empirical risk by a convex, Lipschitz upper bound in order to make the bound easier to minimize (this is a standard approach in classification):
\begin{multline*}
\mathbb{E}_{\theta\sim\rho}[r(\theta)] = \mathbb{E}_{\theta\sim\rho}\left[\frac{1}{n}\sum_{i=1}^n \mathbf{1}(f_\theta(X_i)\neq Y_i)\right]
\\
\leq  \mathbb{E}_{\theta\sim\rho}\left[\frac{1}{n}\sum_{i=1}^n \frac{\log\left( 1 + {\rm e}^{-Y_i f_\theta(X_i)} \right)}{\log 2} \right] .
\end{multline*}
 \item the use of the Stochastic Gradient Algorithm (SGD) to minimize the bound (up to our knowledge, the first authors to use of SGD to minimize a PAC-Bayes bound was \citet{ger2009} for linear classification). Of course, this is standard in deep learning, but there is a crucial observation that SGD tends to converge to flat minima. This is very important, because around a flat minima $w^*$, we have $r(w)\simeq r(w^*)$ and thus $\mathbb{E}_{\theta\sim \rho_{w,s^2}}[r(\theta)] \simeq r(\theta^*)$ even for quite large values of $s^2$. On the other hand, for a sharp minimum $w^*$, $\mathbb{E}_{\theta\sim \rho_{w,s^2}}[r(\theta)] \simeq r(\theta^*)$ only for very small values of $s^2$ which tends to make the PAC-Bayes bound larger (see Example~\ref{exm:lipschitz:gauss} above).
 \item finally, the authors used a data dependent prior: $\mathcal{N}(w_0,\sigma^2 I)$, where $\sigma^2$ is chosen to minimize the bound (this is justified in theory thanks via a union bound argument as in Subsection~\ref{subsec:choice:lambda}). The mean $w_0$ is not optimized, but the authors point out that the choice $w_0=0$ is not suitable and they actually draw $w_0$ randomly, as is usually done in non-Bayesian approaches to initialize SGD.
\end{itemize}

On the MNIST data set, the authors obtain empirical bounds between $0.16$ and $0.22$, thus, non vacuous. The classification performance of their posterior is actually around $0.03$, so they conclude that there is still room for improvement. Indeed, since then, a wide range of papers, from very theoretical to very computational, studied PAC-Bayes bounds for deep networks~\citep{lon2017,ney2017,zhou2018non,let2019,via2019,rivasplata2019pac,lan2020,tsu2020,big2020,for2020,per2020,suzuki2020generalization,pmlr-v119-pitas20a,clerico2021wide,jin2022,clerico2022conditionally,trabs2022,gluch2023bayes,zhang2023auto}. We discuss recent results by \citet{per2020,clerico2022conditionally} below, but first, let us discuss in detail one of the most important ingredients above: the data-dependent prior.

\subsection{Bounds with data-dependent priors}
\enlargethispage{\baselineskip}
To use data in order to improve the prior is actually an old idea: we found such approaches works by \citet{seeger2003bayesian,cat2003,cat2004,zha2006,NIPS2006_3f5ee243,cat2007,lever2010,par2012,lev2013,dzi2017,dzi2018,pmlr-v130-karolina-dziugaite21a}. Note that the original PAC-Bayes bounds do not allow to take a data-dependent prior. Thus, some additional work is required to make this possible (e.g the union bound on $\sigma^2$ by \citet{dzi2017} discussed above). The first occurence of this idea is due to \citet[Chapter 4]{seeger2003bayesian}. Seeger proposed to split the sample in two parts. The first part is used to define $\Theta$ and the prior $\pi$, and the PAC-Bayes bound is applied on the second part of the sample (that is, conditionally on the first part). Seeger used this technique to study the generalization of Gaussian processes. Later \citet{cat2003} used it to prove generalization error bounds on compression schemes. We will here describe in detail two other approaches: first, Catoni's ``localization'' technique, because it will also be important in Sections~\ref{section:oracle} and~\ref{section:related}, and then a recent bound by \citet{dzi2018}.

\enlargethispage{\baselineskip}
First, let us discuss the intuition leading to Catoni's method. As can be seen in the elementary bounds in Section~\ref{section:first:step}, for example in Example~\ref{exm:first:finite}, the bound is tighter for parameters $\theta$ for which $\pi(\theta)$ is large, and less tight for parameters $\theta$ for which $\pi(\theta)$ is small: in the finite case, we recall that the Kullback-Leibler divergence led to a term in $\log(1/\pi(\theta))$ in the bound. Based on this idea, we might want to construct a prior $\pi$ that gives a large weight to the relevant parameters, that is, to parameters such that $R(\theta)$ is small. A possible prior is $\pi_{-\beta R}$ for some $\beta>0$, where $\pi_{-\beta R}$ is given by
$$
\frac{{\rm d}\pi_{-\beta R}}{{\rm d}\pi}(\theta) = \frac{{\rm e}^{-\beta R(\theta)}}{\mathbb{E}_{\vartheta\sim \pi}[{\rm e}^{-\beta R(\vartheta)}] }.
$$
This priori is often refered to as a \textit{localized} prior~\citep{cat2003,cat2007} or as a distribution-dependent prior~\citep{lever2010,lev2013}.
This choice is {\it not} data-dependent, and thus allowed by theory. But in practice, it cannot be used, because $R(\theta) =\mathbb{E}_{(X,Y)\sim P}[\ell(f_\theta(X),Y)] $ is of course unknown (still, we use the prior $\pi_{-\beta R}$ in Section~\ref{section:oracle} below, in theoretical bounds that are not meant to be evaluated on the data). For empirical bounds, Catoni proved that $KL\left(\rho\|\pi_{-\beta R}\right) $ can be upper bounded, with large probability, by $KL\left(\rho\|\pi_{-\xi r}\right) $ for $\xi = \beta / (\lambda+ g(\lambda/n)\lambda^2/n) $, plus some additional terms ($g$ being Bernstein's function defined in Lemma~\ref{lemma:bernstein}). Plugging this result into Theorem~\ref{thm:first:bound}, he obtained the following ``localized bound''~\citep[Lemma 6.2]{cat2003}:

\noindent
\begin{multline*}
\mathbb{P}_{\mathcal{S}}
\Biggl(
\forall\rho\text{, }
\mathbb{E}_{\theta\sim\rho}[ R(\theta)] 
\\
\leq  \frac{ (1-\xi)\mathbb{E}_{\theta\sim\rho}[ r(\theta)] + KL\left(\rho\| \pi_{-\xi \lambda r}\right) + (1+\xi)\log\frac{2}{\delta} }{(1-\xi)\lambda + (1+\xi)g\left( \frac{\lambda}{n} \right) \frac{\lambda^2}{n}}
\Biggr) 
\geq 1-\delta
\end{multline*}
which means that we are allowed to use $\pi_{-\xi r}$, that is data-dependent, as a prior! This bound is a little scary at first, because it depends on many parameters. We will provide simpler localized bounds in Section~\ref{section:oracle} in order to explain their benefits (in particular, it allows to remove some $\log(n)$ terms in the rates of convergences). For now, simply accept that the bound is usually tighter than Theorem~\ref{thm:first:bound}, but in practice, we have to calibrate both $\lambda$ and $\beta$, which makes it a little more difficult to use. Thus, I am not aware of any application of this technique to neural networks, but we will show in Section~\ref{section:oracle} that, used on PAC-Bayes oracle inequalities, it leads to an improvement of the order of the bound. See \citet{cat2003,zha2006,cat2007} for more details on the localization technique.

\citet{dzi2018} proved that any data-dependent prior can actually be used in Seeger's bound, under a differential privacy condition, at the cost of a small modification of the bound.

\begin{theorem}[Theorem 4.2, \citet{dzi2018}]
Assume we have a function $\Pi$ that maps any sample $s=((x_1,y_1),\dots,(x_n,y_n))$ into a prior $\pi=\Pi(\mathcal{S})$. Remind that the data is $\mathcal{S} =((X_1,Y_1),\dots,(X_n,Y_n)) $ and define, for any $i\in\{1,\dots,n\}$, $\mathcal{S}'_i$ a copy of $\mathcal{S}$ where $(X_i,Y_i)$ is replaced by $(X_i',Y_i')\sim P$ independent from $\mathcal{S}$.
Assume that there is an $\eta>0$ such that, for any $i\in\{1,\dots,n\}$, for any measurable set $B$,
$$ \mathbb{P}_{\mathcal{S}} (\Pi(\mathcal{S}) \in B ) \leq {\rm e}^\eta \mathbb{P}_{\mathcal{S}'_i} (\Pi(\mathcal{S}'_i) \in B ) $$
(we say that $\Pi$ is $\eta$-differentially private). Then, for any $\delta> 0$,

\noindent
\begin{multline*}
\mathbb{P}_{\mathcal{S}}
\Biggl(
\forall\rho\text{, }
 kl\left( \mathbb{E}_{\theta\sim\rho}[ R(\theta)] \left\|  \mathbb{E}_{\theta\sim\rho}[ r(\theta)]\right.\right)
\leq \frac{KL(\rho\|\Pi(\mathcal{S})) + \log \frac{4\sqrt{n}}{\delta} }{n}
\\
+ \frac{\eta^2}{2} + \eta\sqrt{\frac{\log\frac{4}{\delta}}{2n}}
\Biggr) \geq 1-\delta.
\end{multline*}
\end{theorem}
For more on PAC-Bayes and differential privacy, see \citet{oneto2020randomized,banerjee2021information}.

\subsection{Comparison of the bounds and tight certificates}

Recently, \citet{per2020} trained neural networks by minimizing PAC-Bayes bounds, on the MNIST and CIFAR-10 datasets. They obtain state of the art test errors ($0.02$ on MNIST), and improve the generalization bounds of \citet{dzi2017}: $0.0279$ on MNIST, a very tight bound!. In order to do so, they build on many ideas of \citet{dzi2017} but also provide a systematic comparison of many of the PAC-Bayes bounds listed so far. In all their experiments, the bound of \citet{thi2017} is the tightest one (Theorem~\ref{thm:thieman:bound} above), taking advantage of situations where the empirical risk is very small. Since then, even tighter results were reported by~\citet{clerico2022conditionally} with a training based on the Gaussian process approximation of neural networks.

All these results are obtained by sample splitting, an approach proposed by \citet{pmlr-v130-karolina-dziugaite21a}: a large part of the dataset is used to learn a prior, centered around the weights of a neural network trained in a non-Bayesian way. The minimization of the PAC-Bayes bound is done on the second part of the dataset. A recurring criticism against this approach is that the prior is so good, that the PAC-Bayes step becomes barely necessary: the posterior learnt is extremely close to the prior. In other words, we could almost learn the weights using the first part of the dataset, and use Hoeffding or Bernstein's inequality (without a union bound) on the second part of the dataset and still obtain a reasonnably tight certificate.

I still believe these results are very motivating for PAC-Bayes as they show that very tight bounds are possible. The numerical comparison of the various bounds in the context of deep-learning is also of interest. Other experimental comparisons of the bounds are available, see for example \citet{foong2021tight} in the small-data regime.

\chapter{PAC-Bayes Oracle Inequalities and Fast Rates}
\label{section:oracle}

As explained in Section~\ref{subsec:twotypes}, empirical PAC-Bayes bounds are very useful as they provide a numerical certificate for randomized estimators or aggregated predictors. But we also mentioned another type of bounds: oracle PAC-Bayes bounds. In this section, we provide examples of PAC-Bayes oracle bounds. Interestingly, the first PAC-Bayes oracle inequality we state below is actually derived from empirical PAC-Bayes inequality.

\section{From Empirical Inequalities to Oracle Inequalities}

As for empirical bounds, we can prove oracle bounds in expectation, and in probability. We will first present a simple version of each. Later, we will focus on bounds in expectation for the sake of simplicity: these bounds are much shorter to prove. But all the results we will prove in expectation have counterparts in probability, see for example the results of \citet{cat2003,cat2007}.

\subsection{Bound in expectation}

We start by a reminder of (the second claim of) Theorem~\ref{thm:first:bound:exp}: for any $\lambda>0$,
\begin{equation*}
 \mathbb{E}_\mathcal{S} \mathbb{E}_{\theta\sim\hat{\rho}_\lambda}[ R(\theta)] \leq  \mathbb{E}_\mathcal{S} \left[ \inf_{\rho\in\mathcal{P}(\theta)}\left\{ \mathbb{E}_{\theta\sim\rho}[ r(\theta)] +  \frac{\lambda C^2}{8n} + \frac{KL(\rho\|\pi) }{\lambda} \right\} \right],
\end{equation*}
where we remind that $\hat{\rho}_\lambda$ is the Gibbs posterior defined in~\eqref{equa:def:gibbs:posterior}. From there, we have the following:
\begin{align*}
 \mathbb{E}_\mathcal{S} \mathbb{E}_{\theta\sim\hat{\rho}_\lambda}[ R(\theta)]
 &
 \leq  \mathbb{E}_\mathcal{S} \left[ \inf_{\rho\in\mathcal{P}(\theta)}\left\{ \mathbb{E}_{\theta\sim\rho}[ r(\theta)] +  \frac{\lambda C^2}{8n} + \frac{KL(\rho\|\pi) }{\lambda} \right\} \right]
 \\
 &
 \leq \inf_{\rho\in\mathcal{P}(\theta)} \left[ \mathbb{E}_\mathcal{S} \left\{ \mathbb{E}_{\theta\sim\rho}[ r(\theta)] +  \frac{\lambda C^2}{8n} + \frac{KL(\rho\|\pi) }{\lambda} \right\} \right]
 \\
 &
 = \inf_{\rho\in\mathcal{P}(\theta)} \left[ \mathbb{E}_\mathcal{S} \left\{ \mathbb{E}_{\theta\sim\rho}[ r(\theta)]\right\} +  \frac{\lambda C^2}{8n} + \frac{KL(\rho\|\pi) }{\lambda}  \right]
 \\
 &
 = \inf_{\rho\in\mathcal{P}(\theta)} \left[  \mathbb{E}_{\theta\sim\rho} \left\{ \mathbb{E}_\mathcal{S} [ r(\theta)]\right\} +  \frac{\lambda C^2}{8n} + \frac{KL(\rho\|\pi) }{\lambda}  \right]
\end{align*}
where we used Tonelli's theorem in the last equality. But, by definition, $\mathbb{E}_\mathcal{S} [ r(\theta)]=R(\theta)$. Thus, we obtain the following theorem.

\begin{theorem}
 \label{thm:oracle:bound:exp}
 For any $\lambda>0$,
 \begin{equation*}
 \mathbb{E}_\mathcal{S} \mathbb{E}_{\theta\sim\hat{\rho}_\lambda}[ R(\theta)] \leq   \inf_{\rho\in\mathcal{P}(\theta)}\left\{ \mathbb{E}_{\theta\sim\rho}[ R(\theta)] +  \frac{\lambda C^2}{8n} + \frac{KL(\rho\|\pi) }{\lambda} \right\} .
\end{equation*}
\end{theorem}

\begin{example}[Finite case, continued]
\label{exm:oracle:finite}
In the context of Example~\ref{exm:first:finite}, that is, ${\rm card}(\Theta)=M<+\infty$, with $\lambda = \sqrt{8n/(C^2\log(M))}$ and $\pi$ uniform on $\Theta$ we obtain the bound:
 \begin{equation*}
 \mathbb{E}_\mathcal{S} \mathbb{E}_{\theta\sim\hat{\rho}_\lambda}[ R(\theta)] \leq   \inf_{\theta\in\Theta} R(\theta) + C\sqrt{\frac{\log(M)}{2n}} .
\end{equation*}
From this, we don't have a numerical certificate on $\mathbb{E}_\mathcal{S} \mathbb{E}_{\theta\sim\hat{\rho}_\lambda}[ R(\theta)] $. But on the other hand, we know that our predictions are the best theoretically possible, up to at most $C\sqrt{\log(M)/2n}$ (such an information is not provided by an empirical PAC-Bayes inequality).
\end{example}

A natural question after Example~\ref{exm:oracle:finite} is: is it possible to improve the rate $1/\sqrt{n}$? Is it possible to ensure that our predictions are the best possible up to a smaller term? The answer is ``no'' in the worst case, but ``yes'' quite often. These faster rates will be the object of the following subsections. But first, as promised, we provide an oracle PAC-Bayes bound in probability.

\subsection{Bound in probability}

As we derived the oracle inequality in expectation of Theorem~\ref{thm:oracle:bound:exp} from the empirical inequality in expectation of Theorem~\ref{thm:first:bound:exp}, we will now use the empirical inequality in probability from Theorem~\ref{thm:first:bound} to prove the following oracle inequality in probability. Note, however, that the proof is slightly more complicated, and that this leads to different (and worse) constants within the bound.

\begin{theorem}
 \label{thm:oracle:bound:pac}
For any $\lambda>0$, for any $\delta\in(0,1)$,
\begin{multline*}
 \mathbb{P}_\mathcal{S}\Biggl(\mathbb{E}_{\theta\sim\hat{\rho}_\lambda}[ R(\theta)] 
 \\
 \leq \inf_{\rho\in\mathcal{P}(\Theta)} \left\{ \mathbb{E}_{\theta\sim\rho}[ R(\theta)] +  \frac{\lambda C^2}{4n} + 2 \frac{KL(\rho\|\pi) + \log\frac{2}{\delta}}{\lambda} \right\} \Biggr)
 \\
 \geq 1-\delta.
\end{multline*}
\end{theorem}

\enlargethispage{-3\baselineskip}
\noindent {\it Proof:} first, apply Theorem~\ref{thm:first:bound} to $\rho = \hat{\rho}_\lambda$, as was done to obtain Corollary~\ref{cor:first:bound}. This gives:
\begin{multline}
\label{equa:proof:thm:oracle:bound:pac:1}
 \mathbb{P}_\mathcal{S}\Biggl( \mathbb{E}_{\theta\sim\hat{\rho}_\lambda}[ R(\theta)]
 \\
 \leq \inf_{\rho\in\mathcal{P}(\Theta)}\left[  \mathbb{E}_{\theta\sim\rho}[ r(\theta)] +  \frac{\lambda C^2}{8n} + \frac{KL(\rho\|\pi) + \log\frac{1}{\delta}}{\lambda} \right]\Biggr)
 \\
 \geq 1-\delta.
\end{multline}
We will now prove the reverse inequality, that is:
\begin{multline}
\label{equa:proof:thm:oracle:bound:pac:2}
 \mathbb{P}_\mathcal{S}\Biggl( \forall \rho\in\mathcal{P}(\Theta)\text{, } \mathbb{E}_{\theta\sim\rho}[ r(\theta)] 
 \\
 \leq   \mathbb{E}_{\theta\sim\rho}[ R(\theta)] +  \frac{\lambda C^2}{8n} + \frac{KL(\rho\|\pi) + \log\frac{1}{\delta}}{\lambda} \Biggr)
 \\
 \geq 1-\delta.
\end{multline}
The proof of~\eqref{equa:proof:thm:oracle:bound:pac:2} is {\it exactly} similar to the proof of Theorem~\ref{thm:first:bound}, except that we replace $U_i$ by $-U_i$. So, the reader who is comfortable enough with this kind of proof can skip this part, or prove~\eqref{equa:proof:thm:oracle:bound:pac:2} as an exercise. Still, we provide a complete proof for the sake of completeness. Fix $\theta\in\Theta$ and apply Hoeffding's inequality with $U_i=\ell_i(\theta)-\mathbb{E}[\ell_i(\theta)]$, and $t=\lambda/n$:
 \begin{equation*}
 \mathbb{E}_\mathcal{S} \left[ {\rm e}^{ \lambda [r(\theta)-R(\theta)] } \right] \leq {\rm e}^{\frac{\lambda^2 C^2}{ 8 n}}.
 \end{equation*}
Integrate this bound with respect to $\pi$:
 \begin{equation*}
 \mathbb{E}_{\theta\sim\pi} \mathbb{E}_\mathcal{S} \left[ {\rm e}^{ \lambda [r(\theta)-R(\theta)] } \right] \leq {\rm e}^{\frac{\lambda^2 C^2}{ 8 n}}.
 \end{equation*}
Apply Tonelli:
 \begin{equation*}
 \mathbb{E}_\mathcal{S}  \mathbb{E}_{\theta\sim\pi} \left[ {\rm e}^{ \lambda [r(\theta)-R(\theta)] } \right] \leq {\rm e}^{\frac{\lambda^2 C^2}{ 8 n}}
 \end{equation*}
 and then Donsker and Varadhan's variational formula (Lemma~\ref{lemma:dv}):
 \begin{equation*}
 \mathbb{E}_\mathcal{S} \left[ {\rm e}^{ \sup_{\rho\in\mathcal{P}(\Theta)} \lambda \mathbb{E}_{\theta\sim\rho}[r(\theta)-R(\theta)] -KL(\rho\|\pi) } \right] \leq {\rm e}^{\frac{\lambda^2 C^2}{ 8 n}}.
 \end{equation*} 
 Rearranging terms:
 \begin{equation*}
 \mathbb{E}_\mathcal{S} \left[ {\rm e}^{ \sup_{\rho\in\mathcal{P}(\Theta)} \lambda \mathbb{E}_{\theta\sim\rho}[r(\theta)-R(\theta)] -KL(\rho\|\pi)-\frac{\lambda^2 C^2}{ 8 n} } \right] \leq 1.
 \end{equation*}
Chernoff bound gives: \enlargethispage{-3\baselineskip}
 \begin{equation*}
 \mathbb{P}_\mathcal{S} \left[ \sup_{\rho\in\mathcal{P}(\Theta)} \lambda \mathbb{E}_{\theta\sim\rho}[r(\theta)-R(\theta)] -KL(\rho\|\pi)-\frac{\lambda^2 C^2}{ 8 n}  > \log\frac{1}{\delta} \right] \leq \delta.
 \end{equation*} 
 Rearranging terms:
 \begin{multline*}
 \mathbb{P}_\mathcal{S} \Biggl[ \exists \rho\in\mathcal{P}(\Theta) \text{, } \mathbb{E}_{\theta\sim\rho}[ r(\theta)] > \mathbb{E}_{\theta\sim\rho}[ R(\theta)] 
 \\
 +  \frac{\lambda C^2}{8n} + \frac{KL(\rho\|\pi) + \log\frac{1}{\delta}}{\lambda} \Biggr] \leq \delta.
 \end{multline*}
 Take the complement to get~\eqref{equa:proof:thm:oracle:bound:pac:2}.
 
 Consider now~\eqref{equa:proof:thm:oracle:bound:pac:1} and~\eqref{equa:proof:thm:oracle:bound:pac:2}. A union bound gives:
\begin{multline*}
\mathbb{P}_{\mathcal{S}}\left(
\begin{array}{c}
 \mathbb{E}_{\theta\sim\hat{\rho}_\lambda}[ R(\theta)] \leq \inf_{\rho\in\mathcal{P}(\Theta)}\left[  \mathbb{E}_{\theta\sim\rho}[ r(\theta)] +  \frac{\lambda C^2}{8n} + \frac{KL(\rho\|\pi) + \log\frac{1}{\delta}}{\lambda} \right]
 \\
 \text{ and simultaneously }
 \\
 \forall \rho\in\mathcal{P}(\Theta)\text{, } \mathbb{E}_{\theta\sim\rho}[ r(\theta)] \leq   \mathbb{E}_{\theta\sim\rho}[ R(\theta)] +  \frac{\lambda C^2}{8n} + \frac{KL(\rho\|\pi) + \log\frac{1}{\delta}}{\lambda}
\end{array}
\right)
\\
\geq 1-2\delta
\end{multline*}
Plug the upper bound on $\mathbb{E}_{\theta\sim\rho}[ r(\theta)] $ from the second line into the first line to get:
\begin{multline*}
\mathbb{P}_{\mathcal{S}}\Biggl(
 \mathbb{E}_{\theta\sim\hat{\rho}_\lambda}[ R(\theta)] \leq \inf_{\rho\in\mathcal{P}(\Theta)}\Biggl[  \mathbb{E}_{\theta\sim\rho}[ r(\theta)]
 \\
 +  2 \frac{\lambda C^2}{8n} + 2 \frac{KL(\rho\|\pi) + \log\frac{1}{\delta}}{\lambda} \Biggr]
\Biggr)
\geq 1-2\delta.
\end{multline*}
Just replace $\delta$ by $\delta/2$ to get the statement of the theorem.
$\square$

\section{Bernstein Assumption and Fast Rates}

As mentioned above, the rate $1/\sqrt{n}$ that we have obtained in many PAC-Bayes bounds seen so far is not always the tightest possible. Actually, this can be seen in Tolstikhin and Seldin's bound (Theorem~\ref{thm:tol:bound}): there, if $\mathbb{E}_{\theta\sim\rho}[ r(\theta)] =0$ for some $\rho$, then the bound is actually in $1/n$.

It appears that rates in $1/n$ are possible in a more general setting, under an assumption often refered to as Bernstein assumption. This is well known for (``non Bayesian'') PAC bounds~\citep{bartlett2003large}, we will show that this is also true with PAC-Bayes bounds.
\begin{definition}
\label{dfn:bernstein}
 From now, we will let $\theta^*$ denote a minimizer of $R$ when it exists:
 $$ R(\theta^*) = \min_{\theta\in\Theta} R(\theta). $$
 When such a $\theta^*$ exists, and when there is a constant $K$ such that, for any $\theta\in\Theta$,
 \begin{equation*}
  \mathbb{E}_{\mathcal{S}}\left\{ \left[  \ell_i(\theta) - \ell_i(\theta^*) \right]^2 \right\}
  \leq K [R(\theta)-R(\theta^*) ]
 \end{equation*}
 we say that Bernstein assumption is satisfied with constant $K$.
\end{definition}

PAC-Bayes oracle bounds based using explicitly Bernstein's assumption were proven by \citet{cat2003,zha2006,cat2007,grun2020,grunwald2021pac}. Before we state such a bound, let us explore situations where this assumption is satisfied.

\begin{example}[Classification without noise]
Consider classification with the $0-1$ loss: $\ell_i(\theta)=\mathbf{1}(Y_i\neq f_{\theta}(X_i))$. If the optimal classifier does not make any mistake, that is, if $ R(\theta^*)=0$, we have necessarily $\ell_i(\theta^*)=0$ almost surely. We refer to this situation as ``classification without noise''. In this case, we have obviously:
 \begin{align*}
  \mathbb{E}_{\mathcal{S}}\left\{ \left[  \ell_i(\theta) - \ell_i(\theta^*) \right]^2 \right\}
  &  = \mathbb{E}_{\mathcal{S}}\left\{ \left[ \mathbf{1}(Y_i\neq f_{\theta}(X_i)) - 0 \right]^2 \right\}
  \\
  & = \mathbb{E}_{\mathcal{S}}\left\{  \mathbf{1}(Y_i\neq f_{\theta}(X_i)) \right\}
  \\
  & = R(\theta)
  \\
  & = K[R(\theta)-R(\theta^*)]
  \end{align*}
 if we put $K=1$. So, Bernstein assumption is satisfied with constant $K=1$. Actually, this can be extended beyond the $0-1$ loss: for any loss $\ell$ with values in $[0,C]$, if $R(\theta^*)=0$, then Bernstein assumption is satisfied with constant $K=C$.
\end{example}

\begin{example}[Mammen and Tsybakov margin assumption]
More generally, still in classification with the $0-1$ loss, consider the function
$$ \eta(x) = \mathbb{E}_{\mathcal{S}}(Y_i|X_i=x). $$
\citet{mammen1999smooth} proved that, if $|\eta(X_i)-1/2|\geq \tau $ almost surely for some $\tau>0$, then Bernstein assumption holds for some $K$ that depends on $\tau$. The case $\tau=1/2$ leads back to the previous example (noiseless classification), but $0<\tau<1/2$ is a more general assumption.
\end{example}

\begin{example}[Lipschitz and strongly convex loss function]
Assume that $\Theta$ is convex. Assume that there are function $d_i:\Theta^2\rightarrow \mathbb{R}_+$, where each $d_i$ might depend of $(X_i,Y_i)$, and assume that $\ell_i $ satisfies:
\begin{equation}
\label{bernstein:strcvx}
 \forall \theta \in\Theta \text{, }
 \frac{\ell_i(\theta)+\ell_i(\theta^*)}{2} - \ell_i\left(\frac{\theta + \theta^*}{2} \right) \geq \frac{1}{8\alpha} d_i^2(\theta,\theta^*)
\end{equation}
and
\begin{equation}
\label{bernstein:lipschitz}
 \forall \theta\in\Theta\text{, } |\ell_i(\theta)-\ell_i(\theta^*)| \leq L d_i(\theta,\theta^*).
\end{equation}
In the special case where $d_i(\theta,\theta')$ is a metric on $\Theta$,~\eqref{bernstein:strcvx} will be satisfied if the loss is $\alpha$-strongly convex in $\theta$, and~\eqref{bernstein:lipschitz} will be satisfied if the loss is $L$-Lipschitz in $\theta$ with respect to $d_i$. For example, when using the quadratic loss $\ell_i(\theta) = (Y_i - \left<\theta,X_i\right>)^2$,~\eqref{bernstein:strcvx} holds with $d_i(\theta,\theta^*) = \left<X_i,\theta-\theta^*\right>^2$ and $\alpha = 1/2$ (it is actually an equality in this case, and~\eqref{bernstein:lipschitz} holds if $X_i$, $Y_i$ and $\Theta$ are bounded).

\citet{bartlett2003large} proved that, under these assumptions, Bernstein assumption is satisfied with constant $K=4L^2 \alpha$. The proof is so luminous than I cannot resist giving it:
 \begin{align*}
  \mathbb{E}_{\mathcal{S}} & \left\{ \left[  \ell_i(\theta) - \ell_i(\theta^*) \right]^2 \right\}
  \\
  & \leq   L^2 \mathbb{E}_{\mathcal{S}}\left\{ d_i(\theta,\theta^*)^2 \right\} \text{ by \eqref{bernstein:lipschitz}}
  \\
  & \leq 8 L^2 \alpha \mathbb{E}_{\mathcal{S}}\left\{ \frac{\ell_i(\theta)+\ell_i(\theta^*)}{2} - \ell_i\left(\frac{\theta + \theta^*}{2} \right) \right\} \text{ by \eqref{bernstein:strcvx}}
  \\
  & = 8 L^2 \alpha \left[ \frac{R(\theta)+R(\theta^*)}{2} - R\left(\frac{\theta + \theta^*}{2} \right) \right]
  \\
  & \leq 8 L^2 \alpha \left[ \frac{R(\theta)+R(\theta^*)}{2} - R\left(\theta^* \right) \right]
\end{align*}
where in the last equation, we used $R(\theta^*) \leq R\left(\frac{\theta + \theta^*}{2} \right)$. Thus:
\begin{equation*}
   \mathbb{E}_{\mathcal{S}}\left\{ \left[  \ell_i(\theta) - \ell_i(\theta^*) \right]^2 \right\}
   \leq 4 L^2 \alpha \left[ R(\theta) - R\left(\theta^* \right) \right]
\end{equation*}
and thus Bernstein assumption is satisfied with constant $K=4L^2 \alpha$.
\end{example}

\begin{theorem}
\label{thm:oracle:bound}
Assume Bernstein assumption is satisfied with some constant $K>0$. Take $\lambda= n/\max(2K,C)$, we have:
\begin{multline*}
\mathbb{E}_\mathcal{S} \mathbb{E}_{\theta\sim\hat{\rho}_{\lambda}} [R(\theta) ]-R(\theta^*) 
  \\
  \leq 2  \inf_{\rho\in\mathcal{P}(\Theta)}  \left\{  \mathbb{E}_{\theta\sim\rho} [R(\theta) ]-R(\theta^*) + \frac{\max(2K,C) KL(\rho\| \pi) }{n} \right\}.
\end{multline*}
\end{theorem}

We postpone the proof to the end of this section (page~\pageref{lemma:bernstein}), and the applications to Section~\ref{subsec:appli:oracle}. First, a quick explanation on how we will use this bound: we only have to find a $\rho$ such that $ \mathbb{E}_{\theta\sim\rho} [R(\theta) ] \simeq R(\theta^*)$ to obtain:
\begin{equation*}
\mathbb{E}_\mathcal{S} \mathbb{E}_{\theta\sim\hat{\rho}_{\lambda}} [R(\theta) ]
   \lesssim R(\theta^*) + \frac{2 \max(2K,C) KL(\rho\| \pi) }{n} ,
\end{equation*}
hence the rate in $1/n$. We will provide more accurate statements in Subsection~\ref{subsec:appli:oracle}. The quantity $R(\theta)-R(\theta^*)$ is known as the ``excess risk'' of $\theta$. Thus, Theorem~\ref{thm:oracle:bound} is often refered to as ``excess risk PAC-Bayes bound''.

\begin{remark}
There is a more general version of Bernstein condition: where there are constants $K>0$ and $\kappa\in[1,+\infty)$ such that, for any $\theta\in\Theta$,
 \begin{equation*}
  \mathbb{E}_{\mathcal{S}}\left\{ \left[  \ell_i(\theta) - \ell_i(\theta^*) \right]^2 \right\}
  \leq K [R(\theta)-R(\theta^*) ]^{\frac{1}{\kappa}}
 \end{equation*}
 we say that Bernstein assumption is satisfied with constants $(K,\kappa)$. We will not study the general case here, but we mention that, in the case of classification, this can also be interpreted in terms of margin~\citep{mammen1999smooth}. Under such an assumption, some oracle PAC-Bayes inequalities for classification are proven by \citet[Corollary 1.4.7 page 40]{cat2007} that leads to rates in $1/n^{\kappa/(2\kappa-1)}$. These results were later extended to general losses~\citep{alq2008}. These rates are known to be optimal in the minimax sense~\citep{lecue2007aggregation}. Finally, for recent results and a comparison of all the type of conditions leading to fast rates in learning theory (including situations with unbounded losses), see \citet{grun2020}.
\end{remark}

\begin{remark}
All the PAC-Bayes inequalities seen before Section~\ref{section:oracle} were empirical. Most of them led to the rate $1/\sqrt{n}$, except when the empirical error is close to zero, where we obtained the rate $1/n$. In this section, we built:
\begin{itemize}
 \item an oracle inequality with rates $1/\sqrt{n}$. Note that an empirical inequality was part of the proof.
 \item an oracle inequality with rate $1/n$. It is important to note that the proof we will propose {\it does not directly involve any empirical inequality}. (The reader might remark that~\eqref{equa:proof:almost:empirical} in the proof below is almost an empirical inequality, but the term $r(\theta^*)$ is not empirical as it depends on the unknown $\theta^*$).
\end{itemize}
It is thus natural to ask: are there empirical inequalities leading to rates in $1/n$ or $1/n^{\kappa/(2\kappa-1)}$ beyond the noiseless case? The answer is ``yes'' for ``non-Bayesian'' PAC  bounds~\citep{bartlett2006empirical}, based on Rademacher complexity: there are empirical bounds on $R(\hat{\theta}_{{\rm ERM}})-R(\theta^*)$. In the PAC-Bayesian case, it is a little more complicated (unless one uses the bound to control the risk of a non-Bayesian estimator such as the ERM). This is discussed by \citet{grunwald2021pac}.
\end{remark}

We will now prove Theorem~\ref{thm:oracle:bound}. The proof relies on Bernstein's inequality (Lemma~\ref{lemma:bernstein}).

\noindent {\it Proof of Theorem~\ref{thm:oracle:bound}:}
We follow the general proof scheme for PAC-Bayes bounds, with some important differences. Fist, Hoeffding's inequality will be replaced by Bernstein's inequality. But another very important point is to use the inequality on the ``relative losses'' $ \ell_i(\theta^*) - \ell_{i}(\theta) $ instead of the loss $\ell_i(\theta)$ (for this reason, excess risk bounds are also sometimes called ``relative bounds''). This is to ensure that we can use Bernstein condition. So, let us fix $\theta\in\Theta$ and apply Lemma~\ref{lemma:bernstein} to $U_i= \ell_i(\theta^*) - \ell_{i}(\theta)  $. As $\mathbb{E}(U_i)=R(\theta^*)-R(\theta)$, we obtain:
\begin{equation*}
 \mathbb{E}_{\mathcal{S}}\left( {\rm e}^{ t n [R(\theta) - R(\theta^*) -r(\theta) + r(\theta^*)] } \right)
\leq {\rm e}^{ g\left(C t\right) n t^2  {\rm Var}_{\mathcal{S}}(U_i)}.
\end{equation*}
Put $\lambda = t n $ and note that
\begin{align*}
{\rm Var}_{\mathcal{S}}(U_i)
& \leq \mathbb{E}_{\mathcal{S}}(U_i^2)
\\
& = \mathbb{E}_{\mathcal{S}}\left\{[\ell_i(\theta^*) - \ell_{i}(\theta)]^2 \right\}
\\
& \leq K \left[ R(\theta) - R(\theta^*) \right]
\end{align*}
thanks to Bernstein condition. Thus:
\begin{equation*}
 \mathbb{E}_{\mathcal{S}}\left( {\rm e}^{\lambda [R(\theta) - R(\theta^*) -r(\theta) + r(\theta^*)] } \right)
\leq {\rm e}^{ g\left(\frac{\lambda C}{n}\right) \frac{\lambda^2}{n} K \left[ R(\theta) - R(\theta^*) \right] }.
\end{equation*}
Rearranging terms:
\begin{equation}
\label{equa:proof:thm:oracle}
 \mathbb{E}_{\mathcal{S}}\left( {\rm e}^{\lambda \left\{ \left[ 1- K g\left(\frac{\lambda C}{n}\right) \frac{\lambda}{n} \right]\left[ R(\theta) - R(\theta^*)\right] -r(\theta) + r(\theta^*)\right\} } \right)
\leq 1.
\end{equation}
The next steps are now routine: we integrate $\theta$ with respect to $\pi$ and apply Tonelli's theorem and Donsker and Varadhan's variational formula to get:
\begin{equation*}
 \mathbb{E}_{\mathcal{S}}\left( {\rm e}^{\lambda \sup_{\rho\in\mathcal{P}(\Theta)} \left( \mathbb{E}_{\theta\sim\rho} \left\{ \left[ 1- K g\left(\frac{\lambda C}{n}\right) \frac{\lambda}{n} \right]\left[ R(\theta) - R(\theta^*)\right] -r(\theta) + r(\theta^*)\right\} -KL(\rho\|  \pi) \right)} \right)
\leq 1.
\end{equation*}
In particular for $\rho=\hat{\rho}_\lambda$ the Gibbs posterior of~\eqref{equa:def:gibbs:posterior}, we have, using Jensen and rearranging terms:
\begin{multline*}
  \left[ 1- K g\left(\frac{\lambda C}{n}\right) \frac{\lambda}{n} \right] \left\{ \mathbb{E}_\mathcal{S} \mathbb{E}_{\theta\sim\hat{\rho}_{\lambda}} [R(\theta) ]-R(\theta^*) \right\}
 \\
 \leq \mathbb{E}_\mathcal{S}  \left[  \mathbb{E}_{\theta\sim\hat{\rho}_{\lambda}} [r(\theta) ]-r(\theta^*) + \frac{KL(\hat{\rho}_\lambda\| \pi)}{\lambda} \right].
\end{multline*}
From now we assume that $\lambda$ is such that $\left[ 1- K g\left(\frac{\lambda C}{n}\right) \frac{\lambda}{n} \right]>0$, thus
\begin{equation*}
\mathbb{E}_\mathcal{S} \mathbb{E}_{\theta\sim\hat{\rho}_{\lambda}} [R(\theta) ]-R(\theta^*) 
 \leq \frac{ \mathbb{E}_\mathcal{S}  \left\{  \mathbb{E}_{\theta\sim\hat{\rho}_{\lambda}} [r(\theta) ]-r(\theta^*) + \frac{KL(\hat{\rho}_\lambda\| \pi)}{\lambda} \right\}}{1- K g\left(\frac{\lambda C}{n}\right) \frac{\lambda}{n}}.
\end{equation*}
In particular, take $\lambda = n/\max(2K,C)$. We can check that: $\lambda \leq n/(2K) \Rightarrow K \lambda / n \leq 1/2$ and $\lambda \leq n/C \Rightarrow g(\lambda C/n) \leq g(1) \leq 1 $, so
$$  K g\left(\frac{\lambda C}{n}\right) \frac{\lambda}{n} \leq \frac{1}{2} $$
and thus
\begin{equation}
\label{equa:proof:almost:empirical}
\mathbb{E}_\mathcal{S} \mathbb{E}_{\theta\sim\hat{\rho}_{\lambda}} [R(\theta) ]-R(\theta^*) 
 \leq 2\mathbb{E}_\mathcal{S}  \left\{  \mathbb{E}_{\theta\sim\hat{\rho}_{\lambda}} [r(\theta) ]-r(\theta^*) + \frac{KL(\hat{\rho}_\lambda\| \pi)}{\lambda} \right\}.
\end{equation}
Finally, note that $\hat{\rho}_\lambda$ minimizes the quantity in the expectation in the right-hand side. Thus,~\eqref{equa:proof:almost:empirical} can be rewritten as
\begin{align*}
\mathbb{E}_\mathcal{S} & \mathbb{E}_{\theta\sim\hat{\rho}_{\lambda}} [R(\theta) ]
 -R(\theta^*) 
 \\
 &  \leq 2 \mathbb{E}_\mathcal{S}  \inf_{\rho\in\mathcal{P}(\Theta)}  \left\{  \mathbb{E}_{\theta\sim\rho} [r(\theta) ]-r(\theta^*) + \frac{\max(2K,C) KL(\rho\|\pi) }{n}\right\}
 \\
 & \leq  2  \inf_{\rho\in\mathcal{P}(\Theta)}  \mathbb{E}_\mathcal{S} \left\{  \mathbb{E}_{\theta\sim\rho} [r(\theta) ]-r(\theta^*) +\frac{\max(2K,C) KL(\rho\| \pi) }{n}\right\}
\\
& = 2  \inf_{\rho\in\mathcal{P}(\Theta)}  \left\{  \mathbb{E}_{\theta\sim\rho} [R(\theta) ]-R(\theta^*) + \frac{\max(2K,C) KL(\rho\| \pi) }{n} \right\}. \text{ }\square 
\end{align*}

\section{Applications of Theorem~\ref{thm:oracle:bound}}
\label{subsec:appli:oracle}

\begin{example}[Finite set of predictors]
 We come back to the setting of Example~\ref{exm:first:finite}: ${\rm card}(\Theta)=M$ and $\pi$ is the uniform distribution over $\Theta$. Assuming Bernstein condition holds with constant $K$, we apply Theorem~\ref{thm:oracle:bound} and, as was done in Example~\ref{exm:first:finite}, we restrict the supremum to $\rho\in\{\delta_\theta,\theta\in\Theta\}$. This gives, for $\lambda=n/\max(2K,C)$,
 
 \vspace{-0.8\baselineskip}
 \small
 \begin{equation*}
 \mathbb{E}_\mathcal{S} \mathbb{E}_{\theta\sim\hat{\rho}_{\lambda}} [R(\theta) ]-R(\theta^*) 
   \leq 2  \inf_{\theta\in\Theta}  \left\{ R(\theta)-R(\theta^*) + \frac{\max(2K,C) \log(M) }{n} \right\}.
\end{equation*}\normalsize
In particular, for $\theta=\theta^*$, this becomes:
 \begin{equation*}
 \mathbb{E}_\mathcal{S} \mathbb{E}_{\theta\sim\hat{\rho}_{\lambda}} [R(\theta) ] \leq R(\theta^*) 
  + \frac{2 \max(2K,C) \log(M) }{n}.
\end{equation*}
Note that the rate $\sqrt{\log(M)/n}$ from Example~\ref{exm:first:finite} becomes $\log(M)/n$ under Bernstein assumption.
\end{example}

\begin{example}[Lipschitz loss and Gaussian priors]
\label{exm:lipschitz:gauss:continued}
We now tackle the setting of Example~\ref{exm:lipschitz:gauss} under Bernstein assumption with constant $K$. Let us remind that $\Theta=\mathbb{R}^d$, $\theta\mapsto\ell(f_\theta(x),y)$ is $L$-Lipschitz for any $(x,y)$, and $\pi=\mathcal{N}(0,\sigma^2 I_d)$. We apply Theorem~\ref{thm:oracle:bound}, for $\lambda= n/\max(2K,C)$:
\begin{multline*}
 \mathbb{E}_\mathcal{S}
 \mathbb{E}_{\theta\sim\hat{\rho}_\lambda}[ R(\theta)] \leq R(\theta^*) 
 \\
 + 2 \inf_{
 \begin{tiny}
 \begin{array}{c}
 \rho=\mathcal{N}(m,s^2 I_d)
 \\
 m\in\mathbb{R}^d, s>0
 \end{array}
 \end{tiny}
 }\left[ R(\theta) - R(\theta^*) + \frac{\max(2K,C) KL(\rho\|\pi) }{n} \right].
\end{multline*}
Following the same derivations as in Example~\ref{exm:lipschitz:gauss}, with $m=\theta^*$,
\begin{multline*}
 \mathbb{E}_\mathcal{S}
 \mathbb{E}_{\theta\sim\hat{\rho}_\lambda}[ R(\theta)] \leq R(\theta^*) 
 \\
 + 2 \inf_{
s>0
 }\left[ Ls\sqrt{d} + \frac{\max(2K,C) \left\{  \frac{\|\theta^*\|^2}{2\sigma^2} + \frac{d}{2}\left[\frac{s^2}{\sigma^2} + \log\frac{\sigma^2}{s^2} -1 \right] \right\}}{n} \right].
\end{multline*}
Here again, we would seek for the exact optimizer in $s$, but for example $s=\sqrt{d}/n$ we obtain:
\begin{multline*}
 \mathbb{E}_\mathcal{S}
 \mathbb{E}_{\theta\sim\hat{\rho}_\lambda}[ R(\theta)] 
\\
\leq R(\theta^*) + 2 \left[ \frac{Ld}{n} + \frac{\max(2K,C) \left\{  \frac{\|\theta^*\|^2}{2\sigma^2} + \frac{d}{2}\left[\frac{d}{n^2\sigma^2} + \log\frac{\sigma^2 n^2}{d}\right] \right\}}{n} \right]
\end{multline*}
that is
\begin{align*}
 \mathbb{E}_\mathcal{S}
 \mathbb{E}_{\theta\sim\hat{\rho}_\lambda}[ R(\theta)] 
 & \leq R(\theta^*) + \frac{2 d}{n} \left[\frac{\max(2K,C)}{2} \log\left(\frac{\sigma^2 n^2}{d}\right) + L + \frac{d}{2n\sigma^2} \right] 
 \\
 & \qquad + \frac{\max(2K,C)\|\theta^*\|^2}{n\sigma^2}
 \\
 & =  R(\theta^*) + \mathcal{O}\left(\frac{d\log(n)}{n} \right).
\end{align*}
\end{example}

\begin{example}[Lipschitz loss and Uniform priors]
\label{exm:lipschitz:uniform}
We propose a variant of the previous example, with a different prior. We still assume Bernstein assumption with constant $K$, $\theta\mapsto\ell(f_\theta(x),y)$ is $L$-Lipschitz for any $(x,y)$, and this time $\Theta=\{\theta\in\mathbb{R}^d:\|\theta\|\leq B\}$ and $\pi$ is uniform on $\Theta$. We apply Theorem~\ref{thm:oracle:bound} with $\lambda= n/\max(2K,C)$ and restrict the infimum to $\rho = U(\theta_0,s)$ the uniform distribution on $\{\theta:\|\theta-\theta_0\|\leq s\}=\mathcal{B}(\theta_0,s)$. We obtain:
\begin{multline*}
 \mathbb{E}_\mathcal{S}
 \mathbb{E}_{\theta\sim\hat{\rho}_\lambda}[ R(\theta)] \leq R(\theta^*)
 \\
 + 2 \inf_{
 \begin{tiny}
 \begin{array}{c}
 \rho=U(\theta_0,s)
 \\
 \theta_0\in\mathbb{R}^d, s>0
 \end{array}
 \end{tiny}
 }\left[ R(\theta) - R(\theta^*) + \frac{\max(2K,C) KL(\rho\|\pi) }{n} \right].
\end{multline*}
For any $s>0$, there exists $\theta_0\in\Theta$ such that $ \theta_* \in \mathcal{B}(\theta_0,s) \subset \Theta$  and we have:
$$
\mathbb{E}_{\theta\sim U(\theta_0,s)}[ R(\theta)-R(\theta^*) ] \leq Ls,
$$
so
\begin{equation*}
 \mathbb{E}_\mathcal{S}
 \mathbb{E}_{\theta\sim\hat{\rho}_\lambda}[ R(\theta)] \leq R(\theta^*)  + 2 \inf_{
s>0
 }\left[ Ls + \frac{\max(2K,C) d\log\left(\frac{B}{s}\right) }{n} \right].
\end{equation*}
The minimum of the right-hand side is exactly reached for $s = \frac{\max(2K,C) d}{Ln}$ and we obtain, for $n$ large enough (in order to ensure that $s\leq B$):
\begin{equation*}
 \mathbb{E}_\mathcal{S}
 \mathbb{E}_{\theta\sim\hat{\rho}_\lambda}[ R(\theta)] \leq R(\theta^*)  +  \frac{2 \max(2K,C) d\log\left( \frac{{\rm e} B L n}{\max(2K,C) }\right) }{n} .
\end{equation*}
\end{example}

We will end this section by a (non-exhaustive!) list of more sophisticated models where a rate of convergence was derived thanks to an oracle PAC-Bayes inequality:
\begin{itemize}
 \item model selection: for classification, see \citet[Chapter 5]{cat2004}, and \citet{cat2007}. The minimization procedure in Example~\ref{exm:model:selection} above is not always optimal. A selection based on Lepski's procedure~\citep{lepski1} and PAC-Bayes bounds is used by \citet{cat2007} for classification and by \citet{alq2008} for general losses.
 \item density estimation~\citep[Chapter 4]{cat2004}.
 \item scoring/ranking~\citep{rid2014}.
 \item least-square regression~\citep[Chapter 5]{cat2004} and robust variants~\citep{aud2011,catoni2017dimension}.
 \item sparse linear regression: \citet{dal2008} prove a rate of convergence similar to the one of the LASSO for the Gibbs posterior, under more general assumptions. \citet{dal2012b,alq2011,dal2018} provided many variants and improvements, see also \citet{luu2019pac} for group-sparsity.
 \item single-index regression, in small dimension~\citep{gaiffas2007optimal}, and in high-dimension, with sparsity~\citep{alquier2013sparse}.
 \item additive non-parametric regression~\citep{gue2013}.
 \item matrix regression~\citep{suz2012,alquier2013bayesian,dal2018,DalEWA2018,mai2023bilinear}.
 \item matrix completion: continuous case~\citep{mai2015,maithesis,mai2021bayesian} and binary case~\citep{cottet20181}; more generally tensor completion can be tackled with related techniques~\citep{pmlr-v37-suzuki15}.
 \item quantum tomography~\citep{mai2017}.
 \item deep learning~\citep{che2020b,trabs2022}.
 \item estimation of the Gram matrix for PCA~\citep{catoni2017dimension,zhivotovskiy2021dimension} and kernel-PCA~\citep{giu2018,haddouche2020upper}.
\end{itemize}

\section{Dimension and Rate of Convergence}
\label{subsec:rate:prior:mass}

Let us recap the examples  seen so far (Sections~\ref{section:first:step} and~\ref{section:oracle}). In each case, we were able to prove a result of the form:
\begin{equation*}
  \mathbb{E}_\mathcal{S}\mathbb{E}_{\theta\sim\hat{\rho}_\lambda}[ R(\theta)] \leq R(\theta^*) + {\rm rate}_n(\pi) \text{ where } {\rm rate}_n(\pi)\xrightarrow[n\rightarrow\infty]{} 0
\end{equation*}
for an adequate choice of $\lambda>0$. The way ${\rm rate}_n(\pi)$ depends on $\Theta$ characterizes the difficulty of learning predictors in $\Theta$ when using the prior $\pi$: it is similar to other approaches in learning theory, where the learning rate depends on the ``complexity of $\Theta$''. More precisely (we remind that all the results seen so far are for a bounded loss function):
\begin{itemize}
 \item when $\Theta$ is finite and $\pi$ is uniform, ${\rm rate}_n(\pi)$ is in $\sqrt{\log(M)/n}$ in general, and in $\log(M)/n$ under Bernstein condition.
 \item when $\Theta=\mathbb{R}^d$ and $\pi=\mathcal{N}(0,\sigma^2 I_d)$, it is in $\sqrt{[\|\theta^*\|^2 + d\log(n)]/n}$ in general, and in $[\|\theta^*\|^2 + d\log(n)]/n$ under Bernstein condition.
 \item still  when $\Theta=\mathbb{R}^d$, \citet{dal2008} proposed a heavy-tailed prior $\pi$ (multivariate Student). Then, ${\rm rate}_n(\pi)$ is in $\sqrt{[\log\|\theta^*\| + d\log(n)]/n}$ in general, and in $[\log\|\theta^*\| + d\log(n)]/n$ under Bernstein condition (this is left as an exercise to the reader).
 \item when $\Theta$ is a compact subset of $\mathbb{R}^d$ and $\pi$ is uniform, ${\rm rate}_n(\pi)$ is in $\sqrt{d\log(n)/n}$ in general, and in $ d\log(n)/n$ under Bernstein condition.
\end{itemize}
The calculations leading to these results are in Examples~\ref{exm:lipschitz:gauss},~\ref{exm:lipschitz:gauss:continued} and~\ref{exm:lipschitz:uniform}. A closer look at these examples reveals a common strategy: we assumed conditions ensuring that it is possible to write
\begin{equation}
\inf_{\rho\in \mathcal{P}(\Theta)} \left[ \mathbb{E}_{\theta\sim\rho}[R(\theta)-R(\theta^*)] + \frac{KL(\rho\|\pi)}{\beta} \right]  \leq \frac{d}{\beta} \log\left( \frac{\beta}{c} \right)
\label{equa:dimension:first}
\end{equation}
for some constants $c$ and $d$ ($d$ being actually the dimension of the model). Then, we can plug this inequality into Theorem~\ref{thm:oracle:bound:exp}, or when Bernstein assumption is satisfied, into Theorem~\ref{thm:oracle:bound}, to obtain a rate of convergence. Let us now turn this into a formal statement under Bernstein assumption (the case without Bernstein assumption is left as an exercise to the reader).
\begin{theorem}
 \label{thm:oracle:bound:dimension}
Assume that there are constants $\beta_0$, $c_\pi$ and $d_\pi$ such that, for any $\beta\geq \beta_0$,
\begin{equation}
\label{equa:dimension:first:prime}
\beta \inf_{\rho\in \mathcal{P}(\Theta)} \left[ \mathbb{E}_{\theta\sim\rho}[R(\theta)-R(\theta^*)] + \frac{KL(\rho\|\pi)}{\beta} \right]  \leq d_\pi \log\left( \frac{\beta}{c_\pi} \right) .
\end{equation}
Under Bernstein condition with constant $K>0$, with $\lambda = n/\max(2K,$ $C)$, we have as soon as $\lambda \geq \beta_0$,
\begin{equation*}
\mathbb{E}_\mathcal{S} \mathbb{E}_{\theta\sim\hat{\rho}_{\lambda}} [R(\theta) ]
 \leq
 R(\theta^*) 
+\frac{2 d_\pi \max(2K,C) \log\left( \frac{n  }{c_\pi \max(2K,C)} \right) }{n}.
\end{equation*}
\end{theorem}
The proof is direct, by pluging~\eqref{equa:dimension:first:prime} into Theorem~\ref{thm:oracle:bound} with $\lambda=\beta$.

Let us finally dicsuss a connection between the assumption in~\eqref{equa:dimension:first:prime} and other classical assumptions in Bayesian statistics and machine learning. A first direct remark, based on Lemma~\ref{lemma:dv}, is that~\eqref{equa:dimension:first:prime} is equivalent to
$$ f(\beta) := -\log \mathbb{E}_{\theta\sim\pi} \left\{ {\rm e}^{-\beta[R(\theta)-R(\theta^*)]}
 \right\} \leq  d_\pi \log\left( \frac{\beta}{c_\pi} \right). $$
 If the inequality was actually an equality, we would have $ f'(\beta) =  d_\pi/ \beta$, that is, $\beta f'(\beta) = d_\pi$. Actually, $ f'(\beta) =  \mathbb{E}_{\theta\sim \pi_{-\beta R}} \left[ R(\theta)-R(\theta^*)\right] $ (this is proven in the proof of Lemma~\ref{lemma:oracle:bound:dimension} below).
 This motivated the following definition~\citep[(1.5) page 10]{cat2007}.
 \begin{definition}
\label{dfn:complexity:catoni}
 We say that Catoni's dimension assumption is satisfied for dimension $d_\pi>0$ if
 $$ \sup_{\beta\geq 0} \beta \mathbb{E}_{\theta\sim \pi_{-\beta R}} \left[ R(\theta)-R(\theta^*)\right] = d_\pi .$$
\end{definition}
\begin{lemma}
\label{lemma:oracle:bound:dimension}
 Under Catoni's dimension assumption with dimension $d_\pi$, for any $\beta\geq \beta_0 = d_\pi / C$,
 $$ -\log \mathbb{E}_{\theta\sim\pi} \left\{ {\rm e}^{-\beta[R(\theta)-R(\theta^*)]}
 \right\} \leq d_\pi \log\left( \frac{{\rm e}C\beta}{d_\pi} \right) . $$
 In other words, if Catoni's dimension assumption is safisfied, then the assumption of Theorem~\ref{thm:oracle:bound:dimension} given by~\eqref{equa:dimension:first:prime} is satisfied with $c_\pi = d_\pi / ({\rm e} C)$.
\end{lemma}
\noindent {\it Proof:}
Define
$$ f(\xi) = -\log \mathbb{E}_{\theta\sim\pi} {\rm e}^{-\xi[R(\theta)-R(\theta^*)]} $$
for any $\xi\geq 0$. First, note that
$$ f(0) =  -\log \mathbb{E}_{\theta\sim\pi} {\rm e}^{0} = - \log (1) = 0. $$
Moreover, we can check that $f$ is differentiable and that
\begin{align*}
f'(\xi)
& = \frac{\mathbb{E}_{\theta\sim\pi}\left\{ [R(\theta)-R(\theta^*)] {\rm e}^{-\xi[R(\theta)-R(\theta^*)]}\right\} }{\mathbb{E}_{\theta\sim\pi} \left\{ {\rm e}^{-\xi[R(\theta)-R(\theta^*)]}\right\}}
\\
& = \mathbb{E}_{\theta\sim \pi_{-\xi R}} \left[ R(\theta)-R(\theta^*)\right]
\\
& \leq \frac{d_\pi}{\xi}
\end{align*}
where we used Definition~\ref{dfn:complexity:catoni} for the last inequality. This inequality is quite tight for large $\xi$, but useless when $\xi \rightarrow 0$. In this case, we will rather use the (simpler) inequality:
\begin{align*}
f'(\xi)
& = \frac{\mathbb{E}_{\theta\sim\pi}\left\{ [R(\theta)-R(\theta^*)] {\rm e}^{-\xi[R(\theta)-R(\theta^*)]}\right\} }{\mathbb{E}_{\theta\sim\pi} \left\{ {\rm e}^{-\xi[R(\theta)-R(\theta^*)]}\right\}}
\\
& \leq \frac{\mathbb{E}_{\theta\sim\pi}\left\{ C {\rm e}^{-\xi[R(\theta)-R(\theta^*)]}\right\} }{\mathbb{E}_{\theta\sim\pi} \left\{ {\rm e}^{-\xi[R(\theta)-R(\theta^*)]}\right\}}
\\
& = C.
\end{align*}
Combining both bounds, $f'(\xi) \leq \min(C,d_\pi/\xi)$.
Integrating for $0\leq \xi \leq \beta$ gives:
\begin{align*}
f(\beta) & = f(\beta) - f(0)
\\
& = \int_{0}^{\beta} f'(\xi) {\rm d}\xi
\\
& \leq \int_{0}^{\frac{d_\pi}{C}} C {\rm d}\xi + \int_{\frac{d_\pi}{C}}^{\beta} \frac{d_\pi}{\xi} {\rm d}\xi
\\
& = C\frac{d_\pi}{C} + d_\pi\log(\beta) - d_\pi \log\left(\frac{d_\pi}{C}\right)
\\
& = d_\pi \log\left( \frac{{\rm e}C\beta}{d_\pi} \right)
 \text{. } \square
\end{align*}

Another classical condition is as follows.
\begin{definition}
\label{lemma:oracle:bound:dimension:2}
 We say that the prior mass condition is satisfied with constants $c$ and $d_\pi$ if there is $r_0>0$ such that, for any $r\leq r_0$,
 $$  \pi (\{\theta\in\Theta: R(\theta)-R(\theta^*)  \leq r\}) \geq \left(\frac{r}{c} \right)^{d_\pi} .$$
\end{definition}
This type of condition is classical to analyze the asymptotics of Bayesian estimators in statistics~\citep{gho2017}.
\begin{lemma}
 Under the prior mass condition with constants $c$ and $d_\pi$, ~\eqref{equa:dimension:first:prime} is satisfied with $c_\pi = d_\pi/({\rm e} c)$, for any  $\beta\geq \beta_0 = d_\pi/r_0 $.
\end{lemma}
\noindent {\it Proof:}
The proof mimics the strategy of Example~\ref{exm:lipschitz:uniform}, but in a more general setting. We define, for any $r>0$, $\rho_r$ as the restriction of $\pi$ to $\{\theta\in\Theta:  R(\theta)-R(\theta^*)  \leq r \}$. Then,
\begin{align*}
 f(\beta) & =
 \inf_{\rho\in \mathcal{P}(\Theta)} \left[ \mathbb{E}_{\theta\sim\rho}[R(\theta)-R(\theta^*)] + \frac{KL(\rho\|\pi)}{\beta} \right] 
 \\
 & \leq  \inf_{r>0} \left[ \mathbb{E}_{\theta\sim\rho}[R(\theta)-R(\theta^*)] + \frac{KL(\rho_r\|\pi)}{\beta} \right]
 \\
 & \leq \inf_{r>0} \left[ r + \frac{\log \frac{1}{ \pi (\{\theta\in\Theta: R(\theta)-R(\theta^*)  \leq r\})  } }{\beta} \right].
\end{align*}
We put  $r=d_\pi/\beta $. Note that as $\beta\geq d_\pi / r_0$, $r\leq r_0$ and thus
$$
f(\beta) \leq r + \frac{d_\pi \log \frac{c}{ r } }{\beta} = \frac{d_\pi}{\beta} + \frac{d_\pi}{\beta} \log\left(  \frac{c \beta }{d_\pi} \right) =  \frac{d_\pi}{\beta} \log\left(  \frac{{\rm e} c \beta }{d_\pi} \right) . \quad \square$$

\section{Getting Rid of the $\log$ Terms: Catoni's Localization Trick}

We have seen in Subsection~\ref{subsec:neural:nets} Catoni's idea to replace the prior by $\pi_{-\beta R}$ for some $\beta>0$, where $\pi_{-\beta R}$ is given by Definition~\ref{dfn:complexity:catoni}. This technique is called ``localization of the bound'' by Catoni. Used in empirical bounds, this trick can lead to tighter bound. We will study its effect on oracle bounds. Let us start by providing a counterpart of Theorem~\ref{thm:oracle:bound} using this trick (with $\beta=\lambda/4$).

\begin{theorem}
\label{thm:oracle:bound:local}
Assume that Bernstein condition holds for some $K>0$, and take $\lambda=n/\max(2K,C)$. Then
 \begin{multline*}
  \mathbb{E}_{\mathcal{S}}\left\{ \mathbb{E}_{\theta\sim\hat{\rho}_\lambda} [R(\theta)] - R(\theta^*)\right\}
 \\
 \leq
 \inf_{\rho\in\mathcal{P}(\Theta)} \left\{
 3 \mathbb{E}_{\theta\sim\rho} [R(\theta) - R(\theta^*)] +\frac{4 \max(2K,C) KL\left(\rho\|\pi_{-\frac{\lambda}{4}R}\right) }{n} \right\}.
\end{multline*}
\end{theorem}

Before we give the proof, we will show a striking consequence: the $\log(n)$ terms in the last bullet point in the list of rates of convergence can be removed:
\begin{itemize}
 \item when Catoni's dimension $d_\pi<\infty$, ${\rm rate}_n(\pi)$ is in $\sqrt{d_\pi /n}$ in general, and in $ d_\pi /n$ under Bernstein condition, {\it if we use a localized bound}.
\end{itemize}
Indeed, take $\rho=\pi_{-\frac{\lambda}{4}R} = \pi_{-\{n/[4\max(2K,C)]\} R }$ in the right-hand side of Theorem~\ref{thm:oracle:bound:local}:
 \begin{equation*}
  \mathbb{E}_{\mathcal{S}}\left\{ \mathbb{E}_{\theta\sim\hat{\rho}_\lambda} [R(\theta)] - R(\theta^*)\right\}
 \leq
 3 \mathbb{E}_{\theta\sim \pi_{-\frac{\lambda}{4}R}} [R(\theta) - R(\theta^*)] + \frac{0}{n}.
\end{equation*}
Using Definition~\ref{dfn:complexity:catoni} we obtain the following corollary.

\begin{corollary}
 Assume that Catoni's dimension condition is satisfied with dimension $d_\pi>0$.
 Assume that Bernstein condition holds for some $K>0$, and take $\lambda=n/\max(2K,C)$, then:
  \begin{equation*}
  \mathbb{E}_{\mathcal{S}}  \mathbb{E}_{\theta\sim\hat{\rho}_\lambda} [R(\theta)] \leq R(\theta^*)
 +
\frac{12 d_\pi \max(2K,C)}{n}.
\end{equation*}
\end{corollary}

We can also briefly detail the consequence of the bound in the finite case.
\begin{example}[The finite case]
\label{exm:finite:localized}
 When ${\rm card}(\Theta)=M$ is finite and $\pi$ is uniform on $\Theta$, Theorem~\ref{thm:oracle:bound:local} applied to $\rho=\delta_{\theta^*}$ gives:
  \begin{align}
  \mathbb{E}_{\mathcal{S}} & \left\{ \mathbb{E}_{\theta\sim\hat{\rho}_\lambda} [R(\theta)] - R(\theta^*)\right\}
  \nonumber
  \\
& \leq
\frac{4 \max(2K,C) KL\left(\delta_{\theta*}\|\pi_{-\frac{\lambda}{4}R}\right) }{n} \nonumber
\\
& = \frac{4 \max(2K,C) \log \sum_{\theta\in\Theta} {\rm e}^{-\frac{\lambda}{4} [R(\theta)-R(\theta^*)]} }{n} 
\nonumber
\\
& = \frac{4 \max(2K,C) \log \sum_{\theta\in\Theta} {\rm e}^{-\frac{n}{4\max(2K,C)} [R(\theta)-R(\theta^*)]} }{n}
\label{equa:local:finite}
.
\end{align}
Of course, we have:
$$\sum_{\theta\in\Theta} {\rm e}^{-\frac{n}{4\max(2K,C)} [R(\theta)-R(\theta^*)]} \leq M $$
and thus we recover the rate in $\log(M)/n$:
$$
  \mathbb{E}_{\mathcal{S}}\left\{ \mathbb{E}_{\theta\sim\hat{\rho}_\lambda} [R(\theta)] - R(\theta^*)\right\} \leq \frac{4 \max(2K,C) \log(M) }{n}.
$$
On the other hand, in some situations, we can do better from~\eqref{equa:local:finite}. Fix a threshold $\tau>0$ and define $m_\tau = {\rm card}(\{\theta\in\Theta : R(\theta)-R(\theta^*) \leq \tau \} )\in\{1,\dots,M\} $. Then we obtain the bound:
\begin{multline*}
  \mathbb{E}_{\mathcal{S}}\left\{ \mathbb{E}_{\theta\sim\hat{\rho}_\lambda} [R(\theta)] - R(\theta^*)\right\} 
  \\
  \leq \frac{4 \max(2K,C) \log\left( m_\tau + {\rm e}^{-\frac{n \tau}{4\max(2K,C)}}(M-m_\tau) \right) }{n}
\end{multline*}
which will be much smaller for large $n$. A similar result was obtained using a different technique by \citet{lecueM2013optimality}.
\end{example}

\enlargethispage{-3\baselineskip}
\noindent {\it Proof of Theorem~\ref{thm:oracle:bound:local}:} we follow the proof of Theorem~\ref{thm:oracle:bound} until~\eqref{equa:proof:thm:oracle} that we remind here:
\begin{equation*}
 \mathbb{E}_{\mathcal{S}}\left( {\rm e}^{\lambda \left\{ \left[ 1- K g\left(\frac{\lambda C}{n}\right) \frac{\lambda}{n} \right]\left[ R(\theta) - R(\theta^*)\right] -r(\theta) + r(\theta^*)\right\} } \right)
\leq 1.
\end{equation*}
Now, we integrate this with respect to $ \pi_{-\beta R}$ for some $\beta>0$ and use Fubini:
\begin{equation*}
 \mathbb{E}_{\mathcal{S}}\mathbb{E}_{\theta\sim\pi_{-\beta R}}\left( {\rm e}^{\lambda \left\{ \left[ 1- K g\left(\frac{\lambda C}{n}\right) \frac{\lambda}{n} \right]\left[ R(\theta) - R(\theta^*)\right] -r(\theta) + r(\theta^*)\right\} } \right)
\leq 1
\end{equation*}
and Donsker and Varadhan's formula:
\begin{multline*}
 \mathbb{E}_{\mathcal{S}}\exp \Biggl(  \sup_{\rho\in\mathcal{P}(\Theta)} \Biggl( \lambda \mathbb{E}_{\theta\sim\rho} \Biggl\{ \left[ 1- K g\left(\frac{\lambda C}{n}\right) \frac{\lambda}{n} \right]\left[ R(\theta) - R(\theta^*)\right]
 \\
 -r(\theta) + r(\theta^*)\Biggr\} -KL(\rho\| \pi_{-\beta R}) \Biggr) \Biggr)
\leq 1.
\end{multline*}
At this point, we write explicitly
\begin{align*}
 KL(\rho\| \pi_{-\beta R}) & =
 \mathbb{E}_{\theta\sim\rho}\left[ \log\left( \frac{{\rm d} \rho}{{\rm d} \pi_{-\beta R}}(\theta)\right)\right]
 \\
 & =  \mathbb{E}_{\theta\sim\rho}\left[ \log\left( \frac{{\rm d} \rho}{{\rm d} \pi}(\theta) \frac{\mathbb{E}_{\vartheta\sim \pi}[{\rm e}^{-\beta[R(\vartheta)-R(\theta^*)]}] }{{\rm e}^{-\beta[R(\theta)-R(\theta^*)]}} \right)\right]
 \\
 & = 
 KL(\rho\| \pi) + \beta \mathbb{E}_{\theta \sim\rho}[R(\theta)-R(\theta^*)]
 \\
 & \qquad + \log \mathbb{E}_{\vartheta\sim \pi}[{\rm e}^{-\beta[R(\vartheta)-R(\theta^*)]}] 
\end{align*}
which, plugged in the last formula, gives:
\begin{multline*}
 \mathbb{E}_{\mathcal{S}} \exp \Biggl( \lambda \sup_{\rho\in\mathcal{P}(\Theta)} \Biggl( \mathbb{E}_{\theta\sim\rho} \Biggl\{ \left[ 1- K g\left(\frac{\lambda C}{n}\right) \frac{\lambda}{n} - \frac{\beta}{\lambda} \right]\left[ R(\theta) - R(\theta^*)\right]
 \\
 -r(\theta) + r(\theta^*)\Biggr\}
 -KL(\rho\| \pi)-\log \mathbb{E}_{\vartheta\sim \pi}[{\rm e}^{-\beta[R(\vartheta)-R(\theta^*)]}]  \Biggr) \Biggr)
\leq 1.
\end{multline*}
We apply Jensen and rearrange terms to obtain, for any randomized estimator $\hat{\rho}$,
\begin{multline*}
  \left[ 1- K g\left(\frac{\lambda C}{n}\right) \frac{\lambda}{n} - \frac{\beta}{\lambda} \right] \mathbb{E}_{\mathcal{S}}\left\{ \mathbb{E}_{\theta\sim\hat{\rho}} [R(\theta)] - R(\theta^*)\right\}
  \\
  \leq \mathbb{E}_{\mathcal{S}} \left\{  \mathbb{E}_{\theta\sim\hat{\rho}} [r(\theta)] - r(\theta^*) +\frac{KL(\hat{\rho}\| \pi) }{\lambda} \right\}
  \\
  + \frac{\log \mathbb{E}_{\vartheta\sim \pi}\left[{\rm e}^{-\beta[R(\vartheta)-R(\theta^*)]}\right]}{\lambda}.
\end{multline*}
Here again, the r.h.s. is minimized for $\hat{\rho}=\hat{\rho}_\lambda$ the Gibbs posterior, and we obtain:
\begin{align*}
  \Biggl[ 1- & K g\left(\frac{\lambda C}{n}\right) \frac{\lambda}{n} - \frac{\beta}{\lambda} \Biggr] \mathbb{E}_{\mathcal{S}}\left\{ \mathbb{E}_{\theta\sim\hat{\rho}_\lambda} [R(\theta)] - R(\theta^*)\right\}
  \\
 & \leq \mathbb{E}_{\mathcal{S}} \left\{ \inf_{\rho\in\mathcal{P}(\Theta)}\left[ \mathbb{E}_{\theta\sim\rho} [r(\theta)] - r(\theta^*) +\frac{KL(\rho\|\pi) }{\lambda} \right] \right\}
 \\
 & \qquad \qquad
 + \frac{\log \mathbb{E}_{\vartheta\sim \pi}\left[{\rm e}^{-\beta[R(\vartheta)-R(\theta^*)]}\right]}{\lambda}
  \\
 & \leq  \inf_{\rho\in\mathcal{P}(\Theta)} \left\{ \mathbb{E}_{\mathcal{S}} \left[ \mathbb{E}_{\theta\sim\rho} [r(\theta)] - r(\theta^*) +\frac{KL(\rho\|\pi) }{\lambda} \right] \right\}
 \\
 & \qquad \qquad
 + \frac{\log \mathbb{E}_{\vartheta\sim \pi}\left[{\rm e}^{-\beta[R(\vartheta)-R(\theta^*)]}\right]}{\lambda}
 \\
 & = \inf_{\rho\in\mathcal{P}(\Theta)}\left[ \mathbb{E}_{\theta\sim\rho} [R(\theta)] - R(\theta^*) +\frac{KL(\rho\|\pi) }{\lambda} \right]
 \\
 & \qquad \qquad
 + \frac{\log \mathbb{E}_{\vartheta\sim \pi}\left[{\rm e}^{-\beta[R(\vartheta)-R(\theta^*)]}\right]}{\lambda}
 \\
 & = \inf_{\rho\in\mathcal{P}(\Theta)}\left\{ \mathbb{E}_{\theta\sim\rho} [R(\theta) - R(\theta^*)] \left(1 - \frac{\beta}{\lambda} \right) +\frac{KL(\rho\|\pi_{-\beta R}) }{\lambda} \right\}
\end{align*}
where we used again the formula on $KL(\rho\|\pi_{-\beta R})$ for the last step. So, for $\beta$ and $\lambda$ such that
$$    K g\left(\frac{\lambda C}{n}\right) \frac{\lambda}{n} - \frac{\beta}{\lambda} < 1 $$
we have: \enlargethispage{\baselineskip}
\begin{multline*}
 \mathbb{E}_{\mathcal{S}}\left\{ \mathbb{E}_{\theta\sim\hat{\rho}_\lambda} [R(\theta)] - R(\theta^*)\right\}
 \\
 \leq
 \inf_{\rho\in\mathcal{P}(\Theta)}
 \frac{ \left(1 - \frac{\beta}{\lambda} \right)\mathbb{E}_{\theta\sim\rho} [R(\theta) - R(\theta^*)] +\frac{KL(\rho\|\pi_{-\beta R}) }{\lambda}  }{1- K g\left(\frac{\lambda C}{n}\right) \frac{\lambda}{n} - \frac{\beta}{\lambda} }.
\end{multline*}
For example, for $\lambda = n/\max(2K,C)$ we have already seen that
$$ K g\left(\frac{\lambda C}{n}\right) \frac{\lambda}{n} \leq \frac{1}{2} $$
and taking $\beta = \lambda/4$ leads to
\begin{multline*}
 \mathbb{E}_{\mathcal{S}}\left\{ \mathbb{E}_{\theta\sim\hat{\rho}_\lambda} [R(\theta)] - R(\theta^*)\right\}
 \\
 \leq
 \inf_{\rho\in\mathcal{P}(\Theta)}
 \frac{ \frac{3}{4} \mathbb{E}_{\theta\sim\rho} [R(\theta) - R(\theta^*)] +\frac{\max(2K,C) KL\left(\rho\left\| \pi_{-\frac{\lambda}{4}R} \right. \right) }{n}  }{ 1 - \frac{3}{4} }
\end{multline*}
that is
\begin{multline*}
  \mathbb{E}_{\mathcal{S}}\left\{ \mathbb{E}_{\theta\sim\hat{\rho}_\lambda} [R(\theta)] - R(\theta^*)\right\}
 \\
 \leq
 \inf_{\rho\in\mathcal{P}(\Theta)} \left\{
 3 \mathbb{E}_{\theta\sim\rho} [R(\theta) - R(\theta^*)] +\frac{4 \max(2K,C) KL\left(\rho\left\| \pi_{-\frac{\lambda}{4}R} \right. \right) }{n} \right\}
\end{multline*}
which ends the proof.
$\square$

We refer the reader to~\citet{loc2023} for more recent results on localization.
We end up this section with the following comment page 15 in Catoni's book~\citep{cat2007}: ``some of the detractors of the PAC-Bayesian approach (which, as a newcomer, has sometimes received a suspicious greeting among statisticians) have argued that it cannot bring anything that elementary union bound arguments could not essentially provide. We do not share of course this derogatory opinion, and while we think that allowing for non atomic priors and posteriors is worthwhile, we also would like to stress that the upcoming local and relative bounds could hardly be obtained with the only help of union bounds''.

\chapter{Beyond ``Bounded Loss'' and ``i.i.d. Observations''}
\label{section:not:bounded:iid}

If you follow the proof of the PAC-Bayesian inequalities seen so far, you will see that the ``bounded loss'' and ``i.i.d. observations'' assumptions are used only to apply Lemma~\ref{lemma:hoeffding} (Hoeffding's inequality) or Lemma~\ref{lemma:bernstein} (Bernstein's inequality). In other words, in order to prove PAC-Bayes inequalities for unbounded losses or dependent observations, all we need is a result similar to Hoeffding or Bernstein's inequalities (also called exponential moment inequalities) in this context.

\enlargethispage{-\baselineskip}
In the past 15 years, many variants of PAC-Bayes bounds were developped for various applications based on this remark. In this section, we provide some pointers. In the end, some authors now prefer to assume directly that the data is such that it satisfies a given exponential inequality. One of the merits of Theorem~\ref{thm:bound:germain} above, from~\citet{ger2009}, is to make it very explicit: the exponential moment appears in the bound. A similar approach is used by~\citet{alq2016}: a ``Hoeffding assumption'' and a ``Bernstein assumption'' are defined, that corresponds to data satisfying a Hoeffding type inequality, or a Bernstein type inequality + the usual Bernstein condition (Definition~\ref{dfn:bernstein}). A similar point of view is used by \citet{riv2020}.

\begin{remark}
It is possible to prove a PAC-Bayes inequality like Theorem~\ref{thm:first:bound} starting directly from~\eqref{equa:proof:catoni:0}, that is, assuming that an exponential moment inequality is satisfied in average under the prior $\pi$, rather than uniformly on $\theta$. We will not develop this approach here, examples are detailed by \citet{alq2016,haddouche2020pac,riv2020}.
\end{remark}

\section{``Almost'' Bounded Losses (Sub-Gaussian and Sub-gamma)}

\subsection{The sub-Gaussian case}

Hoeffding's inequality for $n=1$ variable $U_1$ taking values in $[a,b]$ simply states that
$$ \mathbb{E}\left( {\rm e}^{t[U_1-\mathbb{E}(U_1)]} \right) \leq {\rm e}^{\frac{t^2 (b-a)^2}{8}}.  $$
Then the general case is obtained by:
\begin{align*}
 \mathbb{E}\left( {\rm e}^{t\sum_{i=1}^n[U_i-\mathbb{E}(U_i)]} \right)
 & = \mathbb{E}\left( \prod_{i=1}^n {\rm e}^{t[U_i-\mathbb{E}(U_i)]} \right) \\
 & = \prod_{i=1}^n \mathbb{E}\left( {\rm e}^{t[U_i-\mathbb{E}(U_i)]} \right) \text{ (by independence)}
 \\
 & \leq \prod_{i=1}^n {\rm e}^{\frac{t^2 (b-a)^2}{8}}
 \\
 & = {\rm e}^{\frac{n t^2 (b-a)^2}{8}}.
\end{align*}

Alternatively, if we simply {\it assume} that, for some $C>0$,
\begin{equation}
\label{equa:subgauss}
\mathbb{E}\left( {\rm e}^{t[U_1-\mathbb{E}(U_1)]} \right) \leq {\rm e}^{C t^2}
\end{equation}
for some constant $C$, similar derivations lead to:
\begin{equation*}
 \mathbb{E}\left( {\rm e}^{t\sum_{i=1}^n[U_i-\mathbb{E}(U_i)]} \right)
 \leq  {\rm e}^{n C t^2},
\end{equation*}
on which we can build PAC-Bayes bounds. We can actually rephrase Hoeffding's inequality by: ``if $U_1$ takes values in $[a,b]$, then~\eqref{equa:subgauss} is satisfied for $C=(b-a)^2/8$''.

It appears that~\eqref{equa:subgauss} is satisfied by some unbounded variables. For example, it is well known that, if $U_i\sim\mathcal{N}(m,\sigma^2)$ then
\begin{equation*}
\mathbb{E}\left( {\rm e}^{t[U_1-\mathbb{E}(U_1)]} \right) = {\rm e}^{\frac{\sigma^2 t^2}{2}},
\end{equation*}
that is~\eqref{equa:subgauss} with $C=\sigma^2/2$. Actually, it can be proven that a variables $U_1$ will satisfy~\eqref{equa:subgauss} if and only if its tails $ \mathbb{P}(|U_1|\geq t)$ converge to zero (when $t\rightarrow\infty$) as fast as the ones of a Gaussian variable, that is $ \mathbb{P}(|U_1|\geq t)\leq \exp(-t^2/C') $ for some $C'>0$, see e.g. \citet[Chapter 1]{chafai2012interactions}. This is the reason behind the following terminology.
\begin{definition}
 A random variable $U$ such that
 $$ \mathbb{E}\left( {\rm e}^{t[U-\mathbb{E}(U)]} \right) \leq {\rm e}^{C t^2} $$ for some finite $C$  is called a sub-Gaussian random variable (with constant $C$).
\end{definition}

Based on this definition, we can state for example the following variant of Theorem~\ref{thm:first:bound} that will be valid for (some!) unbounded losses.
\begin{theorem}
\label{thm:bound:subgauss}
Assume that for any $\theta$ the $\ell_i(\theta)$ are independent and sub-Gaussian random variables with constant $C$. Then for any $\delta\geq 0$, for any $\lambda>0$,
\begin{multline*}
 \mathbb{P}_\mathcal{S}\Biggl(\forall\rho\in\mathcal{P}(\Theta) \text{, } \mathbb{E}_{\theta\sim\rho}[ R(\theta)] \leq \mathbb{E}_{\theta\sim\rho}[ r(\theta)]
 \\
 +  \frac{\lambda C^2}{n} + \frac{KL(\rho\|\pi) + \log\frac{1}{\delta}}{\lambda} \Biggr)
 \geq 1-\delta.
\end{multline*}
\end{theorem}
(We don't provide the proof as all the ingredients were explained to the reader).

For example, the PAC-Bayes bounds of~\citet{alq2016} were explicitly stated for sub-Gaussian losses.

\subsection{The sub-gamma case}

We will not provide details, but variables satisfying inequalities similar to Bernstein's inequality are called sub-gamma random variables, sometimes sub-exponential random variables. A possible characterization is: $ \mathbb{P}(|U_1|\geq t)\leq \exp(-t/C') $ for some $C'>0$. Such variables include: gamma (and exponential) random variables, Gaussian variables and bounded variables.

\citet[Chapter 2]{bou2013} provide a very detailed and pedagogical overview of exponential moment inequalities for independent random variables, and in particular we refer the reader to their Section 2.4 for more details on sub-gamma variables (but I have to warn you, this book is so cool you will find difficult to stop at the end of Chapter 2 and will end up reading everything).

In the literature, PAC-Bayes bounds for sub-gamma random variables can be found as early as 2001, see \citet[Chapter 5]{cat2004}. These are these bounds that are used to prove minimax rates in various parametric and non-parametric problems in the aforementioned~\citep{alquier2013sparse,mai2015,trabs2022}.

\subsection{Remarks on exponential moments}

Finally, exponential moments inequalities for random variables such that $ \mathbb{P}(|U_1|\geq t)\leq \exp(-t^\alpha/C') $ where $\alpha\geq 1$ are studied by \citet[Chapter 1]{chafai2012interactions}. The set of such variables is called an Orlicz space.

Still, all these random variables are defined in such a way that they satisfy more or less the same exponential inequalities as bounded variables. And indeed, for these variables, $ \mathbb{P}(|U_1|\geq t)$ is very small when $t$ is large -- hence the title of this section: {\it almost} bounded variables. We will now discuss briefly how to go beyond this case.

\section{Heavy-tailed Losses}

By heavy-tailed variables, we mean typically random variables $U_1$ such that $\mathbb{P}(|U_1|\geq t) $ is for example in $t^{-\alpha}$ for some $\alpha>0$.

\subsection{The truncation approach}

In my PhD thesis~\citep{alquier2006transductive}, I studied a truncation technique for general losses $\ell_i(\theta)$. That is, write:
$$ \ell_i(\theta) = \ell_i(\theta) \mathbf{1}(\ell_i(\theta)\leq s) + \ell_i(\theta) \mathbf{1}(\ell_i(\theta)>s) $$
for some $s>0$.
The first term is bounded by $s$, so we can use exponential moments inequalities on it, while I used inequalities on the tails $\mathbb{P}(|\ell_i(\theta)|\geq s)$ to control the second term. For the sake of completeness I state one of the bounds obtained by this technique (with $s=n/\lambda$).
\begin{theorem}[Corollary 2.5, \citet{alquier2006transductive}]
Define
$$ \Delta_{n,\lambda}(\theta) = \mathbb{E}_{(X,Y)\sim P}\left\{ \max\left[\ell(f_{\theta}(X),Y)-\frac{n}{\lambda},0\right] \right\}  $$
and
$$
\tilde{r}_{\lambda,n}(\theta) = \frac{1}{n}\sum_{i=1}^n  \Psi_{\frac{\lambda}{n}}\left[\min \left( \ell(f_{\theta}(X_i),Y_i),\frac{n}{\lambda} \right)\right],
$$
where
$$
\Psi_{\alpha}(u) := \frac{-\log(1-u \alpha)}{\alpha} \text{ and thus } \Psi^{-1}_{\alpha}(v) = \frac{1-{\rm e}^{-\alpha v}}{\alpha}.
$$
Then, for any $\delta>0$, for any $\lambda>0$,
\begin{multline*}
\mathbb{P}_{\mathcal{S}}
\Biggl(\forall \rho\in\mathcal{P}(\Theta)\text{, }
\mathbb{E}_{\theta\sim\rho}[R(\theta)]
\\
\leq
\Psi_{\frac{\lambda}{n}}^{-1} \left\{
\mathbb{E}_{\theta\sim\rho}[\tilde{r}_{\lambda,n}(\theta)]
+
\frac{KL(\rho\|\pi) + \log\frac{1}{\delta}}{\lambda}
\right\} + \mathbb{E}_{\theta\sim\rho}[\Delta_{n,\lambda}(\theta)]
\Biggr)
\\
\geq 1-\delta.
\end{multline*}
\end{theorem}
Note that $\tilde{r}_{\lambda,n}(\theta)$ is an approximation of $r(\theta)$ when $\lambda/n$ is small enough (usually $\lambda \sim \sqrt{n}$ in this bound). The function $\Psi_\alpha$ plays a role similar to the function $\Phi_\alpha$ in Catoni's bound (Theorem~\ref{thm:catoni:2}), and more explicit inequalities can be derived by upper-bounding $\Psi^{-1}_\alpha$. Finally, $\Delta_{n,\lambda}(\theta)$ corresponds to the tails of the loss function. Actually, for a bounded loss, we will have $\Delta_{n,\lambda}(\theta)=0$ for $n/\lambda$ large enough. In the sub-exponential setting, $\Delta_{n,\lambda}(\theta)>0$ but will usually not be the dominant term in the right-hand side. However, \citet{alquier2006transductive} also provided upper bounds on $\Delta_{n,\lambda}(\theta)$ in $\mathcal{O}((\lambda/n)^{s-1})$ where $s$ is such that $\mathbb{E}(\ell_i^s)<+\infty$, but this terms is dominant in this case (and thus, it slows down the rate of convergence). This truncation argument is reused by \citet{alq2008} but only the oracle bounds are provided there. \citet{sel2012c} used distribution smoothing as an alternative truncation approach to truncate the unbounded log-loss in density estimation.

\subsection{Bounds based on moment inequalities}

Based on techniques developped of \citet{hon2014,beg2016}, \citet{alq2018} proved inequalities similar to PAC-Bayes bounds, that hold for heavy-tailed losses (they can also hold for non i.i.d. losses, we will discuss this point later). Curiously, these bounds depend no longer on the Kullback-Leibler divergence, but on other divergences. An example of such an inequality is provided here and relies only on the assumption that the losses have a variance. We recall a notation from Section~\ref{section:tight}: ${\rm V}(\theta) = {\rm Var}(\ell_i(\theta))$.
\begin{theorem}[Corollary 1, \citet{alq2018}]
\label{bound:chi2}
Assume that the $\ell_i(\theta)$ are independent and such that ${\rm V}(\theta)  \leq \kappa < \infty$. Then, for any $\delta>0$,
\begin{multline*}
\mathbb{P}_{\mathcal{S}}
\left(\forall \rho\in\mathcal{P}(\Theta)\text{, }
\mathbb{E}_{\theta\sim\rho}[R(\theta)]
\leq
\mathbb{E}_{\theta\sim\rho}[r(\theta)]
+
\sqrt{\frac{\kappa\left( 1+\chi^2(\rho\|\pi) \right)}{n\delta}}
\right) \\
\geq 1-\delta,
\end{multline*}
where $\chi^2(\rho\|\pi) $ is the chi-square divergence:
$$
\chi^2(\rho\|\pi) =
\left\{
\begin{array}{l} \int \biggl[\left(\frac{{\rm d}\rho}{{\rm d}\pi}(\theta)\right)^2-1 \biggr] \pi({\rm d}\theta) \text{ if } \rho \ll\pi \text{, and }\\
 \chi^2(\rho\|\pi) = +\infty \text{ otherwise.}
 \end{array}
\right.
$$
\end{theorem}
Interestingly, the minimization of the bound with respect to $\rho$ leads to an explicit solution~\citep{alq2018}. However the dependence of the rate in $\delta$ is much worse than in Theorem~\ref{thm:first:bound}. This was later dramatically improved by \citet{ohn2021}. As for the truncation approach described earlier, this approach leads to slow rates of convergence for heavy-tailed variables.

\subsection{Bounds based on robust losses}

\citet{cat2012} proposed a robust loss function $\psi$ used to estimate the mean of heavy-tailed random variables (this is based on ideas from an earlier paper by \cite{aud2011}). As a result, \citet{cat2012}  obtains, for the mean of heavy-tailed variables, confidence intervals very similar to the ones of estimators of the mean of a Gaussian.

More recently, \citet{hol2019} derives a full PAC-Bayesian theory for possibly heavy-tailed losses based on Catoni's technique. The idea is as follows. Put
$$
 \psi(u) = \left\{
 \begin{array}{l}
  \frac{-2\sqrt{2}}{3} \text{ if } u<-\sqrt{2}, \\
  u - \frac{u^3}{6} \text{ if } -\sqrt{2}\leq u \leq \sqrt{2}, \\
  \frac{2\sqrt{2}}{3} \text{ otherwise}
 \end{array}
\right.
$$
and, for any $s>0$,
$$ r_{\psi,s}(\theta) = \frac{s}{n}\sum_{i=1}^n \psi\left(\frac{\ell_i(\theta)}{s}\right). $$
The idea is that, even when $\ell_i(\theta)$ is unbounded, the new version of the risk, $r_{\psi,s}(\theta)$, is bounded. Thus, the study of the deviations of  $r_{\psi,s}(\theta)$ can be done with the standard tools for bounded losses. There is some additional work to connect $\mathbb{E}_{\mathcal{S}}[r_{\psi,s}(\theta)]$ to $R(\theta)$ for a well chosen $s$, and \citet{hol2019} obtains the following result.
\begin{theorem}[Theorem 9, \citet{hol2019}]
Let $\delta>0$.
Assume that the $\ell_i(\theta)$ are independent and
\begin{itemize}
 \item $\mathbb{E}(\ell_i(\theta)^2)\leq M_2<+\infty$ and $\mathbb{E}(\ell_i(\theta)^3)\leq M_3<+\infty$,
 \item for any $\theta\in\Theta$, $ R(\theta) \leq  \sqrt{n M_2 /(4\log(1/\delta))}$,
 \item $\delta \leq {\rm e}^{-1/9} \simeq 0.89$,
\end{itemize}
and put
$$ \pi^*_{n,s}(\Theta) = \frac{
\mathbb{E}_{\theta\sim\pi}\left[{\rm e}^{\sqrt{n}[R(\theta)-r_{\psi,s}(\theta)]} \right]
}{
\mathbb{E}_{\theta\sim\pi}\left[{\rm e}^{R(\theta)-r_{\psi,s}(\theta)} \right]
} $$
then, for $s:= n M_2/[2\log(1/\delta)]$,
\begin{multline*}
\mathbb{P}_{\mathcal{S}}\Biggl(
\forall \rho\in\mathcal{P}(\Theta)\text{, }
\mathbb{E}_{\theta\sim\rho}[R(\theta)]
\\
\leq
\mathbb{E}_{\theta\sim\rho}[r_{\psi,s}(\theta)]
+ \frac{
KL(\rho\|\pi) + M_2 + \frac{\log\left(\frac{8\pi M_2}{\delta^2}\right)}{2}+\pi^*_{n,s}(\Theta) -1
}{\sqrt{n}}
\Biggr)
\\
\geq 1-\delta.
\end{multline*}
\end{theorem}
On the contrary to Theorem~\ref{bound:chi2} above, the bound is very similar to the one in the bounded case (heavy-tailed variables do not lead to slower rates). In particular, we have a good dependence of the bound in $\delta$, and the presence of $KL(\rho\|\pi)$ that is much smaller than $\chi^2(\rho\|\pi)$. The only notable difference is the restriction in the range of $\delta$ (which is of no consequence in practice), and the term $\pi_n^*(\Theta)$. Unfortunately, as discussed by \citet[Remark 10]{hol2019}, this term will deterioriate the rate of convergence when the $\ell_i(\theta)$ are not sub-Gaussian (to my knowledge, it is not known if it will lead to better or worse rates than the ones obtained through truncation).

\section{Dependent Observations}

\subsection{Inequalities for dependent variables}

There are versions of Hoeffding and Bernstein's inequalities for dependent random variables, under various assumptions on this dependence. This can be used in the case where the observations are actually a time series, or a random field.

For example, we learnt auto-regressive predictors of the form $\hat{X_t} = f_{\theta}(X_{t-1},\dots,X_{t-k})$ for weakly dependent time series with a PAC-Bayes bound~\citep{alq2012}. The proof relies on Rio's version of Hoeffding's inequality~\citep{rio2000inegalites}. In our work, only slow rates in $ 1/\sqrt{n}$ are provided. Fast rates in $1/n$ were proven in a subsequent paper~\citep{alq2013} for (less general) mixing time series thanks to Samson's version of Bernstein's inequality~\citep{samson2000concentration}.

More exponential moment inequalities (and moment inequalities) for dependent variables can be found in the survey by \citet{win2010} and in monographs dedicated to weak dependence~\citep{dedecker2007weak}. Other time series models where PAC-Bayes bounds were used include martingales~\citep{sel2012}, Markov chains~\citep{ban2021}, continuous dynamical systems~\citep{hau2021}, LTI systems~\citep{eri2021}. Concentration inequalities on martingales are a very important tool to derive PAC-Bayes bounds in this context. Interestingly, they also allow to derive bounds that can hold when the sample size $n$ is not deterministic but instead might be chosen based on the observations seen so far (that is, $n$ is a stopping time), as observed by~\citet{chugg2023unified}.

\subsection{An example}

The weak-dependence conditions are quite general but they are also quite difficult to understand and the definitions are sometimes cumbersome. We only provide a much simpler example based on a more restrictive assumption, $\alpha$-mixing. This result is due to \citet{alq2018} and extends Theorem~\ref{bound:chi2} to time series.

\begin{definition}
Given two $\sigma$-algebras $\mathcal{F}$ and $\mathcal{G}$, we define
\begin{multline*}
\alpha(\mathcal{F},\mathcal{G}) = \sup\{{\rm Cov}(U,V); 0\leq U \leq 1 \text{, } U \text{ is } \mathcal{F}\text{-measurable,}
\\
0\leq V \leq 1 \text{, } V \text{ is } \mathcal{G}\text{-measurable}\}.
\end{multline*}
\end{definition}
Observe that if $\mathcal{F}$ and $\mathcal{G}$ are independent, $\alpha(\mathcal{F},\mathcal{G})=0$.
\begin{definition}
Given a time series $U=(U_t)_{t\in\mathbb{Z}}$ we define its $\alpha$-mixing coefficients by
$$ \forall h\in\mathbb{Z}\text{, } \alpha_h(U) = \sup_{t\in\mathbb{N}} \alpha(\sigma(U_t),\sigma(U_{t+h})). $$
\end{definition}

\begin{theorem}[Corollary 2, \citet{alq2018}]
\label{bound:chi2:time:series}
Let $X=(X_t)_{t\in\mathbb{Z}}$ be a real-valued stationary time series. Define, for $\theta=(\theta_1,\theta_2)\in\mathbb{R}^2$, $\ell_t(\theta) = (X_t - \theta_1 - \theta_2 X_{t-1})^2$ and $R(\theta) = \mathbb{E}_{X}[\ell_t(\theta)] $ (it does not depend on $t$ as the series is stationary). Define
$$ r(\theta)=\sum_{t=1}^n \ell_t(\theta) $$
the empirical risk based on the observation of $(X_0,\dots,X_n)$. Assume the prior $\pi$ is chosen such that
$$ \int \|\theta\|^6 \pi({\rm d}\theta) \leq M_6 <+\infty $$
(for example a Gaussian prior). Assume that $X$ the $\alpha$-mixing coefficients of $X$ satisfy:
$$ \sum_{t\in\mathbb{Z}}  \left[\alpha_t(X) \right]^{\frac{1}{3}} \leq \mathcal{A} < +\infty.   $$
Assume that $\mathbb{E}(X_t^6)\leq C <+\infty$. Define $ \nu = 32 C^{\frac{2}{3}} \mathcal{A} (1+4M_6)$.
Then, for any $\delta>0$,

\vspace{-0.8\baselineskip}
\small
$$
\mathbb{P}_{\mathcal{S}}
\left(\forall \rho\in\mathcal{P}(\Theta)\text{, }
\mathbb{E}_{\theta\sim\rho}[R(\theta)]
\leq
\mathbb{E}_{\theta\sim\rho}[r(\theta)]
+
\sqrt{\frac{\nu\left( 1+\chi^2(\rho\|\pi) \right)}{n\delta}}
\right) \geq 1-\delta.
$$\normalsize
\end{theorem}

\section{Other Non i.i.d. Settings}

\subsection{Non identically distributed observations}

When the data is independent, but non-identically distributed, that is, $(X_i,Y_i)\sim P_i$, we can still introduce
$$ r(\theta) = \frac{1}{n}\sum_{i=1}^n \ell_i(\theta) = \frac{1}{n}\sum_{i=1}^n \ell(f_{\theta}(X_i),Y_i) $$
and
\begin{equation*}
R(\theta) = \mathbb{E}[r(\theta)] = \frac{1}{n}\sum_{i=1}\mathbb{E}_{(X_i,Y_i)\sim P_i}[\ell(f_{\theta}(X_i),Y_i) ].
\end{equation*}
The proofs of most exponential inequalities still hold in this setting (for example, Hoeffding inequality when the losses are bounded). Based on this remark, all the results of \citet{cat2007} are written for independent, but not necessarily identically distributed observations. Of course, if we actually have $P_i = P$ for any $i$, we recover the usual case $ R(\theta) = \mathbb{E}_{(X,Y)\sim P}[\ell(f_{\theta}(X),Y) ] $.

\subsection{Shift in the distribution}

A common problem in machine learning practice is the shift in distribution: one learns a classifier based on i.i.d. observations $(X_i,Y_i)\sim P$. But in practice, the data to be predicted are drawn from another distribution $Q$, that is: $R(\theta) = \mathbb{E}_{(X,Y)\sim Q}[\ell(f_{\theta}(X),Y) ] \neq \mathbb{E}_{(X,Y)\sim P}[\ell(f_{\theta}(X),Y) ]$. There is still a lot of work to do to address this practical problem, but an interesting approach is proposed by \citet{ger2016b}: the authors use a technique called domain adaptation to allow the use of PAC-Bayes bounds in this context.

\subsection{Meta-learning}

Meta-learning is a scenario when one solves many machine learning tasks simulataneously, and the objective is to improve this learning process for yet-to-come tasks. A popular formalization (but not the only one possible) is:
\begin{itemize}
 \item each task $t\in\{1,\dots,T\}$ corresponds to a probability distribution $P_t$. The $P_t$'s are i.i.d. from some $\mathcal{P}$.
 \item for each task $t$, an i.i.d. sample $(X_1^t,Y_1^t),\dots,(X_n^t,Y_n^t)$ is drawn from $P_t$. Thus, we observe the empirical risk of task $t$:
 $$
 r_t(\theta) = \frac{1}{n}\sum_{i=1}^n \ell(f_{\theta}(X_i^t),Y_i^t)
 $$
 and use PAC-Bayes bounds to learn a good $\theta_t$ for this task.
 \item based on the data of tasks $\{1,\dots,T\}$, we want to improve the learning process for a yet non-observed task $P_{T+1}\sim\mathcal{P}$.
\end{itemize}
This improvement differs from one application to the other, for example: learn a better parameter space $\Theta_{T+1}\subset \Theta$, learn a better prior, learn better hyperparameters like $\lambda$... PAC-Bayes bounds for meta-learning were studied by \citet{pen2014,ami2018,jos2020,rot2020,liu2021,meunier2021,liu2021statistical,foong2021tight,rezazadeh2022general,riou2023bayes,sucker2023pac}. I believe PAC-Bayes bound are particularly convenient for meta-learning problems, and thus that this direction of research is very promising.

\chapter[Related Approaches in Statistics and Machine Learning \\ Theory]{Related Approaches in Statistics and Machine Learning Theory}
\label{section:related}
\chaptermark{Related Approaches in Statistics and Machine Learning Theory}

In this section, we list some connections between PAC-Bayes theory and other approaches in statistics and machine learning. We will mostly provide references, and will use mathematics more heuristically than in the previous sections. Note that these connections are well-known and were discussed in the literature, see for example \citet{banerjee2006Bayesian}.

\section{Bayesian Inference in Statistics}

In Bayesian statistics, we are given a sample $X_1,\dots,X_n$ assumed to be i.i.d. from some $P_{\theta^*}$ in a model $\{P_\theta,\theta\in\Theta\}$. A prior $\pi$ is given on the parameter set $\Theta$. When each $P_\theta$ has a density $p_\theta$ with respect to a given measure, the likelihood function is defined by
$$ \mathcal{L}(\theta;X_1,\dots,X_n) := \prod_{i=1}^n p_{\theta}(X_i). $$
According to the Bayesian paradigm, all the information on the parameter that can be inferred from the sample is in the posterior distribution
$$ \pi({\rm d}\theta|X_1,\dots,X_n) = \frac{\mathcal{L}(\theta;X_1,\dots,X_n)\pi({\rm d} \theta)}{\int \mathcal{L}(\vartheta;X_1,\dots,X_n)\pi({\rm d} \vartheta) }. $$

A direct remark is that $ \pi({\rm d}\theta|X_1,\dots,X_n)$ can be seen as a Gibbs posterior. Indeed, as
$$ \left[ \prod_{i=1}^n p_{\theta}(X_i) \right] \pi({\rm d} \theta)= {\rm e}^{\sum_{i=1}^n \log p_{\theta}(X_i)} \pi({\rm d} \theta) $$
we can define the loss $\ell_i(\theta) = -\log p_\theta (X_i)$ and the corresponding empirical risk is the negative log-likelihood:
$$
r(\theta) = \frac{1}{n} \sum_{i=1}^n [-\log p_\theta (X_i)],$$
and we have
\begin{align}
\pi(\cdot |X_1,\dots,X_n) & = \pi_{-n r}
\nonumber
\\
& = \argmin_{\rho\in\mathcal{P}(\Theta)} \left\{ \mathbb{E}_{\theta\sim\rho}\left[ \frac{1}{n}\sum_{i=1}^n [- \log p_\theta(X_i)]\right] + \frac{KL(\rho\|\pi)}{n}    \right\}.
\label{equa:prior:bayes}
\end{align}
This connection is for example discussed  by \citet{audPHD,ger2016a}. Note, however, that log-likelihoods are rarely bounded, which prevents to use the simplest PAC-Bayes bounds to study the consistency of $\pi({\rm d}\theta|X_1,\dots,X_n)$.

\subsection{Gibbs posteriors, generalized posteriors}

Independently from the PAC-Bayes community, the Bayesian statistics community proposed to generalize the posterior, by replacing the log-likelihood by minus a risk function. Often, the resulting generalised posterior
is called a Gibbs posterior. This is done for example by \citet{jia2008,syr2019} in order to estimate some parameters of the distributions of the data without having to model the whole distribution. Another instance of such generalized posteriors are fractional, or tempered posteriors, $\pi_{-\alpha n r}$, where $r$ is the negative log-likelihood and $\alpha<1$. \citet{gru2017} proved that in some contexts where the posterior $\pi({\rm d}\theta|X_1,\dots,X_n)$ is not consistent, a tempered posterior for $\alpha$ small enough will be. Gibbs posteriors are discussed in a decision theoretic framework by \citet{bis2016}. An asymptotic study of Gibbs posteriors, using different arguments than PAC-Bayes bounds (but related), is provided by \citet{syring2020gibbs} and some the references therein. Finally, \citet[Theorem 3.4]{bha2019} proved a PAC-Bayes bound for tempered posteriors.

\subsection{Contraction of the posterior in Bayes nonparametrics}

A very active field of research is the study of the contraction of the posterior in Bayesian statistics: the objective is to prove that $\pi({\rm d}\theta|X_1,\dots,X_n)$ concentrates around the true parameter $\theta^*$ when $n\rightarrow\infty$. We refer the reader to \citet{rousseau2016frequentist,gho2017} on this topic, see also \citet{banerjee2021bayesian} on high-dimensional models specifically. Usually, such results require two assumptions:
\begin{itemize}
 \item a technical condition, the existence of tests to discriminate between members of the model $\{P_\theta,\theta\in\Theta\}$,
 \item and the prior mass condition, which states that enough mass is given by the prior to a neighborhood of $\theta^*$. In other words, $\pi(\{\theta:d(\theta,\theta^*) \leq r \})$ does not converge too fast to $0$ when $r \rightarrow 0$, for some distance or risk measure $d$. For example, we can assume that there is a sequence $r_n\rightarrow 0$ when $n\rightarrow \infty$ such that
 \begin{equation}
  \label{prior:mass:condition}
  \pi(\{\theta:d(\theta,\theta^*) \leq r_n \}) \geq {\rm e}^{-n r_n}.
 \end{equation}
\end{itemize}
The prior mass condition is to be compared to Definition~\ref{lemma:oracle:bound:dimension:2} above. We can actually use the prior mass condition in conjunction with PAC-Bayes bounds, instead of Definition~\ref{lemma:oracle:bound:dimension:2}, to show that the bound is small. For example consider the PAC-Bayes inequality of Theorem~\ref{thm:oracle:bound}: under a Bernstein condition with constant $K$ and for a well chosen $ \lambda$,
\begin{multline*}
\mathbb{E}_\mathcal{S} \mathbb{E}_{\theta\sim\hat{\rho}_{\lambda}} [R(\theta) ]-R(\theta^*) 
   \leq 2  \inf_{\rho\in\mathcal{P}(\Theta)}  \Biggl\{  \mathbb{E}_{\theta\sim\rho} [R(\theta) ]-R(\theta^*) 
   \\
   + \frac{\max(2K,C) KL(\rho,\pi) }{n} \Biggr\}.
\end{multline*}
Assume that the prior mass condition holds with $d(\theta,\theta^*) = R(\theta)-R(\theta^*)$ and take
$$
\rho({\rm d}\theta) = \frac{ \pi({\rm d}\theta)\mathbf{1}(\{d(\theta,\theta^*)\leq r_n\}) }{\pi(\{\theta:d(\theta,\theta^*)\leq r_n\})} .
$$
We obviously have:
$$
\mathbb{E}_{\theta\sim\rho} [R(\theta) ]-R(\theta^*) \leq r_n
$$
and a direct calculation gives
$$
KL(\rho\|\pi) = -\log \pi(\{\theta:d(\theta,\theta^*) \leq r_n \}) \leq n r_n
$$
so the bound becomes:
\begin{equation*}
\mathbb{E}_\mathcal{S} \mathbb{E}_{\theta\sim\hat{\rho}_{\lambda}} [R(\theta) ]-R(\theta^*) 
   \leq 2[1 + \max(2K,C)]r_n \xrightarrow[n\rightarrow\infty]{} 0.
\end{equation*}
\citet[Theorem 3.1]{bha2019} proved the contraction of tempered posteriors their PAC-Bayes bound for tempered posteriors together with a prior mass condition.

\subsection{Variational approximations}

In many applications where the dimension of $\Theta$ is large, sampling from $\pi({\rm d}\theta|X_1,\dots,X_n) $ becomes a very difficult task. In order to overcome this difficulty, a recent trend is to approximate this probability distribution by a tractable approximation. Formally, we would chose a set $\mathcal{F}$ of probability distributions (for example, Gaussian distributions with diagonal covariance matrix) and define the following approximation of the posterior:
$$
\hat{\rho} = \argmin_{\rho\in\mathcal{F}} KL(\rho \|\pi(\cdot|X_1,\dots,X_n)).
$$
This is called a {\it variational approximation} in Bayesian statistics, we refer the reader to \citet{ble2017} for recent survey on the topic. Note that, by definition of $\pi({\rm d}\theta|X_1,\dots,X_n) $ we also have
$$
\hat{\rho} = \argmin_{\rho\in\mathcal{F}} \left\{ \mathbb{E}_{\theta\sim\rho}\left[ \frac{1}{n}\sum_{i=1}^n [- \log p_\theta(X_i)]\right] + \frac{KL(\rho\|\pi)}{n}    \right\}
$$
that is, a restricted version of~\eqref{equa:prior:bayes}.

This leads to two remarks:
\begin{itemize}
 \item non-exact minimization of PAC-Bayes bounds, as in Subsection~\ref{subsec:non:exact}, can be interpreted as variational approximations of Gibbs posteriors. Note that this is also the case of the Gaussian approximation that was used for neural networks in~\citep{dzi2017}. This led to a systematic study of the consistency of variational approximation via PAC-Bayes bounds~\citep{alq2016,she2017}.
 \item on the other hand, little was known at that time on the theoretical properties of variational approximations of the posterior in statistics. Using the fact that variational approximations of tempered posteriors are constrained minimizers of the PAC-Bayes bound of \citet{bha2019}, we studied the consistency of such approximations~\citep{alq2020}. As a byproduct we have a generalization of the prior mass condition for variational inference~\citep[(2.1) and (2.2)]{alq2020}. These results were extended to the standard posterior $\pi({\rm d}\theta|X_1,\dots,X_n)$ by \citet{yan2020,zha2020}.
\end{itemize}
More theoretical studies on variational inference (using PAC-Bayes, or not) appeared at the same time or since~\citep{cherief2018consistency,hug2018,mas2019,che2019a,plum2020,che2019b,ban2021,ave2021,ohn2021adaptive,frazier2021loss}.

The most general version of~\eqref{equa:prior:bayes} we are aware of is:
\begin{equation}
\label{rule:of:three}
\hat{\rho} = \argmin_{\rho\in\mathcal{F}} \left\{ \mathbb{E}_{\theta\sim\rho}\left[r(\theta)\right] + \frac{D(\rho\|\pi)}{n}    \right\}
\end{equation}
where $D$ is any distance or divergence between probability distributions. Here, the Bayesian point of view is generalized in three directions:
\begin{enumerate}
 \item the negative log-likelihood is replaced by a more general notion of risk $r(\theta)$, as in PAC-Bayes bounds and in Gibbs posteriors,
 \item the minimization over $\mathcal{P}(\Theta)$ is replaced by the minimization over $\mathcal{F}$, in order to keep things tractable, as in variational inference,
 \item finally, the $KL$ divergence is replaced by a general $D$. Note that this already happened in Theorem~\ref{bound:chi2} above.
\end{enumerate}
This triple generalization, and the optimization point of view on Bayes\-ian statistics is strongly advocated in~\citet{kno2019} (in particular reasons to replace $KL$ by $D$ are given in this paper that seem to me more relevant in practice than Theorem~\ref{bound:chi2}).

In this spirit, \citet{rodriguez2021tighter,chee2021learning} provided PAC-Bayes type bounds where $D$ is the Wasserstein distance.

\begin{remark}
When $D\neq KL$,~\eqref{rule:of:three} is no longer necessarily equivalent to
\begin{equation}
\label{rule:of:three:NO}
\hat{\rho} = \argmin_{\rho\in\mathcal{F}} D\left(\rho\| \pi(\cdot|X_1,\dots,X_n) \right).
\end{equation}
\citet{kno2019} discusses why~\eqref{rule:of:three} is a more natural generalization,
and~\citet{geffner2020difficulty} shows that~\eqref{rule:of:three:NO} leads to difficult minimization problems. Note however that there are also some theoretical results on~\eqref{rule:of:three:NO} by~\citet{jai2020}.
\end{remark}

\section{Empirical Risk Minimization}
\label{subset:ERM:final}

We already pointed out in the introduction the link between empirical risk minimization (based on PAC bounds) and PAC-Bayes.

When the parameter space $\Theta$ is not finite as in Theorem~\ref{thm:erm} above, the $\log(M)$ term is replaced by a measure of the complexity of $\Theta$ called the Vapnik-Chervonenkis dimension (VC-dim). We simply mention that \citet[Section 2]{cat2003} and \citet[Chapter 3, page 115--130]{cat2007} builds a well-chosen data dependent prior such that the VC-dim of $\Theta$ appears explicitly in the PAC-Bayes bound. However, \citet{PACBAYESLIMITATION} provide an example where the VC dimension is finite, yet the PAC-Bayes approach fails. This problem was solved recently by \citet{grunwald2021pac} thanks to ``conditional PAC-Bayes bounds''. This is discussed below together with Mutual Information bounds.

Similarly, generalization bounds for Support Vectors Machines are based on a quantity called the margin. This quantity can also appear in PAC-Bayes bounds~\citep{lan2002,herbrich2002pac,cat2007,biggs2021margins}.

\enlargethispage{\baselineskip}
Audibert and Bousquet studied a PAC-Bayes version of the chaining argument~\citep{audibert2007combining}. See also versions based on Mutual Information bounds~\citep{NEURIPS2018_8d7628dd,clerico2022chained}.

Finally, \citet{NIPS2008_5b69b9cb,yang2019fast} proved PAC-Bayes bounds using Rademacher complexity.

\section{Non-Bayesian Estimators}
\label{subset:ERM:final:bis}

More generally, Catoni developped recently a technique to prove upper bound on the risk of non-Bayesian estimators that relies on PAC-Bayes bounds. Given an estimator $\hat{m}$ of some vector $m^*\in\mathbb{R}^d$, put $B_d = \{v\in\mathbb{R}^d: \|v\|\leq 1\}$, then
\begin{align*}
\|\hat{m}-m^*\| & = \sup_{v\in B_d} \left< v, \hat{m}-m^* \right>
\\
& = \sup_{\rho \in\mathcal{P}(B_d) }  \left<  \mathbb{E}_{v\sim \rho }[v], \hat{m}-m^* \right>
\\
& =  \sup_{\rho \in\mathcal{P}(B_d) } \Bigl\{ \mathbb{E}_{v\sim \rho } [ \left<  v, \hat{m}\right> ] - \mathbb{E}_{v\sim \rho }[ \left<  v, m^* \right> ] \Bigr\}.
\end{align*}
Thus, we can work with PAC-Bayes bound to control $\|\hat{m}-m^*\|$. The technique is detailed by~\citet{catoni2017dimension} to provide robust estimation of the Gram matrix (for PCA), an infinite-dimensional version for kernel-PCA was then provided by~\citet{giu2018}. This builds on earlier work on robust estimation~\citet{aud2011,cat2012}. More results on the estimation of the covariance matrix were proven since using this technique~\citep{zhivotovskiy2021dimension,nakakita2022dimension}.

\section{Online Learning}

\subsection{Sequential prediction}

Sequential classification focuses on the following problem. At each time step $t$,
\begin{itemize}
 \item a new object $x_t$ is revealed,
 \item the forecaster must propose a prediction $\hat{y}_t$ of the label of $y_t$,
 \item the true label $y_t\in\{0,1\}$ is revealed and the forecaster incurs a loss $\ell(\hat{y}_t,y_t)$, and updates his/her knowledge.
\end{itemize}
Similarly, online regression and other online prediction problems are studied.

Prediction strategies are often evaluated through upper bounds on the regret $\mathcal{R}eg(T)$ given by:
$$
\mathcal{R}eg(T) := \sum_{t=1}^{T} \ell(\hat{y}_t,y_t) - \inf_{\theta} \ell(f_\theta(x_t),y_t)
$$
where $\{f_\theta,\theta\in\Theta\}$ is a family of predictors as in Section~\ref{section:introduction} above. However, a striking point is that most regret bounds hold without any stochastic assumption on the data $(x_t,y_t)_{t=1,\dots,T}$: they are not assumed to be independent nor to have any link whatsoever with any statistical model. On the other hand, assumptions on the function $\theta\mapsto \ell(f_\theta(x_t),y_t)$ are unavoidable (depending on the strategies: boundedness, Lipschitz condition, convexity, strong convexity, etc.).

A popular strategy, strongly related to the PAC-Bayesian approach, is the exponentially weighted average (EWA) forecaster, also known as weighted majority algorithm or multiplicate update rule~\citep{lit1989,vov1990,ces1997,kiv1999} (we also refer to \citet{bar1974} on the halving algorithm that can be seen as an ancestor of this method). This strategy is defined as follows. First, let $\rho_1 = \pi$ be a prior distribution on $\Theta$, and fix a learning rate $\eta>0$. Then, at each time step $t$:
\begin{itemize}
 \item the prediction is given by
 $$ \hat{y}_t = \mathbb{E}_{\theta\sim \rho_t} [f_{\theta}(x)] ,$$
 \item when $y_t$ is revealed, we update
 $$ \rho_{t+1}({\rm d}\theta )  = \frac{{\rm e}^{-\eta \ell(f_{\theta}(x),y_t) }\rho_t({\rm d}\theta)  }{ \int_\Theta {\rm e}^{-\eta \ell(f_{\vartheta}(x),y_t) }\rho_t({\rm d}\vartheta) } .$$
\end{itemize}

We provide here a simple regret bound from \citet{ces2006} (stated for a finite $\Theta$ but the extension is direct). Note the formal analogy with PAC-Bayes bounds.
\begin{theorem}
\label{thm:regret}
 Assume that, for any $t$, $0\leq \ell(f_\theta(x_t),y_t) \leq C$ (bounded loss assumption) and $\theta\mapsto \ell(f_\theta(x_t),y_t)$ is a convex function. Then, for any $T>0$,
 $$
 \sum_{t=1}^{T} \ell(\hat{y}_t,y_t) \leq \inf_{\rho\in\mathcal{P}(\Theta)} \left\{ \mathbb{E}_{\theta\sim\rho} [\ell(f_\theta(x_t),y_t)]
 +
 \frac{\eta C^2 T}{8} + \frac{KL(\rho\|\pi)}{\eta} \right\}.
 $$
 In particular, when $\Theta$ is finite with ${\rm card}(\Theta)=M$ and $\pi$ is uniform, the restriction of the infimum to Dirac masses leads to
  $$
 \sum_{t=1}^{T} \ell(\hat{y}_t,y_t) \leq \inf_{\theta\in\Theta}  \ell(f_\theta(x_t),y_t)
 +
 \frac{\eta C^2 T}{8} + \frac{\log(M)}{\eta}
 $$
 and thus with $\eta = \frac{2}{C}\sqrt{\frac{2\log(M)}{T}}$,
 $$
\mathcal{R}eg(T) \leq C \sqrt{\frac{T\log(M)}{2}}.
$$
\end{theorem}
Choices of $\eta$ that do not depend on the time horizon $T$ are possible~\citep{ces2006}. Smaller regret bounds, up to $\log(T)$ or even constants, are known under stronger assumptions~\citep{ces2006,aud2009}. We refer the reader to \citet{shalev2011online,orabona2019modern} for more up to date introductions to online learning.

While there is no stochastic assumption on the data in Theorem~\ref{thm:regret}, it is possible to deduce inequalities in probability or in expectation from it under an additional assumption (for example, the assumption that the data is i.i.d.). This is described for example in \citet[Chapter 5]{shalev2011online}, but does not always lead to optimal rates. For more up-to-date discussion on this topic with more general results, see~\citet{bil2020}. It is also possible to use tools from sequential predictions to derive PAC-Bayes bounds~\citep{jang2023tighter}.

Finally, we mention that many other strategies than EWA are studied: online gradient algorithm, follow-the-regularized-leader (FTRL), online mirror descent (OMD)... EWA is actually derived as a special case of FTRL and OMD in many references~\citep{shalev2011online} and conversely, \citet{hoeven2018many,khan2021bayesian} derive OMD and online gradient as approximations of EWA. Regret bounds for variational approximations of EWA were proved by~\citet{che2019b}. More generally,~\citet{NEURIPS2022_a4d991d5,neu2023online2pac} proved PAC-Bayes regret bounds, that allow to recover regret bounds for EWA, variational approximations and potentially more methods. Here, we point out a similarity to Section~\ref{section:first:step}: we remarked that one can use PAC-Bayes inequalities to provide generalization error bounds on non-Bayesian methods like the ERM. \citet{haddouche2023pac,rod2023} recently used this approach to prove PAC-Bayes bounds with heavy-tailed losses.

\subsection{Bandits and reinforcement learning (RL)}

Other online problems received considerable attention in the past few years. In bandits, the forecaster only receives feedback on the loss of his/her prediction, but not on the losses what he/she would have incurred under other predictions. We refer the reader to \citet{MAL-024} for an introduction to bandits. Note that some strategies used for bandits are derived from EWA. Some authors derived strategies or regret bounds directly from PAC-Bayes bounds~\citep{sel2011,pmlr-v26-seldin12a}. There are also PAC-Bayes bounds for offline bandits~\citep{pmlr-v97-london19a,sakhi2022pac,aouali2023exponential}. Bandits themself are a subclass of a larger family of learning problems: reinforcement learning (RL). PAC-Bayes bounds or related generalization bounds for RL were derived by \citet{pmlr-v97-wang19o,tasdighi2023pac}.

\section{Aggregation of Estimators in Statistics}
\label{subsection:aggregation}

Given a set $\mathcal{E}$ of statistical estimators, the aggregation problem consists of finding  a new estimator, called the aggregate, that would perform as well as the best estimator in $\mathcal{E}$, see \citet{nem2000} for a formal definition and variants of the problem. The optimal rates are derived by \citet{tsybakov2004optimal}. Many aggregates share a formal similarity with the EWA of online learning and with the Gibbs posterior of the PAC-Bayes approach~\citep{nem2000,jud2000,yan2001,meir2003generalization,yang2004aggregating,cat2004,zha2006,leu2006,lecue2007aggregation,dal2008,bun2008,suz2012,dal2012b,dai2012deviation,dal2012,dai2014aggregation,dal2018,luu2019pac,DalEWA2018}. In some of these papers, the connection to PAC-Bayes bounds is explicit, for example \citet[Theorem 1]{dal2008}, which is actually an oracle PAC-Bayes bound in expectation. It leads to fast rates in the spirit of Theorem~\ref{thm:oracle:bound}, but with different assumptions (in particular, the $X_i$'s are not random there).

\section{Information Theoretic Approaches}

A note on the terminology: a huge number of statistical and machine learning results mentioned above rely on tools from information theory: \citet{zha2006} actually proves beautiful PAC-Bayesian bounds under the name ``information theoretic bounds'' (maybe for this reason, it's not always listed with the other PAC-Bayes bounds in the publications on the topic). My goal here is not to classify what should be called ``PAC-Bayes'' and what should not, I'm certainly not qualified for that. I simply want to point out the connection to two families of methods inspired directly from information theory.

\subsection{Minimum description length}

In Rissanen's Minimum Description Length (MDL) principle, the idea is to penalize the empirical error of a classifier by its shortest description~\citep{ris1978}. We refer the reader to \citet{bar1998,grunwald2007minimum} for more recent presentations of this very fruitful approach. Note that given a prefix-free code on a finite alphabet $\Theta$, it is possible to build a probability distribution $\pi(\theta) \simeq 2^{-L(\theta)}$ where $L(\theta)$ is the length of the code of $\theta$, so codes provide a way to define priors in PAC-Bayes bounds, see for example \citet[Chapter 1]{cat2004}.

\subsection{Mutual information bounds (MI)}
\label{subsubsec:MI}
\enlargethispage{\baselineskip}
Recently, some generalization error bounds appeared where the complexity is measured in terms of the mutual information between the sample and the estimator.
\begin{definition}
 Let $U$ and $V$ be two random variables with joint probability distribution $P_{U,V}$. Let $P_U$ and $P_V$ denote the marginal distribution of $U$ and $V$ respectively. The mutual information (MI) between $U$ and $V$ is defined as:
 $$ I(U,V) := KL( P_{U,V} \| P_U\otimes P_V  ). $$
\end{definition}

\citet{russo2019much} introduced generalization error bounds (in expectation) that depend on the mutual information between the predictors and the labels. In particular, when $\hat{f}$ is obtained by empirical risk minimization, they recover bounds depending of the VC-dimension of $\Theta$.

Russo and Zou's result was improved and extended by \citet{raginsky2016information,xu2017MI}. In particular, \citet[Subsection 4.3]{xu2017MI} proved a powerful MI inequality, from which many existing and new results can  be derived. Unfortunately, it is not pointed out that the results on Gibbs posteriors could also be obtained directly from a PAC-Bayes bound in expectation such as Theorem~\ref{thm:first:bound:exp} above. Thus, the connection between information bounds and PAC-Bayes bounds might have been missed by part of the information bounds community (in the same way that part of the PAC-Bayes community might have missed the ``information theoretic bounds'' of \citet{zha2006}).

It turns out that \citet{langford2003microchoice} discussed the minimization of a PAC-Bayes bound in expectation (such as Theorem~\ref{thm:first:bound:exp} above) with respect to the prior $\pi$. We will see below that the optimal prior turns out to be $\pi_{-\lambda R}$ (which is the prior that appears in Catoni's localized bounds such as Theorem~\ref{thm:oracle:bound:local} above). This idea also appears in~\citet[page 14 and 51]{cat2007}, where it is also noticed that the optimized $KL$ term boils down to the mutual information between the sample and the parameter. All these derivations will be detailed below, and essentially lead to an inequality similar to the one of~\citet{russo2019much}.

More recent work on MI bounds mention explicitly the connection to PAC-Bayes bounds~\citep{negrea2019information,dziugaite2021role,banerjee2021information}. Let us for example cite a result state by \citet{negrea2019information}, or rather, a simplified version (by setting their parameter $m$ to $0$).
\begin{theorem}[Theorem 2.3, \citet{negrea2019information}, with $m=0$]
Using the notations and assumptions of Section~\ref{section:introduction}, assume that the losses $\ell_i(\theta)$ are sub-Gaussian with parameter $C$, then, for any data-dependent $\tilde{\rho}$,
$$
\mathbb{E}_{\mathcal{S}}\Bigl\{
\mathbb{E}_{\theta\sim\tilde{\rho}}[ R(\theta)] - \mathbb{E}_{\theta\sim\tilde{\rho}}[ r(\theta)]
\Bigr\}
\leq
\sqrt{\frac{2C I(\theta,\mathcal{S}) }{n}}
\leq
\sqrt{\frac{2C \mathbb{E}_{\mathcal{S}}[KL(\tilde{\rho}\|\pi)]}{n}}
.
$$
\end{theorem}
A few comments on this result:
\begin{itemize}
 \item the first inequality is from \citet[Theorem 1]{xu2017MI}. Note however that Theorem 2.3 of \citet{negrea2019information} contains more information, as setting their parameter $m\neq 0$ allows to get a data-dependent prior. The paper contains more new results, and a beautiful application to derive empirical bounds on the performance of stochastic gradient descent (SGD). On this topic, see also the recent works by \citet{bu2020tightening,NEURIPS2020712a3c98,wang2021pac,neu2021information,haghifam2023limitations}.
 \item we can see here that the MI bound
 \begin{equation}
 \label{bound:mi:mi}
 \mathbb{E}_{\mathcal{S}}\Bigl\{
\mathbb{E}_{\theta\sim\tilde{\rho}}[ R(\theta)] - \mathbb{E}_{\theta\sim\tilde{\rho}}[ r(\theta)]
\Bigr\}
\leq
\sqrt{\frac{2C I(\theta,\mathcal{S}) }{n}}
 \end{equation}
is tighter than the the PAC-Bayes bound in expectation
 \begin{equation}
  \label{bound:mi:pacbayes}
\mathbb{E}_{\mathcal{S}}\Bigl\{
\mathbb{E}_{\theta\sim\tilde{\rho}}[ R(\theta)] - \mathbb{E}_{\theta\sim\tilde{\rho}}[ r(\theta)]
\Bigr\}
\leq
\sqrt{\frac{2C \mathbb{E}_{\mathcal{S}}[KL(\tilde{\rho}\|\pi)]}{n}}
.
 \end{equation}
However, an MI bound cannot be used as is in practice. Indeed, $I(\theta,\mathcal{S})$ depends on the distribution of the sample $\mathcal{S}$ that is unknown in practice. In order to use~\eqref{bound:mi:mi}, one must upper bound $I(\theta,\mathcal{S})$ by a quantity that does not depend on the sample.
\item \citet[page 14 and 51]{cat2007} discuss the optimization of PAC-Bayes bounds with respect to the prior. In order to explain the discussion done there, apply $\sqrt{ab}\leq a/(2\lambda) + b\lambda/2 $ to~\eqref{bound:mi:pacbayes} to get a ``Catoni style'' bound:
$$
\mathbb{E}_{\mathcal{S}}\Bigl\{
\mathbb{E}_{\theta\sim\tilde{\rho}}[ R(\theta)] - \mathbb{E}_{\theta\sim\tilde{\rho}}[ r(\theta)]
\Bigr\}
\leq
\frac{C\lambda}{2 n} + 
\frac{ \mathbb{E}_{\mathcal{S}}[KL(\tilde{\rho}\|\pi)]}{\lambda}
$$
and thus
$$
\mathbb{E}_{\mathcal{S}}
\mathbb{E}_{\theta\sim\tilde{\rho}}[ R(\theta)]
\leq
\mathbb{E}_{\mathcal{S}}\left\{  \mathbb{E}_{\theta\sim\tilde{\rho}}[ r(\theta)] + \frac{C\lambda}{2 n} + 
\frac{ KL(\tilde{\rho}\|\pi)}{ \lambda}  \right\}
$$
(compare to Theorem~\ref{thm:first:bound:exp} above).
Let $\tilde{\rho}$ be any data-dependent measure that is absolutely continuous with respect to $\pi$ almost-surely, thus
$$ \frac{{\rm d}\tilde{\rho}}{{\rm d}\pi}(\theta) $$
is well-defined. Catoni defines $\mathbb{E}_{\mathcal{S}}(\tilde{\rho}) $ the probability measure defined by
$$
\frac{{\rm d}\mathbb{E}_{\mathcal{S}}(\tilde{\rho})}{{\rm d}\pi}(\theta) = \mathbb{E}_{\mathcal{S}}\left(\frac{{\rm d}\tilde{\rho}}{{\rm d}\pi}(\theta)\right).
$$
Direct calculations show that $ \mathbb{E}_{\mathcal{S}}[KL(\tilde{\rho}\|\pi)] =  \mathbb{E}_{\mathcal{S}}[KL(\tilde{\rho}\|\mathbb{E}_{\mathcal{S}}(\tilde{\rho}))] + KL(\mathbb{E}_{\mathcal{S}}(\tilde{\rho})\|\pi) = I(\theta,\mathcal{S}) +  KL(\mathbb{E}_{\mathcal{S}}(\tilde{\rho})\|\pi) $ and thus:
$$
\mathbb{E}_{\mathcal{S}}
\mathbb{E}_{\theta\sim\tilde{\rho}}[ R(\theta)]
\leq
\mathbb{E}_{\mathcal{S}}\left\{  \mathbb{E}_{\theta\sim\tilde{\rho}}[ r(\theta)] \right\} + \frac{C\lambda}{2 n} + 
\frac{I(\theta,\mathcal{S}) +  KL(\mathbb{E}_{\mathcal{S}}(\tilde{\rho})\|\pi) }{ \lambda}.
$$
So the choice to replace $\pi$ by $\mathbb{E}_{\mathcal{S}}(\tilde{\rho})$ gives the MI bound:
$$
\mathbb{E}_{\mathcal{S}}
\mathbb{E}_{\theta\sim\tilde{\rho}}[ R(\theta)]
\leq
\mathbb{E}_{\mathcal{S}}\left\{  \mathbb{E}_{\theta\sim\tilde{\rho}}[ r(\theta)] \right\} + \frac{C\lambda}{2 n} + 
\frac{I(\theta,\mathcal{S}) }{ \lambda}.
$$
The choice $\lambda = \sqrt{2 n \mathcal{I}(\theta,\mathcal{S}) / C}$ leads to
$$
\mathbb{E}_{\mathcal{S}}
\mathbb{E}_{\theta\sim\tilde{\rho}}[ R(\theta)]
\leq
\mathbb{E}_{\mathcal{S}}\left\{  \mathbb{E}_{\theta\sim\tilde{\rho}}[ r(\theta)] \right\} + \sqrt{\frac{2 C \mathcal{I}(\theta,\mathcal{S})}{n}}.
$$
In other words, MI bounds can be seen as PAC-Bayes bounds optimized with respect to the prior. Of course, as we said above, MI bound cannot be computed in practice. Catoni proposes an interpretation of his localization technique as taking the prior $\pi_{-\beta R}$ to approximate $\mathbb{E}_{\mathcal{S}}(\pi_{-\lambda r})$, and then to upper bound $KL(\rho\|\pi_{-\beta R})$ via empirical bounds. As we have seen in Section~\ref{section:tight}, this leads to empirical bounds with data-dependent priors, and in Section~\ref{section:oracle}, this leads to improved PAC-Bayes oracle bounds. All this is pointed out by \citet{grunwald2021pac}: ``Catoni already mentions that the prior that minimizes a MAC-Bayesian bound is the prior that turns the KL term into the mutual information''.
\end{itemize}

Similarly to PAC-Bayesian bounds, MI bounds can be stated with other divergences than KL~\citep{neu2022MI}.

Thanks to MI bounds, it is also possible to provide an exact formula (not an upper bound) for the generalization error of the Gibbs posterior in terms of the symmetrized version of the KL bound~\citep{aminian2021characterizing}.

Since \citet{pmlr-v83-bassily18a,pmlr-v75-nachum18a}, it is known that MI bounds can fail in some situations where the VC dimension is finite: thus, they suffer the same limitation as PAC-Bayes bounds proven by \citet{PACBAYESLIMITATION}. Recently, expanding on ideas from the PAC-Bayes literature~\citep{audPHD,cat2007,mhammedi2019pac} and in the MI literature \citep{steinke2020reasoning,hellstrom2020generalization}, \citet{grunwald2021pac} unified MI bounds and PAC-Bayes bounds, an they developped ``conditional'' MI and PAC-Bayes bounds. Note in particular that conditional PAC-Bayes bounds are based on the notion of ``exchangeable priors'' introduced by~\citet[Chapter 2 page 31]{cat2003}, see also~\citet[page 105]{audPHD} and developped by~\citet[Definition 3.1.1 page 11]{cat2007} under the name ``exchangeable posteriors'' (to highlight the fact that these distributions can be data-dependent).

The bounds of \citet{grunwald2021pac} are proven to be small for any set of classifiers with finite VC dimension. Thus, they don't suffer the limitations of PAC-Bayes and MI bounds of \citet{pmlr-v83-bassily18a,pmlr-v75-nachum18a,PACBAYESLIMITATION}. To cite the results of \citet{grunwald2021pac} would go beyond the framework of this ``easy introduction'', but one of the main point of this tutorial is to prepare the reader not familiar with PAC-Bayes bounds, Bernstein assumption etc. to read this work. Recent work on sharp mutual information bounds and their connection to minimax rates in classification include~\citet{NEURIPS2021ddbc86dc,haghifam2022understanding}.

\chapter{Conclusion}

We hope that the reader:
\begin{itemize}
 \item has a better view of what a PAC-Bayes bound is, and what can be done with such a bound,
 \item is at least a little convinced that these bounds are quite flexible, that they can be used in a wide range of contexts, and for different objectives in ML,
 \item wants to read many of the references listed above, that provide tighter bounds and clever applications.
\end{itemize}
I believe that PAC-Bayes bounds (and all the related approaches, including mutual information bounds, etc.) will play an important role in the study of deep learning, in the wake of \citet{dzi2017}, in RL~\citep{pmlr-v97-wang19o} and in meta-learning~\citep{pen2014,ami2018,rot2020}.

\begin{acknowledgements}

First, a huge ``thank you'' to both anonymous referees for their careful reading of this manuscript. Their insightful comments helped a lot to improve it.

My PhD advisor Olivier Catoni taught me PAC-Bayes bounds, and so much more.

Of course I learn also about PAC-Bayes and related topics from all my co-authors, friends, students and twitter pals... that I will not even try to make a list. Thanks to all of you, you know who you are!

\enlargethispage{\baselineskip}
Still, I would like to thank specifically, for motivating, providing valuable feedback and helping to improve the manuscript (since the very first draft of Section~\ref{section:first:step}): Mathieu Alain, Pradeep Banerjee, Wessel Bruinsma, David Burt, Badr-Eddine Ch\'erief-Abdellatif, Nicolas Chopin, Arnak Dalalyan, Andrew Foong, Natalie Frank, Avrajit Ghosh, Avetik Karagulyan, Aryeh Kontorovich, The Tien Mai, Thomas M\"ollenhoff, Peter Nickl, Donlapark Ponnoprat, Charles Riou, and Alexandre Tsybakov. A very first draft of Section~\ref{section:first:step} was started in September-October 2008 as I was invited at ENSAE by Alexandre Tsybakov to give talks on PAC-Bayes bounds. I taught a course on PAC-Bayes bounds and online learning at ENSAE Paris from 2014 to 2019, partially based on this preliminary version. I should thank all students for enduring this. Most of the document was written when I was working at RIKEN AIP (Tokyo) in the Approximate Bayesian Inference team led by Emtiyaz Khan: I thank Emti and the whole team for their support.

\end{acknowledgements}                               

%BACKMATTER SEE DOCUMENTATION
\backmatter  % references, restarts sample

\sloppy
\printbibliography

\end{document}